\documentclass[mnsc,nonblindrev]{informs3_without_head}

\usepackage[utf8]{inputenc}
\usepackage[T1]{fontenc}
\usepackage[english]{babel}
\usepackage{microtype}
\usepackage{mathrsfs,mathtools,accents}
\usepackage{xifthen,enumitem,comment}
\setlist[enumerate,1]{label=\normalfont{(\Roman*)},leftmargin=*}
\usepackage{xcolor}
\usepackage{etoolbox}
\makeatletter
\patchcmd{\env@cases}{1.2}{0.96}{}{}
\makeatother
\usepackage{algorithm,algorithmicx,algpseudocode}
\usepackage[hypertexnames=false]{hyperref}
\hypersetup{%
  colorlinks=true,
  linkcolor={blue!66!black},
  linkbordercolor=red,
  citecolor={blue!66!black},
}
\usepackage{tikz,pgfplots,subfigure}
\usetikzlibrary{matrix,decorations.pathreplacing}

\ifx\argmax\undefined\DeclareMathOperator*{\argmax}{arg\,max}\fi
\ifx\argmin\undefined\DeclareMathOperator*{\argmin}{arg\,min}\fi

\newcommand*{\QEG}{%
\leavevmode\unskip\penalty9999 \hbox{}\nobreak\hfill
    \quad\hbox{$\clubsuit$}%
}

\renewcommand{\Pr}{\mathbb{P}}
\providecommand{\E}{\mathbb{E}}

\providecommand{\cA}{\mathcal{A}}
\providecommand{\cB}{\mathcal{B}}
\providecommand{\cC}{\mathcal{C}}
\providecommand{\cD}{\mathcal{D}}

\providecommand{\cF}{\mathcal{F}}

\providecommand{\cI}{\mathcal{I}}

\providecommand{\cL}{\mathcal{L}}

\providecommand{\cO}{\mathcal{O}}

\providecommand{\cS}{\mathcal{S}}

\providecommand{\cX}{\mathcal{X}}

\usepackage{subcaption}

\renewcommand{\Pr}{\mathbb{P}}
\providecommand{\E}{\mathbb{E}}

\providecommand{\cA}{\mathcal{A}}

\providecommand{\cF}{\mathcal{F}}
\providecommand{\cX}{\mathcal{X}}

\providecommand{\cD}{\mathcal{D}}

\providecommand{\cI}{\mathcal{I}}

\providecommand{\cL}{\mathcal{L}}

\providecommand{\cB}{\mathcal{B}}

\providecommand{\cS}{\mathcal{S}}

\providecommand{\CQPC}{\textsf{CQPC}}

\providecommand{\bs}{\boldsymbol{s}}

\providecommand{\dis}{\xi}

\providecommand{\lgamma}{\underline{\gamma}}
\providecommand{\lowh}{\underline{h}}
\providecommand{\barh}{\bar{h}}
\providecommand{\bgamma}{\bar{\gamma}}

\providecommand{\ugamma}{\overline{\gamma}}

\providecommand{\tDelta}{\tilde{\Delta}}

\providecommand{\gtlc}{\textsf{GTLC}}
\newcommand{\htheta}{\hat{\theta}}

\providecommand{\tdis}{\tilde{\xi}}

\providecommand{\tDelta}{\tilde{\Delta}}

\providecommand{\CPRP}{\textsf{CPRP}}

\definecolor{longhorn}{rgb}{0.8, 0.33, 0.0}

\definecolor{airforceblue}{rgb}{0.36, 0.54, 0.66}
\definecolor{tan}{rgb}{0.82, 0.71, 0.55}
\definecolor{palebrown}{rgb}{0.6, 0.46, 0.33}

\usepackage{natbib}
 \bibpunct[, ]{(}{)}{,}{a}{}{,}%
 \def\bibfont{\small}%
\TheoremsNumberedThrough     
\ECRepeatTheorems

\EquationsNumberedThrough    

\begin{document}


 \RUNAUTHOR{Junyu Cao}

\RUNTITLE{A Conformal Approach to Feature-based Newsvendor under Model Misspecification}

\TITLE{A Conformal Approach to Feature-based Newsvendor under Model Misspecification}

\ARTICLEAUTHORS{%
\AUTHOR{Junyu Cao}
\AFF{McCombs School of Business, The University of Texas at Austin. \EMAIL{junyu.cao@mccombs.utexas.edu}} 
} 

\ABSTRACT{%
In many data-driven decision-making problems, performance guarantees often depend heavily on the correctness of model assumptions. Unfortunately, these assumptions often fail in practice. We address this issue in the context of a feature-based newsvendor problem, where demand is influenced by observed features, such as demographics and seasonality. To mitigate the effect of model misspecification, we propose a model-free and distribution-free framework inspired by conformal prediction. Our approach consists of two phases: a training phase, which can use any type of prediction method, and a calibration phase that conformalizes the model bias.
To enhance predictive performance, we explore the balance between data quality and quantity, recognizing the inherent trade-off: More selective training data can improve quality but reduce quantity. Importantly, we provide statistical guarantees for the optimal solution, referred to as the conformalized critical quantile, independent of the correctness of the underlying model. Moreover, we quantify the confidence interval of the conformalized critical quantile, so that its width decreases as data quality and quantity improve. We validate our framework using both simulated data and a real-world dataset from the Capital Bikeshare program in Washington, DC. Across these experiments, our proposed method consistently outperforms benchmark algorithms, reducing newsvendor loss by up to 38.6\% on the simulated data and 47.3\% on the real-world dataset.
}%



\maketitle

\section{Introduction}

The newsvendor problem aims to optimize inventory levels by balancing the costs of overstocking and understocking in the presence of demand uncertainty. Demand is often heavily influenced by various features. For instance, bike-sharing demand varies according to geographic and demographic factors; ice cream sales fluctuate with seasonal changes; and major events, like the Olympics or World Cup, can significantly boost demand for sports equipment. Therefore, considering these varying features when determining optimal inventory levels is essential. We call it the feature-based newsvendor problem.

When the demand distribution is known, the optimal inventory level, referred to as the \emph{critical quantile}, can be explicitly computed. However, in practice, the demand distribution is unknown. With the growing availability of data, companies increasingly rely on the data to make decisions. Two critical questions arise: First, how can we improve the prediction accuracy of the critical quantile? Second, given the inherent uncertainty in these predictions, how can we effectively quantify the confidence interval of the quantile predictor? The overall goal in posing these questions is to leverage historical data on observed features and demand to make the most informed decisions. 

To answer the first question, several studies have focused on addressing the quantile prediction problem by introducing simplifying model assumptions to make the problem tractable. One commonly adopted approach is the use of a parametric model, such as the linear model; here, the assumptions are that all contexts share the same set of parameters and that demand depends linearly on the features. However, this approach may oversimplify the true underlying relationships and lead to issues of model misspecification.
Other studies impose assumptions on the joint distribution of the features $X$ and demand $Y$ or assume that the noise belongs to a specific family of distributions, such as the exponential family. These works investigate efficient and accurate methods for estimating the parameters within these distribution families. Another common approach in the literature is to assume that the noise in the regression model is independent of the context or decision, suggesting that the variability in outcomes remains constant across different contexts and decisions. However, this assumption may not hold true in practical scenarios. For example, demand variability for a product might be significantly higher during special events, such as Black Friday.

In this work, we address these challenges by leveraging ideas from conformal prediction. Conformal prediction is a technique that constructs prediction intervals with valid coverage for finite samples, but without relying on specific distributional or model assumptions. It allows any heuristic notion of uncertainty from any model to be converted into a statistically rigorous measure. The core concept involves training a regression model on a set of training samples and then using the residuals from a held-out validation set to quantify uncertainty in future predictions. We adopt a similar approach, aiming to conformalize model bias.

Conformal prediction offers several advantages, such as ease of implementation and rigorous statistical guarantees; but it cannot be directly applied to the feature-based newsvendor problem. Specifically, conformal prediction is designed to construct a marginal distribution-free prediction interval $\cC(X_{n+1})\subseteq \mathbb{R}$ that has a desired miscoverage rate $\alpha$, ensuring that $\Pr(Y_{n+1}\in \cC(X_{n+1}))\geq 1-\alpha$. However, the objective in the newsvendor problem differs: We seek to predict the critical quantile and the confidence interval of the optimal decision, whereas conformal prediction constructs a confidence interval for the response (i.e., the demand in our context). Moreover, the statistical guarantee that conformal prediction provides is unconditional; thus, it accounts for variability across all contexts but does not guarantee coverage for each individual context $X$. In the newsvendor problem, decisions are made after observing the feature $X$, and ensuring that the decision informed by the predicted quantile performs well for every context is crucial. Another significant limitation of conformal prediction is that the length of the prediction interval is fixed and independent of $X$. Although recent studies have explored the effect of the underlying model on interval length and have tried to create intervals with locally varying lengths, existing methods often result in conformal intervals of either a fixed length or lengths that depend only weakly on the predictors.

In our proposed framework, the data are divided into two sets: a training set to build the prediction model and a calibration set to conformalize and correct the model. We allow for any method of quantile prediction in the training stage—for example, using parametric or nonparametric method, or a black-box algorithm. The prediction model’s function class may be misspecified, leading to biases that cause predictions for some features to be skewed upward and others downward. Our goal is to enhance the performance of quantile prediction by conformalizing and correcting these biases. The fundamental idea is that biases in similar contexts tend to be consistent, allowing us to leverage this consistency to adjust for errors in the predicted quantile. This  raises the question of how to select the pooling region when predicting is conditional on a given context. A larger pooling region allows for more data to be used but may introduce greater variability in bias within the region. Conversely, a smaller pooling region may result in higher quality data but at the expense of a smaller sample size. Thus, we see that the trade-off between the data's quality and quantity must be carefully balanced.

To answer the second question,  our method goes beyond improving the estimation of the critical quantile; it also provides a confidence bound, offering lower and upper limits within which the true quantile is likely to fall. Accurately quantifying the uncertainty of the critical quantile is crucial because different scenarios may require either a more conservative approach or a more optimistic strategy, depending on the context. The confidence interval not only indicates the optimal predicted quantile but also reflects the level of confidence in that prediction. As we demonstrate, the interval length decreases as both the quality and the quantity of data improve.

\subsection{Our Contributions}

Our study contributes to the literature in the following ways:

\begin{enumerate}
  \item \emph{Modeling}: Model misspecification is a common issue that exists in the data-driven decision-making problems. Much of the existing literature relies on the correct model assumption to derive the theoretical guarantees. To address this challenge, we propose a model-free and distribution-free method for solving the feature-based newsvendor problem; our method can take any type of  prediction method and conformalize the bias.
  \item \emph{Theoretical contributions}: 
  \begin{enumerate}
      \item \emph{Statistical guarantees}: Unlike  many existing studies that  assume that the observational noise is independent of the context or decision, suggesting that the variability in outcomes remains constant across different contexts and decisions, we consider a setting where the noise could  be context- and decision-dependent. We provide rigorous statistical guarantees on the conformalized quantile, even in the presence of the model misspecification issue. More specifically, conditional on the feature $X$, we show that the probability of the demand's lying below the conformalized quantile is centered around the critical quantile, and the gap shrinks with the improvement of data quality and data quantity. 
      \item \emph{Confidence interval of critical quantile}: Beyond the point estimation on the critical quantile, we also provide a confidence interval. This confidence interval provides more comprehensive information than just a point estimation does. It tells the decision maker about how confident the machine learning method is regarding the prediction on the critical quantile. During different stages of business operations, the decision maker can choose to make either more optimistic or more robust decisions.
    \item \emph{Balance between data quality and data quantity}: In our proposed framework, there are two main steps: prediction model training and calibration. In the model training step, the prediction quality for a specific context may be significantly influenced by the selection of the historical data to be pooled for training, where both the quality and the quantity matter. Guided by our theory, we propose a way to balance between the two and to characterize its properties.
  
  \end{enumerate}
  \item \emph{Numerical performance}: We conducted extensive numerical experiments on both simulated and real-world datasets to evaluate the performance of four quantile prediction algorithms: Linear Quantile Regression (LQR), Gradient Boosting (GB), Light Gradient Boosting Machine (LightGBM), and Quantile Regression Neural Network (QRNN). Our proposed method, \emph{Global Training with Local Calibration}, consistently demonstrated superior performance across all tested cases. The addition of the local calibration step reduced empirical loss by more than 35\% in certain instances.
In addition, we analyzed the effect of local pooling—specifically, how the size of the pooling region influences overall performance. We further investigated the effect of sample size, examining how varying the dataset size affects both overall performance and the optimal pooling region.

For the real-world dataset, we conducted an in-depth study using data from the Capital Bikeshare program in Washington, DC. The data include hourly bike rental demand along with features, such as temperature, humidity, and wind speed. We tested the four algorithms at quantiles 0.25, 0.5, and 0.75. Our proposed algorithm consistently outperforms the benchmarks, achieving a reduction in empirical loss of up to 47.3\%.
\end{enumerate}

\section{Literature Review}

 \emph{Data-driven Newsvendor}. 
 The newsvendor problem, in which a decision maker needs to make a trade-off between an underage cost and an overage cost, has been extensively studied in the literature \citep{gallego1993distribution,perakis2008regret,qin2011newsvendor,petruzzi1999pricing,khouja1999single}. Including no contexts in the newsvendor problem, 
 \cite{levi2007provably} analyze sampling-based policies and achieve near-optimal performance, while \cite{levi2015data} show a significantly tighter bound.
\cite{lin2022data} later relax the assumption of the probability density function of the demand and establish a logarithmic upper bound on the worst case regret of sample average approximation (SAA) over a large class of demand distributions. 

 When contextual information is available and the demand is correlated with observable features, incorporating covariates can enhance both demand forecasts and decision-making quality. For a broader survey on data-driven decision-making in revenue management, we refer the reader to \cite{chen2023frontiers}.
 \cite{ban2019big} study the feature-based ``big data'' newsvendor problem and propose algorithms based on the empirical risk minimization (ERM) principle, as well as a non-parametric algorithm based on kernel-weights optimization. They also provide a detailed comparison of other inventory papers that incorporate features, including \cite{liyanage2005practical}, \cite{see2010robust}, and \cite{hannah2010nonparametric}. The paper analytically justifies the use of features by showing that ignoring features yields inconsistent decisions.
Some papers study the newsvendor problem from the robust optimization perspective.
\cite{qi2022distributionally} recently proposed a distributionally robust framework for the conditional quantile prediction problem under the linear demand model. 
\cite{zhang2024optimal} study robust policy optimization for the feature-based newsvendor problem.  \cite{fu2021data} and \cite{liu2024newsvendor} study the newsvendor problem under some distributional ambiguity sets and explore the benefits of implementing the robust solution. \cite{liu2024newsvendor} additionally considers the optimal-transport cost for the misspecification when deriving the robust solution.

Recently, \cite{feng2023contextual} develop an operational data analytics (ODA) framework for the contextual newsvendor problem. Specifically, they identify the equivariant class of operational statistics by recognizing the inherent relationship between the demand-and-covariate data and the inventory decision. 
\cite{besbes2023contextual} analyze the broad class of weighted empirical risk minimization (WERM) policies, which weigh past data according to their similarity in the contextual space. For example, looking at the classical ERM policy (when all the weights are equal), the class of $k$-nearest neighbors policies, and kernel methods, they analyze the performance of data-driven algorithms through a notion of context-dependent worst-case expected regret. Other work develops different machine learning algorithms for making the best inventory decision with limited data, including the tree algorithm \citep{kallus2023stochastic} and deep neural networks \citep{han2023deep,oroojlooyjadid2020applying}. 

Our approach differs from these previous studies in several key aspects. First, many existing works impose specific functional forms on the demand model, in which the model can be wrong in practice. Our framework allows the model to be misspecified and even allows for a model-free form. Second, to provide the performance guarantee, some works assume that the demand or the observational noise follows a certain family of distribution, or they require that the distribution satisfy additional conditions. We relax such assumptions and even allow for heavy-tailed noise. Moreover, in our proposed framework, we can use any machine-learning algorithm, including a blackbox algorithm, to predict the quantile, and we provide rigorous theoretical guarantees for the quantile that is computed from the data.

\emph{Conformal Prediction}. The design of our framework is inspired by conformal prediction \citep{shafer2008tutorial,angelopoulos2023conformal,fontana2023conformal}. Conformal prediction uses historical data to construct confidence intervals for new predictions, without making distributional assumptions \citep{vovk2005algorithmic,vovk2009line,lei2014distribution}. The main idea is to fit a regression model on the training data and then to use the residuals from a separate validation set to measure the uncertainty in future predictions.  
The conformal prediction can use any type of regression algorithm for prediction, including random forests and deep neural networks \citep{johansson2013conformal,johansson2014regression,pearce2018high}. 
In our newsvendor problem specifically, we train the model using quantile regression \citep{koenker1978regression}. 
For more recent advances in quantile regression, we refer readers to \cite{pan2021multiplier,he2022scalable,zhang2022element,romano2019conformalized,tan2022high} and \cite{sivakumar2017high}. 

Conformal prediction methods produced prediction intervals that either were a fixed length or were weakly dependent on the predictors, leading to unnecessarily conservative intervals. To address the limitations of these existing methods, \cite{romano2019conformalized} developed conformalized quantile regression, a method that fully adapts to heteroscedasticity. 

Most conformal prediction methods typically do not provide conditional coverage guarantees, which are essential for high-stakes decision making. However, some studies have explored conditional coverage. For instance, several works modify the split conformal calibration step \citep{guan2023localized,barber2023conformal}, while others adjust the initial prediction rule \citep{chernozhukov2021distributional} to construct conditionally valid prediction intervals. \cite{foygel2021limits} and \cite{jung2022batch} propose methods that offer practical guarantees in the form of group-conditional coverage. In addition, \cite{gibbs2023conformal} introduce a novel approach to conditional coverage, reformulating it as coverage under a class of covariate shifts.

In line with previous work, our research shares a similar objective of providing performance guarantees for any given contextual information. However, our approach diverges from this body of work in several significant ways. First, the objectives differ: While the conformal prediction literature primarily focuses on constructing confidence intervals for new predictions, our aim is to optimize decision making to maximize revenue within a specific context. Second, we address the challenge of providing conditional guarantees by introducing a novel balance between data quality and data quantity. Third, we provide rigorous non-asymptotic guarantees for the computed quantiles.

\cite{sun2023predict} propose an interesting predict-then-calibrate framework: The algorithm initially builds a prediction model without considering the downstream risk profile or robustness guarantees; then, calibration (or recalibration) methods are applied to quantify the uncertainty of the prediction. However, the focus in the work of \cite{sun2023predict} is on solving a robust contextual linear programming (LP) problem in which the uncertainty set is constructed through the conformal prediction. Other work in this line of research includes \cite{patel2024conformal,chenreddy2024end}. Moreover, the coverage guarantee in their work is also with respect to a population/average sense (i.e., unconditional coverage guarantees). In our work, we provide individual-level guarantees and also discuss the trade-off between the quality and quantity of data.

\section{Model}
In this section, we present the formal problem setup and introduce the foundational concepts of conformal prediction.

\subsection{Problem Setup}
Suppose we are given $n$ training samples $\cD_n:=\{(X_i, Y_i)\}_{i=1}^n$, where $X_i$ is the contextual information and $Y_i$ is the demand. The relationship between $Y_i$ and $X_i$ is
\[
Y_i = f(X_i) + \epsilon_i,
\]
where the distribution of the noise $\epsilon_i$ potentially could depend on the contextual information $X_i$. 
We assume that all samples are drawn i.i.d.  from an arbitrary joint distribution $F_{XY}$ over the feature vectors $X\in \mathbb{R}^d$ and the response variables $Y\in \mathbb{R}$.

Given context $X_0$, which is drawn from $F_X$, the goal of the decision maker is to make the optimal decision $a\in \cA$ to minimize the expected loss,
\[
\cL(a):= \E_{Y\sim F_{Y|X_0}}[\ell_\alpha(a,Y)]\,,
\]
where $F_{Y|X_0}$ is the conditional distribution of the response, conditional on the context $X_0$; $\ell_\alpha$ is the ``check function'' or ``pinball loss,'' defined by
\[
\ell_\alpha(a,y) = \frac{c_o}{c_o+c_u} \cdot (a-y)^+ + \frac{c_u}{c_o+c_u}\cdot (y-a)^+.
\]
Here, $c_o$ denotes the over-ordering cost, and $c_u$ denotes the under-ordering cost; the  quantile level corresponding to the optimal solution is $\alpha:=c_u/(c_o+c_u)$. It can be shown that the solution that minimizes the expected loss $\cL(\cdot)$ is the $\alpha^{th}$ quantile of the demand distribution, defined as
\begin{equation}\label{eq: optimal quantile}
a^*(X_0):=\inf\{a\in \mathbb{R}| F_{Y|X_0}(a)\geq \alpha\}.
\end{equation}

As implied by Equation \eqref{eq: optimal quantile}, the optimal action relies on the contextual information because the demand distribution largely depends on the context. The additional information provided by the increasing availability of contextual information can help to optimize decisions. The following example demonstrates the value of contextual information.

\begin{example}[Value of Information -- Feature-based Forecast]
Consider the true demand model $Y_i = \theta^\top X_i + \epsilon$, where $\epsilon$ follows the standard normal distribution. If we ignore the feature and focus only on the entire demand distribution, then the optimal decision is uniform across all features. However, conditional on feature $X$, the optimal decision is $\theta^\top X+\Phi^{-1}(\alpha)$, where $\Phi^{-1}(\alpha)$ is the inverse cumulative distribution function of the standard normal distribution at level $\alpha$. When the variance of $\theta^\top X$ is large, failing to account for the feature information can lead to poor decisions.
\QEG
\end{example}

 The feature-based newsvendor problem aims to find an optimal measurable function $q_\alpha^*$, established from function class $\cF$, such that
\[
q^*_\alpha = \argmin_{f\in \cF}\  \E_{X,Y}[\ell_\alpha(f(X),Y)],
\]
and $q^*_\alpha(X_0)$ is the optimal decision, conditional on the context $X_0$.

In practice, solving Equation \eqref{eq: optimal quantile} is generally challenging because the decision maker lacks knowledge of the true demand distribution and must rely solely on the available dataset $\cD_n$. Therefore, the central challenge is determining how to leverage the dataset $\cD_n$ to make the decision $a$ that minimizes the expected loss, given the context $X$.
A well-established and widely adopted approach to estimate the quantile from the data is quantile regression.
That is, 
\begin{equation}\label{Eq: quantile regression}
\hat{q}_\alpha = \argmin_{f\in \cF} \ \frac{1}{n} \sum_{i=1}^n \ell_\alpha(f(X_i), Y_i)+ \lambda_n ,
\end{equation}
where $f(\cdot)$ is the quantile regression function that can be either parametric or non-parametric; $\cF$ is the function class; and $\lambda_n$ is a potential regularizer. Some existing papers investigate the performance of the estimator in Equation \eqref{Eq: quantile regression}, including asymptotic analysis \citep{li20081,kai2011new,wu2009variable,zou2008composite} and non-asymptotic analysis \citep{tan2022high}.

To solve Equation \eqref{Eq: quantile regression}, one can leverage a variety of machine learning methods to learn $\hat{q}_\alpha$ \citep{hunter2000quantile,taylor2000quantile,koenker2001quantile,meinshausen2006quantile,steinwart2011estimating}. Many existing studies assume a specific parametric relationship between $X$ and $Y$, such as linear or quadratic models.
For example, \cite{tan2022high} consider high-dimensional structured quantile regression. Specifically, they assume that, given the response variable $y_i$ and covariates $x_i$, the $\alpha^{th}$ conditional quantile function of $y_i$ is $q_{\alpha}(x_i) = x_i^\top \theta_\alpha^*$, where $\theta_\alpha^*$ is the parameter corresponding to quantile $\alpha$. 
For more information about machine learning methods for quantile regression, we refer readers to \cite{koenker2017handbook}.

However, a significant issue that arises from the assumption on the function class $\cF$ is the risk of model misspecification. Moreover, the regression can be biased because of the selection of the regularization term. For example, $f(x)$ may depend non-linearly on $x$, but the model may assume it is linear for simplicity and tractability. 
We use the following two examples for illustrating this model misspecification issue.

\begin{example}[Model Misspecification -- Uniform bias]
The true model is $q_{\alpha}(X)=|X^\top \theta|+\theta_0$, where $\theta_0>0$ and $\theta\in \mathbb{R}^d$; which is equivalent to assuming that $Y=|X^\top \theta|+\theta_0+\epsilon$, where $\Pr(\epsilon\leq 0|X)=\alpha$. In the training step, the function class is assumed to be $\cF=\{f: f(x)=|x^\top \theta|,\theta\in \mathbb{R}^d\}$, which excludes an intercept term. For each $X$, the predictive model may underestimate or overestimate the $\alpha$-quantile. When the bias term $\theta_0$ becomes large, substantial errors can arise, leading to arbitrarily poor decisions.
\QEG
\end{example}

\begin{example}[Model Misspecification -- Non-uniform bias]
Consider the piece-wise linear model: $q_\alpha(X)=2|X|+2$ if $0<X\leq 4$, and $q_\alpha(X)=4|X|+2$ if $-4\leq X\leq 0$. When the model is misspecified to the linear model (e.g., $q_\alpha(X)=3|X|+2$), then all quantiles for $X<0$ are overestimated and all quantiles for $X>0$ are underestimated. The decision could result in poor performance when the prediction relies on an incorrect model.
\QEG
\end{example}

To tackle the issue of model misspecification, we draw inspiration from conformal prediction. Before presenting our algorithm, we first provide an overview of the conformal prediction framework.

\subsection{Preliminary: Conformal Prediction}
We first describe how conformal prediction \citep{shafer2008tutorial} constructs the prediction intervals that have statistical guarantees. The split conformal method splits the training data into two disjoint subsets: a training set $\cI_1$ and a calibration set $\cI_2$. 
We outline the procedure for conformal prediction:
\begin{enumerate}[label=\arabic*.]
  \item Train the model $\hat{f}\in \cF$ on the training set $\cI_1$; the function class $\cF$ can be either parametric or nonparametric.
  \item Define the score function  $s(x,y;\hat{f})\in \mathbb{R}$ based on the predictor trained from Step 1. For any data point $(X_i,Y_i)\in \cI_2$, compute the calibration score $s_i=s(X_i,Y_i;\hat{f})$. (A larger score indicates worse agreement between $\hat{f}(x)$ and $y$).
  \item Compute $\hat{s}$ as the $\frac{ (n+1)(1-\alpha)}{n}$-quantile of the calibration scores $s_1,\cdots, s_n$.
  \item Use this quantile to form the prediction sets for new contextual information:
  \[
\cC(X_{n+1})= \{y: s(X_{n+1},y;\hat{f})\leq \hat{s}\}.
  \]
\end{enumerate}

The key step in the conformal prediction is to find a good score function according to the task.
Let us consider a specific score function.
Given any algorithm $\cA$, a predictor is trained from the training set:
\[
\hat{f}(x) \leftarrow \cA(\{(X_i,Y_i):i\in \cI_1\}).
\]
After training, the absolute residuals are computed on the calibration set $\cI_2$, as follows:
\[
s_i= |Y_i-\hat{f}(X_i)|, \quad \forall i\in \cI_2.
\]
Let $\bs=\{s_i,i\in \cI_2\}$. Then, we compute a quantile of the empirical distribution of the absolute residuals:
\begin{equation}\label{eq: construct Q}
Q_{1-\alpha}(\bs, \cI_2): = (1-\alpha)(1+1/|\cI_2|)-\text{the empirical quantile of } \{s_i|i\in \cI_2\}.
\end{equation}
The prediction interval at a new point $X_{n+1}$ is given by 
\begin{equation}\label{eq: construct Q2}
\cC(X_{n+1}) = [\hat{f}(X_{n+1})-Q_{1-\alpha}(\bs, \cI_2), \hat{f}(X_{n+1})+Q_{1-\alpha}(\bs, \cI_2)]\,,
\end{equation}
where the interval is guaranteed to satisfy that
\begin{equation}\label{eq: CI lower bound}
\Pr(Y_{n+1}\in \cC(X_{n+1}))\geq 1-\alpha.
\end{equation}

One significant limitation of this method is that the length of $\cC(X_{n+1})$ is fixed at $2Q_{1-\alpha}(\bs,\cI_2)$, regardless of $X_{n+1}$. This uniformity overlooks the fact that observational noise may vary across features; some may exhibit low noise and be easier to predict, while others may have high noise. As a result, a uniform-length interval can be overly conservative for certain features. The naive bound $\hat{\mu}(X_{n+1}) + Q_{1-\alpha}(\bs, \cI_2)$ fails to capture the correct quantile because the true confidence interval may vary in length, depending on the feature. Even more critical is that, as with all conformal inferences, the lower bound in Inequality \eqref{eq: CI lower bound} is feature-independent. Therefore, we cannot guarantee that $\Pr(Y_{n+1} \in \cC(X_{n+1}) \mid X_{n+1}) \geq 1-\alpha$.

To address the issue of heteroscedasticity and to construct variable-width conformal prediction intervals, \cite{romano2019conformalized} proposed conformalized quantile regression. Specifically, they fit two conditional quantile functions, $\hat{q}_{\alpha_{\text{lo}}}$ and $\hat{q}_{\alpha_{\text{hi}}}$, on the training dataset $\cI_1$. Then, they computed conformity scores, 
\[
s_i = \max\{\hat{q}_{\alpha_{\text{lo}}}-Y_i, Y_i-\hat{q}_{\alpha_{\text{hi}}}\},
\]
for each $i\in \cI_2$. The conformity score $s_i$ is the magnitude of the error incurred by the mistake. By constructing confidence interval (CI) through Equations \eqref{eq: construct Q} and \eqref{eq: construct Q2} using the new conformity score, the same level of confidence \eqref{eq: CI lower bound} can be guaranteed. However, the statistical guarantee is still unconditional.

A key question is whether conformal prediction can be effectively leveraged to enhance decision making in data-driven newsvendor problems. Unlike traditional applications of conformal inference, which primarily focus on constructing prediction intervals for the response variable, our goal is to improve quantile prediction to support better decision making. We aim to achieve this goal by integrating the strengths of conformal prediction into the newsvendor framework.

Conformal prediction enables the construction of confidence intervals using any machine learning model, including black-box models, while providing a user-friendly framework for generating statistically rigorous uncertainty sets around predictions. Our proposed framework preserves these key advantages: It is model-free and distribution-free; and it can accommodate any machine learning method for training while also offering theoretical performance guarantees under regularity conditions.

\section{Contextual Quantile Prediction with Calibration}\label{sec: CCQP}
In this section, we introduce Contextual Quantile Prediction with Calibration and provide statistical guarantees under regularity conditions. In particular, we provide performance guarantees of the conformalized quantile and the confidence interval of the true quantile.

\subsection{Algorithm}
The contextual quantile predict process, with calibration, involves three steps. First, we split the data into a proper training set, indexed by $\cI_1$, and a calibration set, indexed by $\cI_2$. Given any quantile regression algorithm $\cA$, we fit conditional quantile function $\hat{q}_{\alpha}$:
\[
\hat{q}_\alpha \leftarrow \cA(\{(X_i, Y_i): i\in \cI_1\}).
\]
Second, we compute conformity scores that quantify the error:
\[
s_i = Y_i-\hat{q}_{\alpha}(X_i)\,, \text{ for }i\in \cI_2.
\]
The score is positive if $Y_i$ is above the estimated quantile and negative if $Y_i$ is below the estimated quantile. Third, given new input data $X_0$, we construct the $\alpha$ quantile as 
\[
\hat{q}_\alpha^c(X_0):= \hat{q}_\alpha(X_0) + Q_{\alpha}(\bs, \cI_2)\,,
\]
where the superscript $c$ denotes ``conformalized'' and where the plug-in prediction interval is conformalized by
\begin{equation}\label{Q score function}
Q_{\alpha}(\bs, \cI_2) = \alpha(1+1/|\cI_2|)\text{-th empirical quantile of }\{s_i, i\in \cI_2\}.
\end{equation}
The pseudo-code is shown in Algorithm \ref{Alg: CCQP}, which we call \emph{Contextual Quantile Prediction with Calibration} (\textsf{CQPC}). The implementation is simple: First we train the model in the training phase, and then we compute the conformity scores during the calibration phase and use them to construct the 
$\alpha$-quantile.

Our intuitive explanation is that if $\hat{q}_\alpha(X_0)$ accurately captures the $\alpha$-quantile, then approximately $\alpha$ proportion of the $s_i$ values will be negative, while $1-\alpha$ proportion will be positive. As a result, $Q_\alpha(\bs, \cI_2)$ will be close to zero. Conversely, if $\hat{q}_\alpha(X_0)$ overestimates the quantile, more than $\alpha$ proportion of the $s_i$ values will be negative, making $Q_\alpha(\bs, \cI_2)$ negative. This result implies a need to reduce $\hat{q}_\alpha(X_0)$ by a bias term. Similarly, if $\hat{q}_\alpha(X_0)$ underestimates the quantile, $Q_\alpha(\bs, \cI_2)$ will be positive, indicating that a bias should be added to correct the underestimation.

\begin{algorithm}
   \caption{\textnormal{\textsf{Contextual Quantile Prediction with Calibration}}}\label{Alg: CCQP}
\begin{algorithmic}[1]
\State Split the data to the training dataset $\cI_1$ and calibration dataset $\cI_2$
\State Apply quantile regression algorithm $\cA$ on the training dataset $\cI$ to get the predictor $\hat{q}_\alpha(\cdot)$
\State Compute the conformity score $s_i = Y_i-\hat{q}_{\alpha}(X_i)\,, \text{ for }i\in \cI_2$
\State Compute $Q_\alpha(\bs, \cI_2)$ according to Equation \eqref{Q score function}
\For{any new data point $X_i$}
\State  $\hat{q}_\alpha^c(X_i):= \hat{q}_\alpha(X_i) + Q_{\alpha}(\bs, \cI_2)$
\EndFor
\end{algorithmic}
\end{algorithm}

\subsection{Statistical Guarantees}

In this section, we provide theoretical guarantees for the \CQPC. First, we present unconditional statistical guarantees in Theorem \ref{theorem: quantile}. Next, we introduce two new concepts -- the gap function of prediction error and the margin condition -- to establish conditional statistical guarantees. Moreover, we characterize the confidence interval for the quantile.

For ease of notation, let $n_1 = |\cI_1|$; let $n_2= |\cI_2|$; and let $n=n_1+n_2$. We first prove that the conformalized predicted quantile satisfies the statistical guarantee in Theorem \ref{theorem: quantile}.

\begin{theorem}\label{theorem: quantile}
If $(X_i, Y_i), i=1,\cdots, n+1$ are i.i.d. and almost surely distinct, then 
\[
\alpha\leq \Pr(Y_{n+1}\leq \hat{q}_\alpha^c(X_{n+1}))\leq \alpha + \frac{1}{n_2+1}.
\]
\end{theorem}

Theorem \ref{theorem: quantile} establishes that the probability that $Y_{n+1}$ is less than $\hat{q}_\alpha^c(X_{n+1})$ falls within the interval $\left[\alpha, \alpha+\frac{1}{n_2+1}\right]$. This result implies that $\hat{q}_\alpha^c(X_{n+1})$ lies between the $\alpha$-quantile and the $\alpha+\frac{1}{n_2+1}$-quantile. However, this probability is unconditional, meaning it does not account for any specific contextual information. In practice, we often are interested in the conditional quantile, which considers contextual information that typically is observable.

To provide statistical guarantees for the conditional quantile with contextual information, we define a function, the gap function of prediction error, that links the estimation error in quantile prediction across different contexts.

\begin{assumption}[Gap Function of Prediction Error]\label{assum: prediction model}
There exists a function $\kappa$ such that, given $n_1$ data points in $\cI_1$, the following holds:
\[
|(\hat{q}_{n_1,\alpha}(x_1)-q_{\alpha}^*(x_1))-(\hat{q}_{n_1,\alpha}(x_2)-q_{\alpha}^*(x_2))|\leq \kappa(n_1, \xi(x_1, x_2)), \quad \forall x_1, x_2\in \cX\,,
\]
where $\xi(x_1,x_2)$ is the distance between contexts $x_1$ and $x_2$. This function $\kappa$ is the gap function of prediction error.
\end{assumption}

The gap in the prediction error is described by the function $\kappa(n_1, \xi(x_1, x_2))$, which depends on both the size of the training data and the distance between contexts. Intuitively, $\kappa$ decreases as $n_1$ increases, and it increases as $\xi(x_1, x_2)$ increases. The reason is that a larger training set generally enhances prediction accuracy, and closer contexts are likely to result in similar prediction errors. Note that the existence of gap function $\kappa$ does not assume model correctness. In other words, the model can be significantly inaccurate, but the bias in prediction for each context is bounded by the function of the distance between contexts. The model may overestimate in some regions and underestimate in others.

We lay out a possible structure of $\kappa$. \cite{pan2021multiplier} study the linear model where $f(X)=X^\top \theta$. In Theorem 2.1, they show that, under regularity conditions, the quantile regression estimator satisfies that
\[
\|\htheta-\theta^*\|_{\Sigma}\leq C\sqrt{\frac{d+t}{n}}
\]
with probability at least $1-2e^{-t}$ when $n$ is large enough, where $\Sigma = \E[XX^\top]$ and $C$ is some constant. This estimation error is derived under the assumption that $\theta^*$ is consistent across different contexts. However, in practice, the true model may vary from one context to another, with closer contexts likely sharing more similar models. Therefore, we can reasonably assume that 
\begin{equation}\label{eq: kappa structure}
\kappa(n_1, \xi(x_1, x_2))=C\sqrt{\frac{\dis^\nu}{n_1}},
\end{equation}
 where $\nu\geq 0$ is a constant. In the special case that all contexts share the same model, $\nu=0$. For $\nu>0$, the difference in the prediction error between different contexts widens as the contexts become more distinct.

Another assumption is on the margin condition. From the definition, the $\alpha$-quantile satisfies that $\Pr(Y\leq q_{\alpha}^*(X)|X)\leq \alpha$. In the neighborhood of the $\alpha$-quantile, we assume a uniform margin condition.

\begin{assumption}[Margin Condition]\label{assum: margin}
For all $X\in \cX$ and any $\Delta>0$, there exist functions $h_-(\cdot)$ and $h_+(\cdot)$ such that
\[
 \alpha + \lowh(\Delta)\leq \Pr(Y\leq q_{\alpha}^*(X)+\Delta|X)\leq \alpha + \barh(\Delta)
 \]
 and 
 \[
\alpha- \barh(\Delta)\leq \Pr(Y\leq q_{\alpha}^*(X)-\Delta|X) \leq \alpha- \lowh(\Delta).
\]
\end{assumption}
The functions $\lowh(\Delta)$ and $\barh(\Delta)$, which together we call the \emph{margin function}, represent how the density function accumulates around the $\Delta$-neighborhood of the $\alpha$-quantile. If the density function remains relatively low,
the bounds are small; if the density is high, the bounds are larger. Note that $\lowh(\Delta)$ and $\barh(\Delta)$ are both monotonically increasing functions. In particular, $\lowh(\Delta)=\barh(\Delta)=0$ when $\Delta=0$. The following three examples, using uniform distribution, Gaussian distribution, and exponential distribution, illustrate the shape of $h$.

\begin{example}[Uniform Distribution]\label{example: uniform}
Given context $X$, suppose $Y$ is uniformly distributed in $[-|X^\top \theta^*|, |X^\top \theta^*|]$, where $0<\lgamma(\cX) \leq |X^\top \theta^*|\leq \bgamma(\cX)<\infty$. Then, $q_\alpha^*(X) = (1-2\alpha)|X^\top \theta^*|$ and 
\[
\Pr(Y\leq q_{\alpha}^*(X)+\Delta|X)=\alpha +\frac{\Delta}{2|X^\top \theta^*|}\in\left[\alpha+\frac{\Delta}{2\bgamma(\cX)}, \alpha+\frac{\Delta}{2\lgamma(\cX)}\right]\,,
\]
and
\[
\Pr(Y\leq q_{\alpha}^*(X)-\Delta|X)=\alpha -\frac{\Delta}{2|X^\top \theta^*|}\in \left[\alpha-\frac{\Delta}{2\lgamma(\cX)}, \alpha-\frac{\Delta}{2\bgamma(\cX)}\right]\,.
\]
In this case, $\barh(\Delta)=\frac{\Delta}{2\lgamma(\cX)}$ and $\lowh(\Delta)=\frac{\Delta}{2\bgamma(\cX)}$.
\QEG
\end{example}

\begin{example}[Gaussian Distribution]
Given context $X$, suppose $Y$ follows Gaussian distribution with mean $X^\top \theta^*$ and variance $(X^\top \theta^*)^2$. Then, $q_\alpha^*(X)=\Phi^{-1}(\alpha)|X^\top \theta^*|+X^\top \theta^*$ and
\[
\begin{aligned}
\Pr(Y\leq q_{\alpha}^*(X)+\Delta|X)&=\Pr(Y\leq \Phi^{-1}(\alpha)|X^\top \theta^*|+X^\top \theta^*+\Delta|X)\\
&= \Pr\left(Y\leq \left(\Phi^{-1}(\alpha)+\frac{\Delta}{|X^\top \theta^*|}\right)|X^\top \theta^*|+X^\top \theta^*|X\right)\\
&=\Phi\left(\Phi^{-1}(\alpha)+\frac{\Delta}{|X^\top \theta^*|}\right)\,.
\end{aligned}
\]
Thus, $\barh(\Delta)=\Phi\left(\Phi^{-1}(\alpha)+\frac{\Delta}{\lgamma(\cX)}\right)$ and $\lowh(\Delta)=\Phi\left(\Phi^{-1}(\alpha)+\frac{\Delta}{\bgamma(\cX)}\right)$.
\QEG
\end{example}

\begin{example}[Exponential Distribution]\label{example: exponential}
Given context $X$, suppose $X$ follows exponential distribution with parameter $|X^\top \theta^*|$. Then, $q_\alpha^*(X) = \frac{\log(1-\alpha)}{-|X^\top \theta^*|}$ and
\[
\begin{aligned}
\Pr(Y\leq q_{\alpha}^*(X)+\Delta|X)&=\Pr\left(Y\leq  \frac{\log(1-\alpha)}{-|X^\top \theta^*|}+\Delta|X\right)\\
&= 1-\exp\left(-|X^\top \theta^*|\cdot \left(\frac{\log(1-\alpha)}{-|X^\top \theta^*|}+\Delta\right)\right)\\
&=1-(1-\alpha)\cdot \exp(-|X^\top \theta^*|\Delta)\\
&=\alpha + (1-\alpha)\cdot (1-\exp(-|X^\top \theta^*|\Delta))\\
&\in \left[\alpha +(1-\alpha)(1-\exp(-\lgamma(\cX)\Delta),\alpha +(1-\alpha)(1-\exp(-\ugamma(\cX)\Delta)\right].
\end{aligned}
\]
Therefore, $\barh(\Delta)=\alpha +(1-\alpha)(1-\exp(-\ugamma(\cX)\Delta))$ and $\lowh(\Delta)=\alpha +(1-\alpha)(1-\exp(-\lgamma(\cX)\Delta))$.
\QEG
\end{example}

If the prediction model exhibits a constant bias across all $x \in \cX$ (i.e., when $\kappa$=0), we obtain a nice result in Theorem \ref{theorem.kappa0}, even if the model itself is incorrect. For instance, the result applies when the true model is $q_\alpha^*(X)=X^\top \theta^* + \theta_0$ where $\theta_0\neq 0$ but the prediction model is $\hat{q}_\alpha^*(X)=X^\top \theta^*$.

\begin{theorem}\label{theorem.kappa0}
Suppose Assumption \ref{assum: prediction model} holds with $\kappa=0$ and Assumption \ref{assum: margin} holds. Then for any $\Delta>0$, it holds that
\[
\alpha-\barh(\Delta)-\exp\left(-2n_2 \lowh(\Delta)^2\right)\leq \Pr(Y_{n+1}\leq \hat{q}_\alpha^c(X_{n+1})|X_{n+1})\leq\alpha + \barh(\Delta)+\exp\left(-2 n_2 \lowh(\Delta)^2\right)\,
\]
for all $X_{n+1}$.
\end{theorem}

Theorem \ref{theorem.kappa0} shows that, when a constant bias exists across all $x\in \cX$, the conformalized quantile $\hat{q}_\alpha^c(X_{n+1})$ lies within the true quantile range given by $\big[q^*_{\alpha-\barh(\Delta)-\exp\left(-2 n_2 \lowh(\Delta)^2\right)}, q^*_{\alpha + \barh(\Delta)+\exp\left(-2 n_2 \lowh(\Delta)^2\right)}\big]$. For fixed $\Delta$, the bound shrinks with the calibration set size $n_2$, and the second term, $\exp\left(-2 n_2 \lowh(\Delta)^2\right)$, shrinks to 0 when $n_2$ goes to infinity. That is, the calibration quality improves with the calibration set size. The trade-off exists when selecting $\Delta$. While $\barh(\Delta)$ increases with $\Delta$, $\exp\left(-2 n_2 \lowh(\Delta)^2\right)$ decreases in $\Delta$. To minimize the gap, $\Delta$ should be chosen such that $\barh(\Delta)+\exp\left(-2 n_2 \lowh(\Delta)^2\right)$ is minimized.

Next, we provide a theorem for a more general case where $\kappa$ is not always equal to 0.

\begin{theorem}[Performance of Conformalized Quantile by Pooling Neighboring Data]\label{thm: heterogeneous}
Suppose Assumptions \ref{assum: prediction model} and \ref{assum: margin} hold. Given context $X_{n+1}$, when pooling all $n(\cB)$  data points in the neighboring ball $\cB$ of $X_{n+1}$ with diameter $\xi(\cB)$, then for any $\Delta>0$, it holds that
\[
\begin{aligned}
\alpha - \phi(\Delta, \cB)\leq &\Pr(Y_{n+1}\leq \hat{q}_\alpha^c(X_{n+1})|X_{n+1})
\leq \alpha + \phi(\Delta, \cB), 
\end{aligned}
\]
where 
\[
\phi(\Delta, \cB) = \barh(\Delta+\kappa(n_1(\cB),\dis(\cB)))+\exp\left(-2 n_2(\cB)\lowh(\Delta)^2\right).
\]
\end{theorem}

Theorem \ref{thm: heterogeneous} establishes the result for the general case, where the bias varies across different contexts. Several factors influence the resulting bounds. First, the bound depends on the data pooling region $\cB$, which affects both the size of the training dataset and the maximum distance between contexts. As the data pooling region $\cB$ expands, both $n_1(\cB)$ and $\xi(\cB)$ increase. However, because $\kappa$ decreases with $n_1$ and increases with $\dis(\cB)$, a trade-off exists when selecting the region $\cB$ to minimize $\kappa$. Moreover, $\cB$ affects the second term, $\exp\left(-2 n_2(\cB)\lowh(\Delta)^2\right)$, in that the calibration dataset size $n_2(\cB)$ increases with $\cB$. Therefore, for any fixed $\Delta$, we can minimize the gap between the lower and upper bounds by choosing $\cB$ to minimize $\phi(\Delta, \cB)$.

Second, the bound depends on the choice of $\Delta$. Because the bound holds for any $\Delta > 0$, tightening the bound involves selecting $\Delta$ to minimize $\phi(\Delta, \cB)$. As in Theorem \ref{theorem.kappa0}, a trade-off exists for $\Delta$ because the first term, $\barh$, increases with $\Delta$, while the second term $\exp\left(-2 n_2(\cB)\lowh(\Delta)^2\right)$ decreases with $\Delta$.

In the special scenario where $\cB=\cX$, we can immediately obtain the bound: 
\[
\alpha - \phi(\Delta, \cX)\leq \Pr(Y_{n+1}\leq \hat{q}_\alpha^c(X_{n+1})|X_{n+1})
\leq \alpha + \phi(\Delta, \cX).
\]
In this case, we can simply select $\tilde{\Delta}$ such that 
\begin{equation}\label{eq: tilde Delta}
\barh(\tilde{\Delta}+\kappa(n_1(\cB),\dis(\cB)))=\exp\left(-2 n_2(\cB)\lowh(\tilde{\Delta})^2\right).
\end{equation}
Such $\tilde{\Delta}$ satisfies that $\phi(\tilde{\Delta},\cX)\leq 2\phi(\Delta^*,\cX)$, where $\Delta^*$ is the optimal $\Delta$ that minimizes $\phi(\Delta,\cX)$. We prove this conclusion in Corollary \ref{Corollary: tilde Delta}.

\begin{corollary}\label{Corollary: tilde Delta}
Suppose Assumptions \ref{assum: prediction model} and \ref{assum: margin} hold. Given context $X_{n+1}$, it holds that 
\[
\begin{aligned}
\alpha-2\phi(\Delta^*, \cX)\leq \alpha - \phi(\tilde{\Delta}, \cX)\leq &\Pr(Y_{n+1}\leq \hat{q}_\alpha^c(X_{n+1})|X_{n+1})
\leq \alpha + \phi(\tilde{\Delta}, \cX)\leq \alpha+2\phi(\Delta^*, \cX),
\end{aligned}
\]
where $\tilde{\Delta}$ satisfies Equation \eqref{eq: tilde Delta} and $\Delta^*$ minimizes $\phi(\Delta,\cX)$.
\end{corollary}

Compared to solving for the optimal $\Delta^*$, computing $\tilde{\Delta}$ is significantly easier. The reason is that this computation allows for the application of the bisection algorithm because the left-hand side of \eqref{eq: tilde Delta} monotonically increases in $\tilde{\Delta}$ while the right-hand side of \eqref{eq: tilde Delta} decreases in $\tilde{\Delta}$. In addition, $\tilde{\Delta}$ comes with a nice theoretical guarantee, as outlined in Corollary \ref{Corollary: tilde Delta}.

Theorem \ref{thm: heterogeneous} and Corollary \ref{Corollary: tilde Delta} establish a performance guarantee for the conformalized quantile, while the following theorem provides a confidence interval for the quantile.

\begin{theorem}[Confidence Interval of the Quantile]\label{Corollary: interval}
Fix $\delta>0$. Given context $X_{n+1}$, when pooling all $n(\cB)$ data points in the neighboring ball $\cB$ of $X_{n+1}$ with diameter $\xi(\cB)$, it holds that
\[
\Pr( \hat{q}_{\alpha-z}^c(X_{n+1})\leq q_\alpha^*(X_{n+1})\leq \hat{q}_{\alpha+z}^c(X_{n+1})|X_{n+1})\geq 1-\delta,
\]
where 
$z=\barh\left(\lowh^{-1}\left(\sqrt{\frac{\log(1/(2\delta))}{2n_2(\cB)}}\right)+\kappa (n_1(\cB), \xi(\cB)\right)$.
\end{theorem}

The confidence interval for the quantile $q_\alpha^*(X_{n+1})$, as characterized by Theorem \ref{Corollary: interval}, depends on both the gap function $\kappa$ and the margin functions $\barh$ and $\lowh$. As more data are collected, the confidence interval shrinks because $z$ decreases with the size of the training and calibration datasets. Specifically, if $\kappa$ converges to zero as $n_1$ increases, the interval will shrink to zero as the sample size approaches infinity. Furthermore, Theorem \ref{Corollary: interval} offers additional practical benefits. By characterizing the confidence interval for the quantile, it enables the decision maker to optimize their decisions within the uncertainty set defined by Theorem \ref{Corollary: interval}, allowing for decisions to be made either optimistically or robustly (max/min problem), depending on the context. However, we want to point out that the gap function and the margin functions could be unknown in practice. In Appendix \ref{Appendix: data-driven}, we develop a data-driven method to estimate these two types of functions.

In implementing the conformalized quantile prediction, the key question is how to select the radius of the neighboring ball, which we call data pooling. In many real-world applications, historical data with more similar features may provide more information. Consider a two-dimensional feature space consisting of temperature and weather conditions. If a decision maker aims to predict ice cream demand for a sunny day with a temperature of 80°F, focusing on similar conditions may be more effective. For example, a rainy day with a temperature below 60°F is likely to offer limited relevant information for this prediction. Similarly, in a meal delivery system, demand is highly influenced by factors such as order time, delivery location, and restaurant cuisine type. Pooling data with similar times, locations, and cuisine preferences can potentially enhance the predictive power of the model.

There is a trade-off between data quality and data quantity. Pooling data selectively results in a smaller dataset, while choosing a larger region and including dissimilar data increases the dataset size but may compromise quality. In the next section, we introduce a framework that balances data quality and data quantity.

\section{Quality vs. Quantity of Data} \label{sec: quality vs quantity}
In this section, we discuss the properties of the optimal pooling region, focusing on how it is influenced by the dataset size, the gap function, and the margin functions. In addition, we present a practical, data-driven framework for selecting the pooling region.

\subsection{Data Pooling}\label{subsec: data pooling}

Data pooling involves a trade-off between data quality and quantity. The key challenge is this: For a given context $X_{n+1}$, how can one efficiently choose the pooling region to get closest to the true quantile?
Theorem \ref{thm: heterogeneous} implies that, to minimize the gap between upper and lower bounds, we choose $\Delta$ and $\cB$ to minimize $\phi(\Delta, \cB)$. 
However, optimizing $\phi(\Delta, \cB)$ is not an easy task because the hardness of the optimization problem depends heavily on the structure of $\kappa$.

To simplify and make the problem tractable, we assume that the pooling region is a neighboring ball. In this case, selecting $\cB$ is equivalent to choosing its diameter $\xi$, so we use $\xi(\cB)$ to denote the diameter of ball $\cB$, and we use $\cB_\xi$ to represent the ball with dimeter $\xi$. There is a one-to-one mapping between the diameter and the ball. 

The closed-form solution of the optimal $\xi$ can be derived when $\barh$, $\lowh$, and $\kappa$ have nice structures. We first characterize the optimal pooling region when both $\barh$ and $\lowh$ are linear (see Example \ref{example: uniform}) and when $\kappa$ has a structure that follows Formula \eqref{eq: kappa structure}.

\begin{proposition}\label{corollary: optimal d}
Suppose $\barh(\Delta)=c_1 \Delta$ and $\lowh(\Delta)=c_2 \Delta$, where $c_1\geq c_2$; $n_1(\cB_{\dis})=\rho n {\dis}^\iota$ and $n_2(\cB_{\dis})=(1-\rho)n {\dis}^\iota$; and $\kappa(n_1,{\dis})=\sqrt{\frac{{\dis}^\nu}{n_1}}$, where $\iota$ and $\nu$ are two constants. Then, 
\begin{enumerate}
\item When $\nu>\iota$, the optimal pooling diameter satisfies that
\[
\sqrt{n {\dis}^{\nu+\iota}}\exp(-c_2' {\dis}^{\nu})=c_1',
\]
where $c_1'=\frac{c_1\iota \sqrt{\rho}}{4(\nu-\iota)(1-\rho)c_2^2}$ and $c_2'=2 c_2^2\frac{(\nu-\iota)^2}{\iota^2} \frac{1-\rho}{\rho}$. \label{optimal d1}
\item When $\nu\leq \iota$, pooling data in the entire region is optimal. \label{optimal d2}
\end{enumerate}
\end{proposition}

Proposition \ref{corollary: optimal d} provides a specific formula for 
$\dis$'s being optimal under varying conditions. When $\nu$ is small, meaning the differences between contexts are small, the optimal pooling region may extend to the entire dataset. In this case, as the pooling region expands, the advantage of having more data for training outweighs the losses caused by model differences. However, when $\nu$ is large and the true model varies significantly between contexts, balancing the size of the training set against the impact of model discrepancies is crucial.
For example, for a grocery store, if demand fluctuates significantly between neighboring locations (i.e., $\nu$ is large), using a smaller pooling region is preferable to capture local variations. Conversely, if demand patterns are similar across a broader area (i.e., $\nu$ is small), a larger pooling region is more effective for leveraging consistent trends.

While Proposition \ref{corollary: optimal d} offers the precise form of the optimal pooling diameter, a closed-form solution does not exist for the general case. A straightforward approach to determining the optimal pooling diameter is to enumerate $\Delta$ and $\dis$ by discretizing the continuous space of these two parameters. Suppose the search spaces for the two parameters are $\cS_\Delta$ and $\cS_\xi$, respectively; then the computational complexity is given by $|\cS_\Delta||\cS_\xi|$.

This naive way can be further improved: To do so, we propose a two-approximation solution.\footnote{We follow the convention that the approximation ratio is no less than 1 for minimization problems.} As we discussed, to ease the computation, one can choose $\tilde{\Delta}(\cB_\dis)$ that satisfies Equation \eqref{eq: tilde Delta}. To solve \eqref{eq: tilde Delta}, we can use the bisection algorithm, because the left-hand side increases in $\tDelta$ and the right-hand side decreases in $\tDelta$. In this case, the complexity would decrease from linear to logarithmic for searching $\Delta$ (i.e., the computational complexity is $|\cS_\xi|\log(|\cS_\Delta|)$).
Suppose $\cB_{\tdis}$ is a neighboring ball that minimizes $\phi(\tilde{\Delta}(\cB_{\tdis}), \cB_{\tdis})$. Then, we show the performance guarantee of $\cB_{\tdis}$ in Proposition \ref{prop: tilde cB}, where $\Delta^*$ and $\cB_{\dis^*}$ is the minimizer of $\phi(\Delta,\cB)$.

\begin{proposition}\label{prop: tilde cB}
$\phi(\tilde{\Delta}(\cB_{\tdis}), \cB_{\tdis})$ provides a two-approximation solution. That is,
\[
\phi(\tilde{\Delta}(\cB_{\tdis}), \cB_{\tdis})\leq 2\phi(\Delta^*,\cB^*).
\]
\end{proposition}

Although the complexity of searching for $\tDelta$ only requires a logarithm order, the complexity of finding the diameter is still linear. The hardness of the optimization problem depends on the structure of $\kappa$. Note that $\kappa$ decreases in $n_1$ and increases in $\xi$. Because $n_1$ increases in the pooling diameter $\xi$, whether $\kappa$ would increase, decrease, or remain the same is unclear. Depending on the structure of $\kappa$, the data pooling strategy changes. We characterize the strategy in Theorem \ref{Prop: choice of d} under general structures of the gap function.

\begin{theorem}\label{Prop: choice of d}
Suppose $n_1(\cB_{\dis}) = \rho n(\cB_{\dis})$ and $n_2(\cB_{\dis}) = (1-\rho) n(\cB_{\dis})$. The following properties hold: 
\begin{enumerate}
  \item If $\kappa(n_1(\cB_{\dis}),{\dis})$ is independent of or monotonically decreasing with ${\dis}$, then pooling all the data is optimal. \label{Prop: choice of d1}
  \item If $\kappa(n_1(\cB_{\dis}),{\dis})$ monotonically increases with ${\dis}$, then the optimal choice of ${\dis}$ could expand, shrink, or remain the same when more data are collected. \label{Prop: choice of d2}
\end{enumerate}
\end{theorem}

Theorem \ref{Prop: choice of d} implies that scenarios where $\kappa(n_1(\cB_{\dis}),{\dis})$ increases monotonically with ${\dis}$ are more complex. As the dataset grows (i.e., as more data are collected), the optimal pooling region may either expand, contract, or remain unchanged, depending on the structure of the function $\kappa$. To demonstrate this relationship, we provide three examples in the proof, each corresponding to a distinct scenario. Although the change in the pooling region is uncertain as the number of data points increases, we know that in the large-sample regime, as the sample size approaches infinity, the optimal pooling diameter converges to zero under certain conditions.

\begin{proposition}[Big-data regime]\label{prop: big data}
Suppose that continuous functions $\kappa(\infty, x)$, $\barh(x)$, and $\lowh(x)>0$ strictly increase in $x>0$. Then, the optimal pooling diameter converges to zero when $n$ goes to infinity.
\end{proposition}

This result is intuitive. When sample size is not a limiting factor, the optimal strategy is to pool data that are as pure as possible. In the case of discrete contexts, if the dataset is sufficiently large, the best approach is to always pool data from identical contexts.

So far, we have focused on identifying the optimal pooling region under the assumption that the gap function $\kappa$ and the margin functions $\barh$ and $\lowh$ have known structures. To summarize, when these functions exhibit well-behaved structures, deriving an explicit solution for the optimal pooling diameter is possible. In cases where the functions have more general structures, we can solve the problem either by enumerating in a discrete space, which has a computational complexity of $O(|\cS_\Delta||\cS_\xi|)$, or by using a two-approximation solution with a reduced complexity of $O(|\cS_\xi|\log(|\cS_\Delta|))$. In addition, we have characterized how the structure of the gap function generally influences the optimal pooling region. Furthermore, in the big-data regime, aligning with intuition, the optimal pooling diameter shrinks to zero as the data size approaches infinity.

In many practical scenarios, the structure of these functions may be unknown. This situation raises the question: Given a dataset and no prior knowledge, how can we empirically select the pooling diameter for implementing \CQPC? To address this question, we propose a data-driven framework designed specifically for practical implementation.

\subsection{Data-Driven Approach}\label{subsec: data-driven}

We propose three data-driven approaches for empirically selecting the optimal pooling region for a given context $X_{n+1}$. Two of these methods do not leverage the structure of the gap or margin functions, while the third approach estimates these functions to determine the optimal pooling region according to the theory we developed in Section \ref{subsec: data pooling}.

\subsubsection{Selecting $\dis$ from the training dataset.}\label{subsec: select parameter} The pooling diameter, denoted as $\dis$, determines the size of the dataset used to predict the quantile for $X_0$. To evaluate the effectiveness of a given $\dis$, we employ a $K$-fold cross-validation procedure. Specifically, we split the data in $\cB$ into $K$ folds. The prediction model is trained on $K-1$ of these folds, and the remaining fold is used for validation. This process is repeated $K$ times, with each fold serving as the validation set once.

Within each training dataset (comprising $K-1$ folds), we further divide the data into two subsets: $\cI_1$ for training the model and $\cI_2$ for calibration. After obtaining the conformal quantile prediction $\hat{q}_\alpha^c(\cdot)$ using Algorithm \ref{Alg: CCQP}, we evaluate its performance by computing the pinball loss on the validation dataset.
This procedure is repeated for different values of $\dis$, and the pooling diameter that results in the lowest loss is selected as the optimal value.

Although this method is straightforward and easy to implement, it can be computationally expensive. First, the model must be retrained for each value of $\dis$, and because training typically is resource-intensive, this retraining process can significantly increase computational costs. In addition, as more data are sequentially collected, the model requires further retraining. According to Theorem~\ref{Prop: choice of d}, the optimal pooling diameter is sensitive to the sample size, meaning that whenever new data are added, the optimal diameter must be recalculated, further compounding the computational burden.

\subsubsection{Global Training with Local Calibration}\label{subsec: global local}

As we noted in Section \ref{subsec: select parameter}, treating $\xi$ as a hyper parameter is computationally expensive because the training process needs to be repeated for each $\xi$. To ease the computational burden, we propose a framework that does not require retraining in the process of selecting the optimal pooling diameter.

Up to this point, we have focused on the procedure of training and calibration using data pooled within the same ball. One big benefit of the conformal approach is that the training process and the calibration process can be decoupled. Therefore, a more general approach can pool data from different balls for training and calibration. 
We propose a framework that does training on the entire region and applies local calibration using data from the neighboring ball. We call this framework \emph{Global Training with Local Calibration} (\gtlc).

Following the procedure outlined in Section \ref{subsec: select parameter}, we can divide the data in $\cI_2$ into $K$ folds, represented by $\cI_2^{(1)},\cdots, \cI_2^{(K)}$. Let the predictor trained on $\cI_1$ be denoted as $\hat{q}(\cdot)$. The calibration process is performed on $K-1$ of these folds, while the remaining fold is reserved for validation. For a fixed diameter $\xi$, and for any point 
$(x,y)$ in the validation set, we identify all data points within the $\xi$-neighboring ball from the $K-1$ calibration folds, denoted as $\cI_2\backslash \cI_2^{(K)}\cap \cB_\xi$. Calibration is applied to this neighboring ball, and the prediction is adjusted to $\hat{q}^c(x)$. We then compute the pinball loss, $\ell_\alpha(\hat{q}^c(x),y)$, for that validation point. This process is repeated for all points in the validation set. The optimal value of $\xi$ that minimizes the pinball loss is selected.

Rather than limiting the calibration selection to the neighboring ball, we include flexibility to allow for other forms of pooling. For instance, 
 we also can choose a certain number of nearby data points for calibration, where the distance between features can be defined using any norm. Specifically, let $m$ represent the number of neighboring points used for calibration, which serves as a hyperparameter. For any point 
$(x,y)$ in the validation set, we identify the $m$ closest data points to $x$ from the $K-1$ calibration folds, denoted as $\cB_m(x, \cI_2\backslash \cI_2^{(K)})$. We then apply calibration to these points and adjust the prediction to $\hat{q}^c(x)$. Next we compute the pinball loss, $\ell_\alpha(\hat{q}^c(x),y)$, for each validation point. This process is repeated for different values of $m$, and the optimal $m$ is the one that minimizes the pinball loss.

The pseudo code of \gtlc\ is provided in Algorithm \ref{Alg: GTLC}.

\begin{algorithm}
   \caption{\textnormal{\textsf{Global Traning with Local Calibration}}}\label{Alg: GTLC}
\begin{algorithmic}[1]
\State Split the data to the training dataset $\cI_1$ and calibration dataset $\cI_2$
\State Apply quantile regression algorithm $\cA$ on the training dataset $\cI_1$ to get the predictor $\hat{q}_\alpha(\cdot)$
\State Split $\cI_2$ to $K$ folds: $\cI_2^{(1)}, \cdots, \cI_2^{(K)}$
\For{different pooling diameter $\xi$ or number of nearby pooling points $m$}
\For{any data point $(X_i,Y_i)$ in the validation set $\cI_2^{(K)}$}
\State Compute the conformity score $s_i = Y_i-\hat{q}_{\alpha}(X_i)\,, \text{ for }(X_i,Y_i)\in \cI_2\backslash \cI_2^{(K)}\cap \cB_\xi$ or $(X_i, Y_i)\in\cB_m(x, \cI_2\backslash \cI_2^{(K)})$
\State Compute $Q_\alpha(\bs, \cI_2\backslash \cI_2^{(K)})$ according to Equation \eqref{Q score function}
\State  $\hat{q}_\alpha^c(X_i):= \hat{q}_\alpha(X_i) + Q_{\alpha}(\bs, \cI_2\backslash \cI_2^{(K)})$
\EndFor
\State Compute loss$=\frac{1}{|\cI_2^{(K)}|}\sum_{i} \ell_\alpha(\hat{q}^c(X_i),Y_i)$
\EndFor
\State Output: Select $\xi$ or $m$ from the lowest loss
\end{algorithmic}
\end{algorithm}

There are alternative methods for localization. For instance, given a context $x$, we can partition the surrounding region into concentric rings $\cO_1, \cdots, \cO_J$ with varying diameters centered at $x$. For each ring $\cO_j$, we compute a local correction term $Q_\alpha(\bs,\cI_2\cup \cO_j)$ for all $1\leq j\leq J$. The final correction term is then determined as a weighted combination of these local terms, with weights based on the distance from each ring to the context $x$. This approach ensures that all data points contribute to the conformalized quantile calculation, with greater weight assigned to points closer to $x$.

\subsubsection{Choosing optimal pooling region by estimating three functions}
The third approach determines the optimal pooling region by estimating three key functions: $\kappa$, $\barh$, and $\lowh$.

The general idea is as follows: We begin by selecting an initial pooling region, $\dis$, using a clustering algorithm, such as $K$-means clustering or spectral clustering. Then, we estimate the functions $\barh$, $\lowh$, and $\kappa$. Based on these estimates, we iteratively update the pooling region $\dis$ according to Equation \eqref{eq: tilde Delta}, repeating this process until satisfactory performance is achieved.

Details of this approach are provided in Appendix \ref{Appendix: data-driven}. We estimate $\barh$ and $\lowh$ across varying pooling region sizes and a sequence of margin levels, and we estimate $\kappa$ using data trained on different clustering diameters and sample sizes. The pseudo-code for this data-driven selection method for the pooling diameter is presented in Algorithm \ref{Alg: Pooling} in Appendix \ref{Appendix: data-driven}.

\section{Numerical Experiments}

In this section, we show the performance of our  algorithm on both the simulated dataset and the real dataset. Through numerical experiments, we show practical insights associated with our developed theory.
\subsection{Simulations}
We first conduct simulations to illustrate the performance of our proposed framework. As we discussed in Section \ref{subsec: data-driven}, multiple data-driven methods can be implemented. For ease of computation, we mainly implement the method discussed in Section \ref{subsec: global local}, which we call \emph{global training and local calibration}. 

\paragraph{Experiment Setup}: Similar to the setting considered in \cite{han2023deep}, we test two types of functions that encompass various feature dimensions and smoothness levels of the function:
\begin{enumerate}
  \item Multivariate Logistic (ML) model: 
  \[
f(x) = e^{x_1 - 0.5} + 2 (x_2 + x_3 - 1)^2 + |x_4 - 0.5| +e^{x_5 - 1} + 2 (x_6 + 3 x_7 - 1)^2 + |x_8 - 0.2| + x_9 ^2 +0.5  x_{10};
  \]
  \item Multivariate additive (MA) model:
\[
f(x) = \frac{\exp(\theta^\top x)}{1+\exp(\theta^\top x)},
\]
  where $\theta=(2, -4, 2, -1, 3, 5, -2,-1,0.5,2)^\top$.
\end{enumerate}
In all scenarios, each dimension of covariate vector $x$ is uniformly generated from [0,1]. The response (i.e., demand) is generated from $Y=f(X)+\epsilon$, where $\epsilon$ follows a standard normal distribution. We generate 2,000 data points, with 90\% of the data for training and 10\% for testing. For \CPRP, we subdivide the 90\%, taking 75\% for training the model and using 15\% for the second-stage calibration. For each test data point, we use the nearest 50 points in the calibration set to calibrate. The distance is measured in Euclidean distance. Each experiment is repeated 100 times.

\paragraph{Tested Algorithms}:
\begin{enumerate}
   \item Linear Quantile Regression (LQR): LQR estimates the relationship between independent variables and a specified quantile of the dependent variable using linear functions. The model is trained on the pinball loss.
  \item Gradient Boosting (GB): GB creates an ensemble of decision trees, and each new tree improves on the predictions of the previous ones. The method works by iteratively adding trees to reduce the residual errors from the current model.
  \item Light Gradient Boosting Machine (LightGBM): LightGBM  is a highly efficient  GB framework designed for large-scale data processing.
  \item Quantile Regression Neural Network (QRNN): Instead of focusing on the linear function class, QRNN estimates the relationship between independent variables and a specified quantile of the dependent variable using neural networks. It can capture complex, non-linear relationships.
\end{enumerate}
We implement two variations of the algorithms on the simulated dataset: the original version without the calibration and the calibrated version. The calibrated version includes an additional step of local calibration.

\paragraph{Metrics:} For each case, we test for quantiles 0.25, 0.5, and 0.75. We use the empirical pinball loss of the test dataset $\cD_{test}=\{(x_i^{test}, y_i^{test})\}$ as the evaluation metric. Suppose $\hat{f}(x)$ is the predicted quantile for context $x$; then the empirical pinball loss is
\[
\frac{1}{|\cD_{test}|}\left[\sum_{i=1}^{|\cD_{test}|}\frac{c_o}{c_o+c_u} \cdot (\hat{f}(x_i^{test})-y_i^{test})^+ + \frac{c_u}{c_o+c_u}\cdot (y_i^{test}-\hat{f}(x_i^{test}))^+\right].
\]

\begin{figure}[h!]
    \centering
    \subfigure[GB]
    {\includegraphics[width = .48\linewidth]{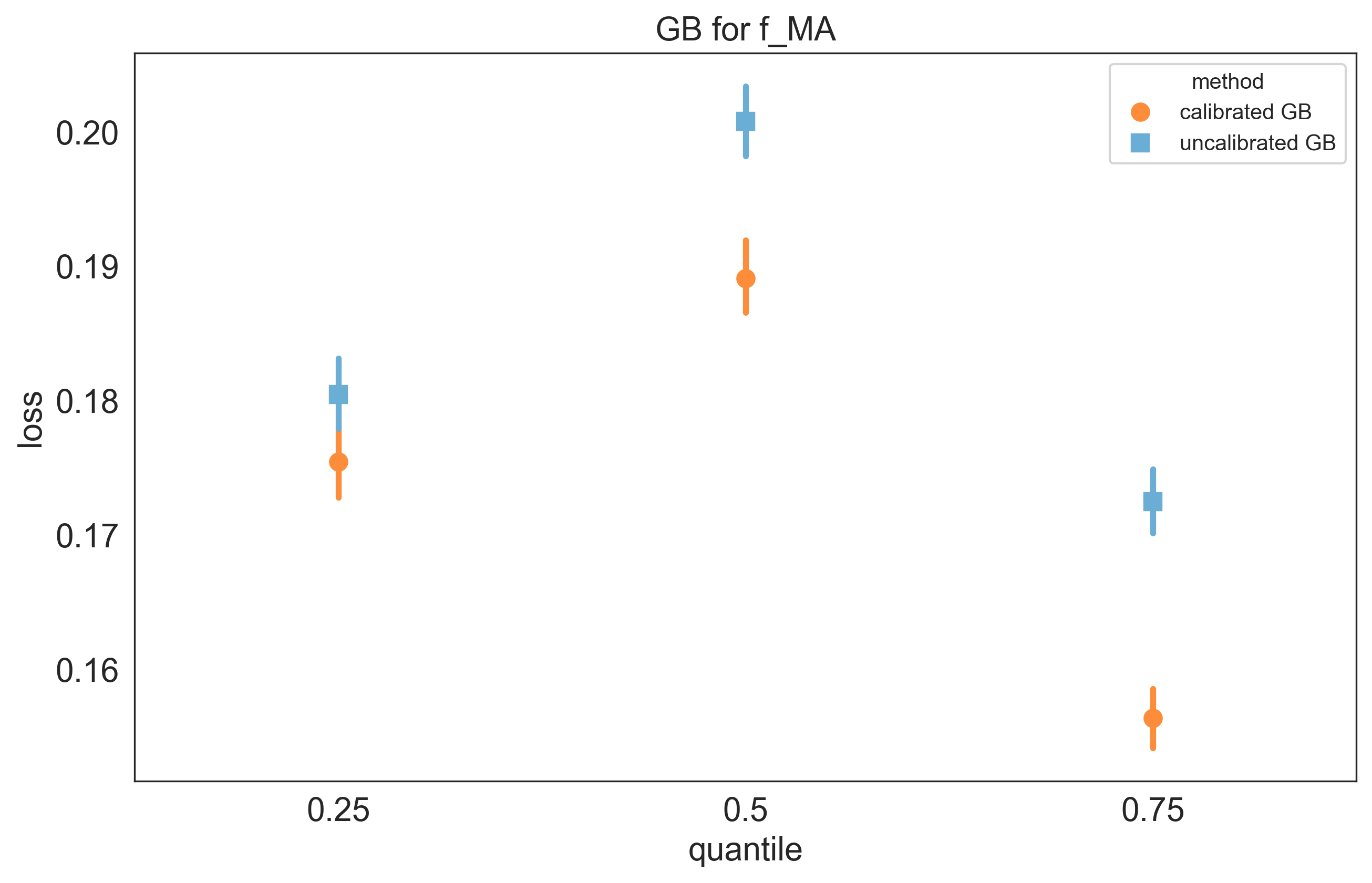}}
    \subfigure[LightGBM]
    {\includegraphics[width = .48\linewidth]{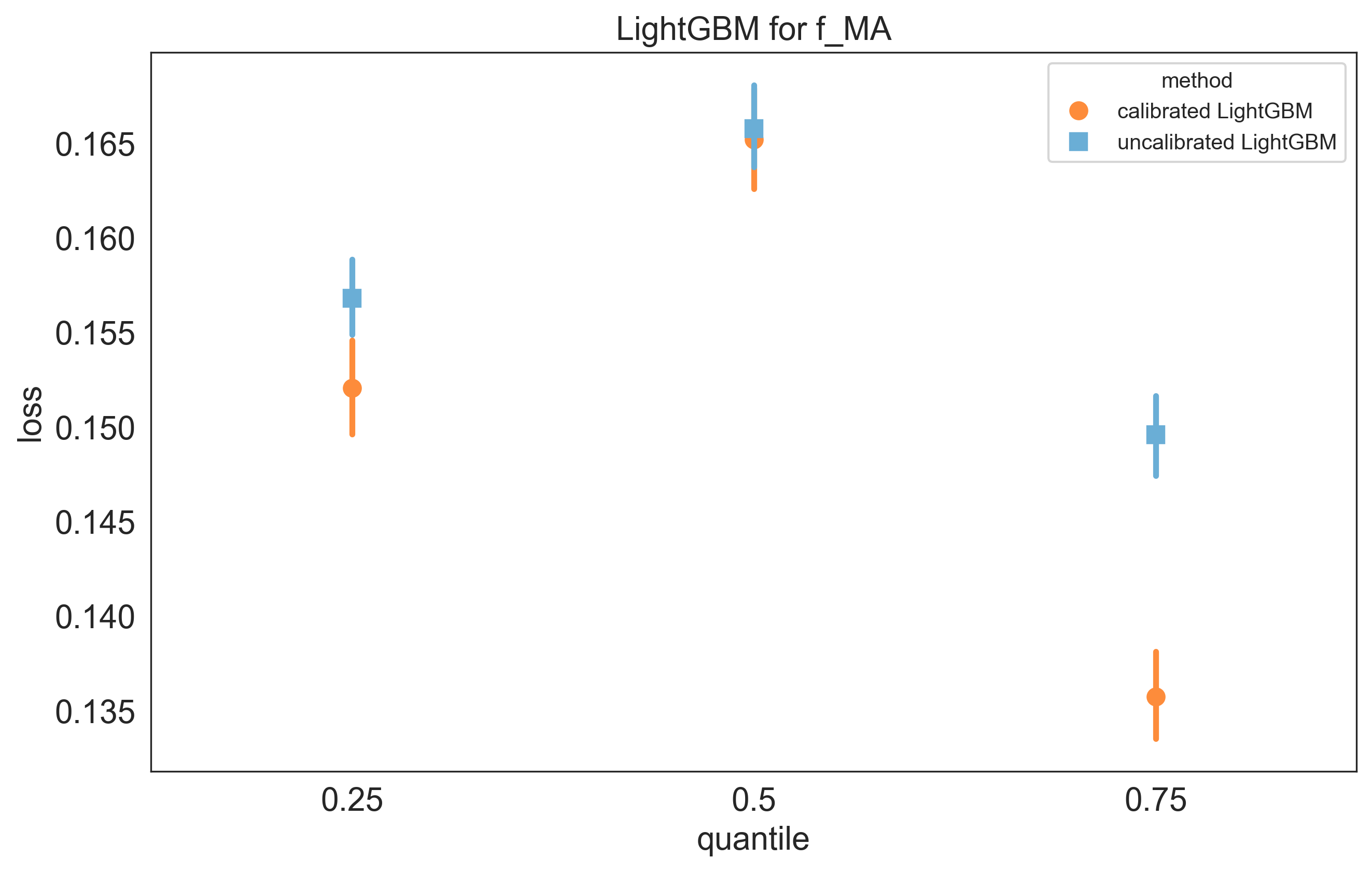}}
    \subfigure[LinearQR]
    {\includegraphics[width = .48\linewidth]{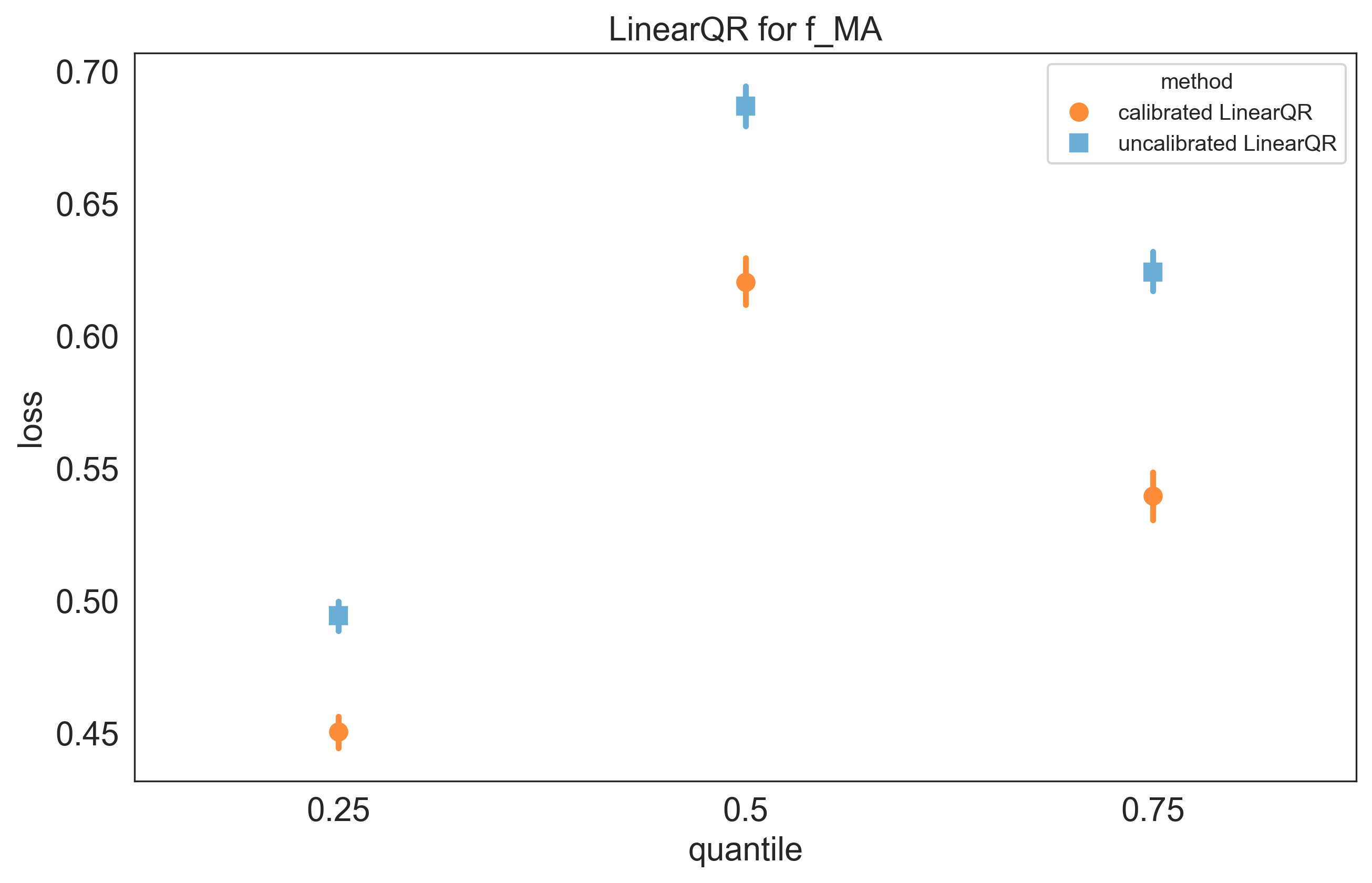}}
     \subfigure[QRNN]
    {\includegraphics[width = .48\linewidth]{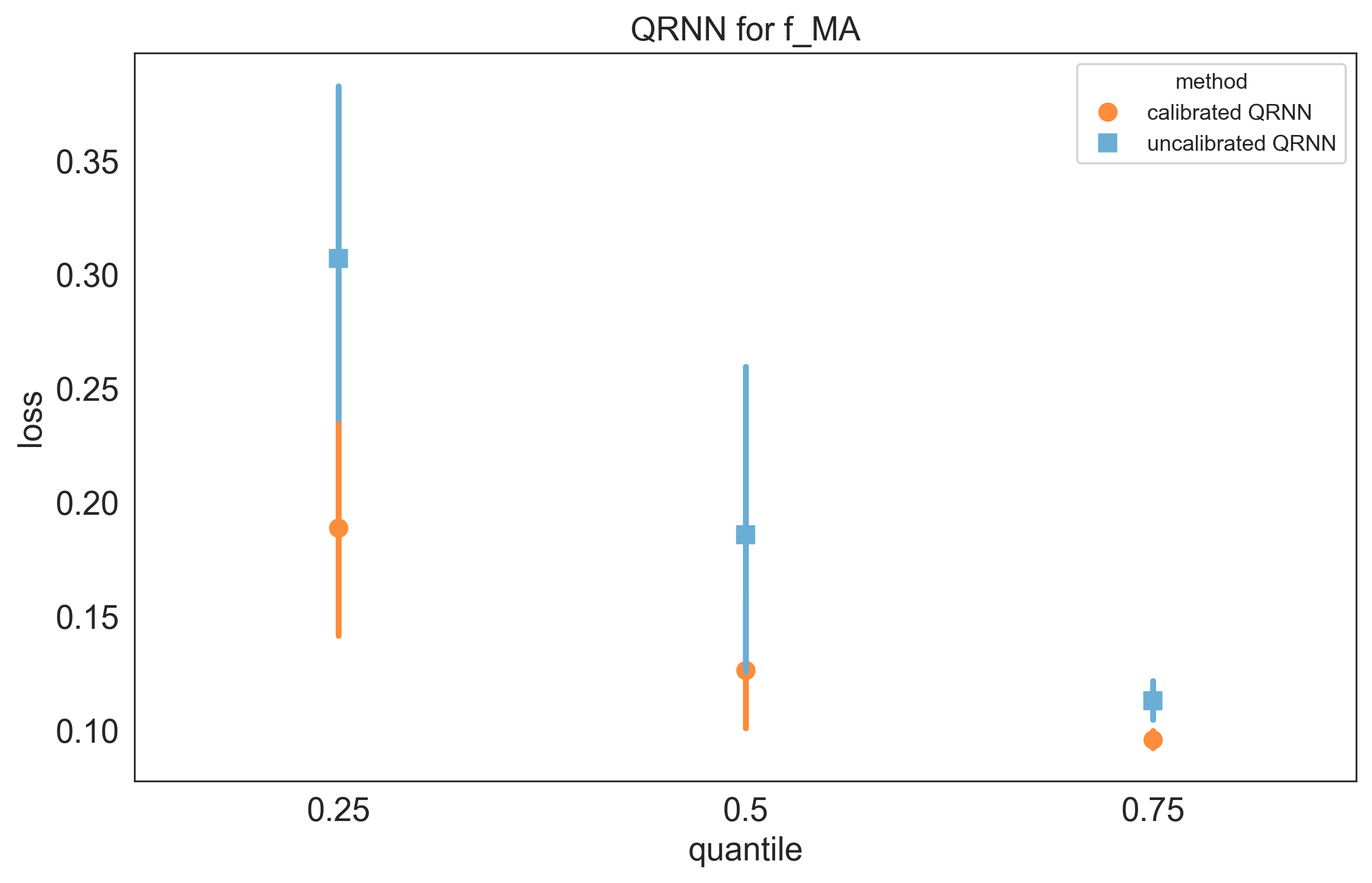}}
    \caption{Empirical pinball loss for MA model.}
    \label{fig: MA model}
\end{figure}

\paragraph{Results:} Figure \ref{fig: MA model} presents the empirical pinball loss of the MA model. Across all algorithms, our proposed method consistently outperforms the benchmark approach without the calibration step. Notably, for GB, the advantage of using the calibrated GB increases with higher quantiles, reducing the empirical loss by approximately 9.3\% at the 0.75 quantile level. For both LightGBM and LinearQR, the empirical loss is similarly reduced -- by about 9.2\% and 13.5\%, respectively, at the 0.75 quantile. Interestingly, QRNN demonstrates a different pattern, where the performance gap is more pronounced at lower quantiles. Specifically, at the 0.25 quantile, calibrated QRNN reduces the empirical loss by more than 38.6\%; at the 0.5 quantile, calibrated QRNN reduces the empirical loss by around 31.9\%, marking a significant improvement.

\begin{figure}[h!]
    \centering
    \subfigure[GB]
    {\includegraphics[width = .48\linewidth]{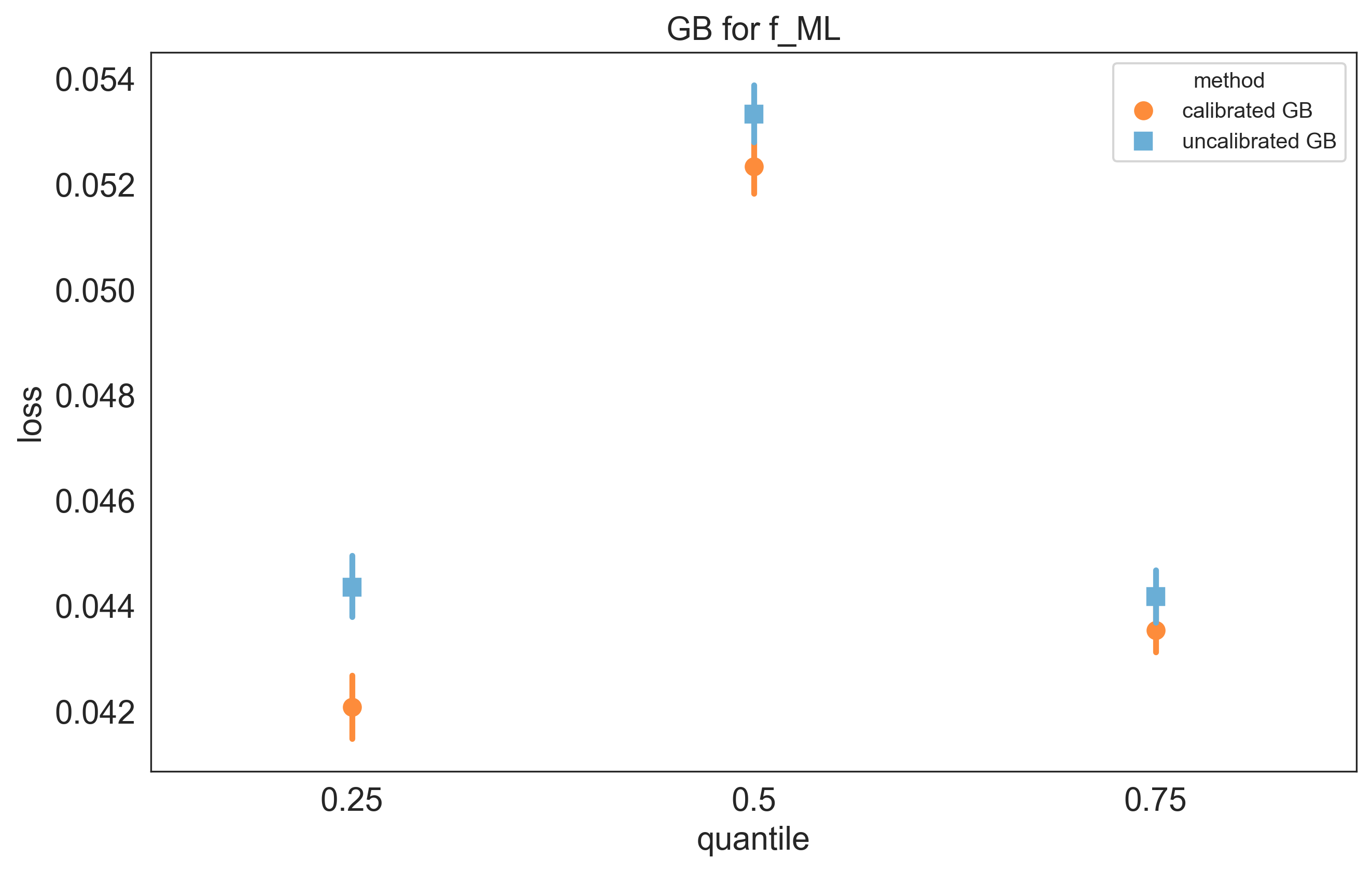}}
    \subfigure[LightGBM]
    {\includegraphics[width = .48\linewidth]{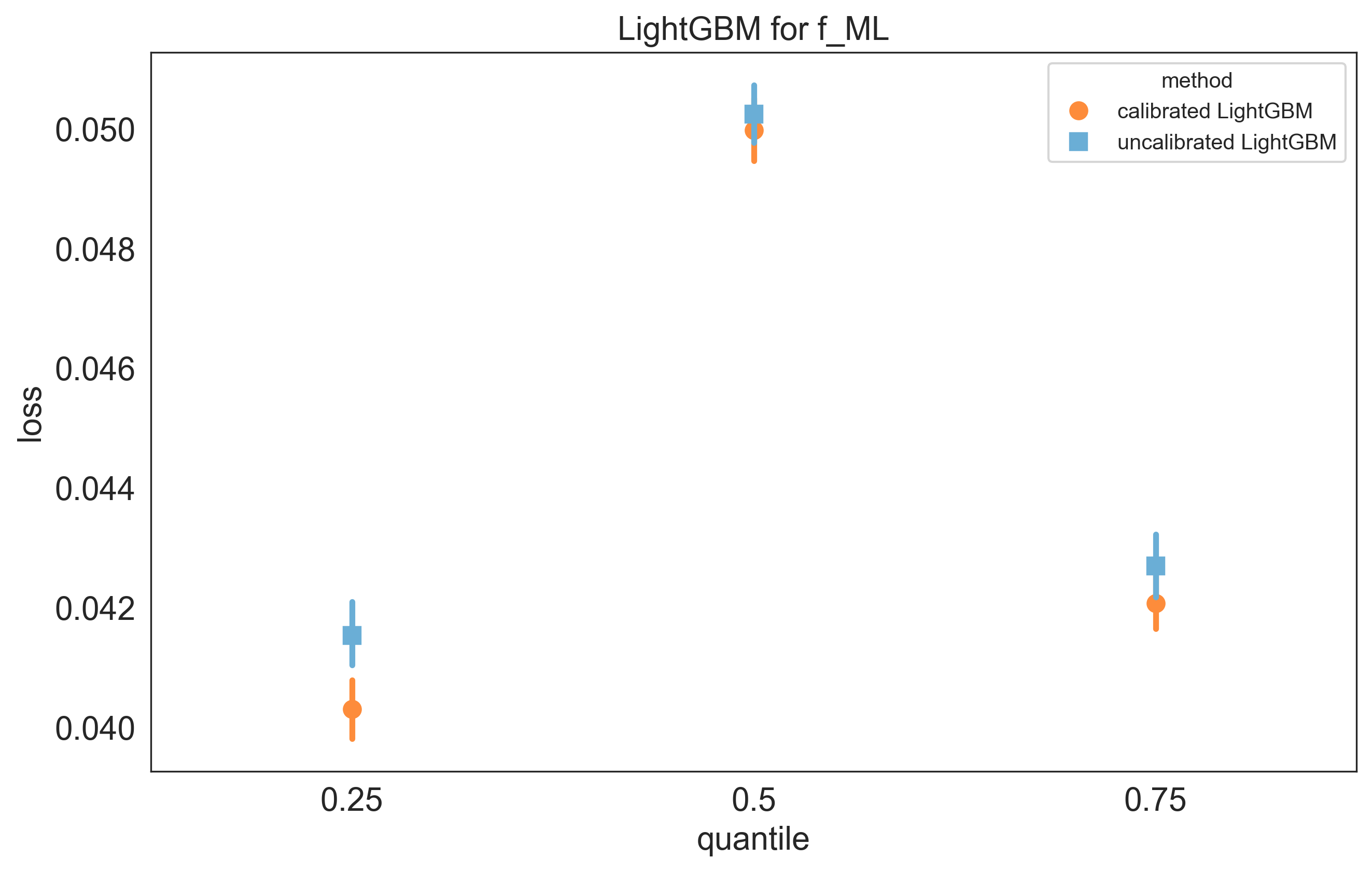}}
    \subfigure[LinearQR]
    {\includegraphics[width = .48\linewidth]{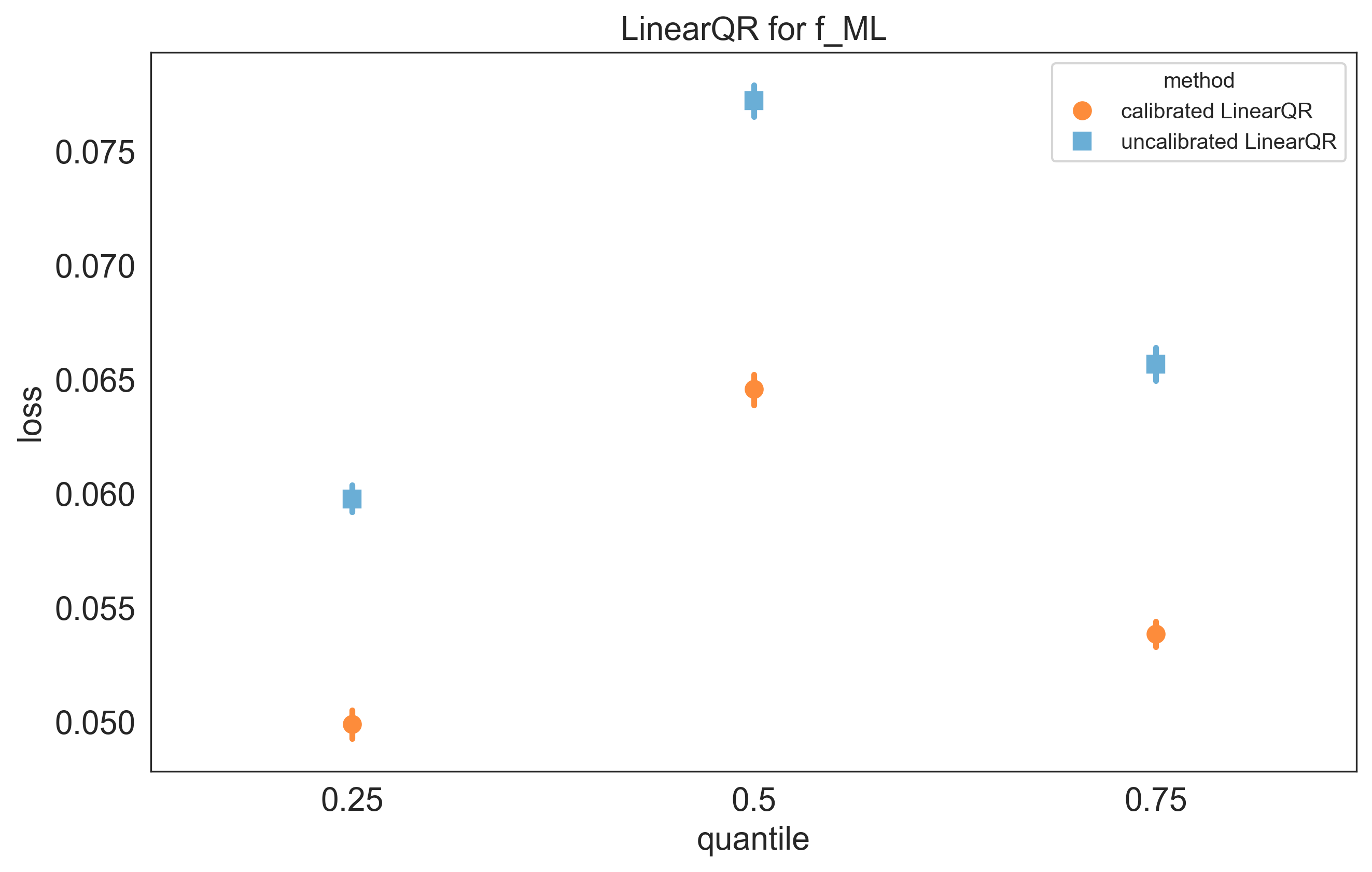}}
     \subfigure[QRNN]
    {\includegraphics[width = .48\linewidth]{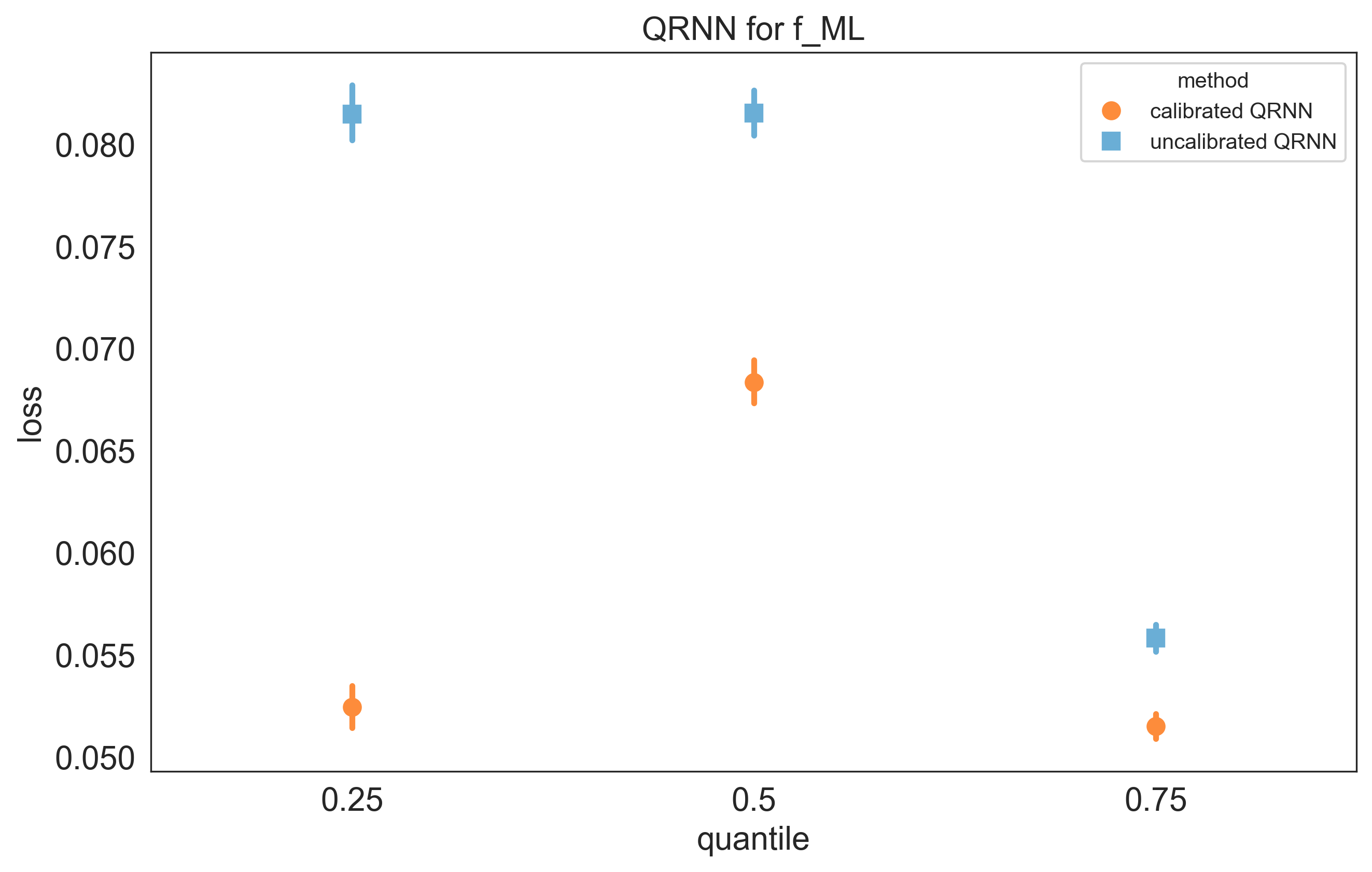}}
    \caption{Empirical pinball loss for ML model.}
    \label{fig: ML model}
\end{figure}

Figure \ref{fig: ML model} presents the empirical pinball loss of the ML model. All quantile regression algorithms show better performance with the additional local calibration step, but the improvement varies among different algorithms and quantiles. The improvements of QRNN and LinearQR are more pronounced. Specifically, calibrated QRNN reduces the empirical loss of the uncalibrated QRNN by about 35.6\%, 16.2\%, and 7.7\% for quantiles 0.25, 0.5, and 0.75, respectively; the calibrated LinearQR improves over the  uncalibrated LinearQR by about 16.5\%, 16.4\%, and 18.0\% for quantiles 0.25, 0.5, and 0.75, respectively.

\subsubsection{The effect of local pooling}
\begin{figure}[h!]
    \centering
    \subfigure[quantile 0.25]
    {\includegraphics[width = .32\linewidth]{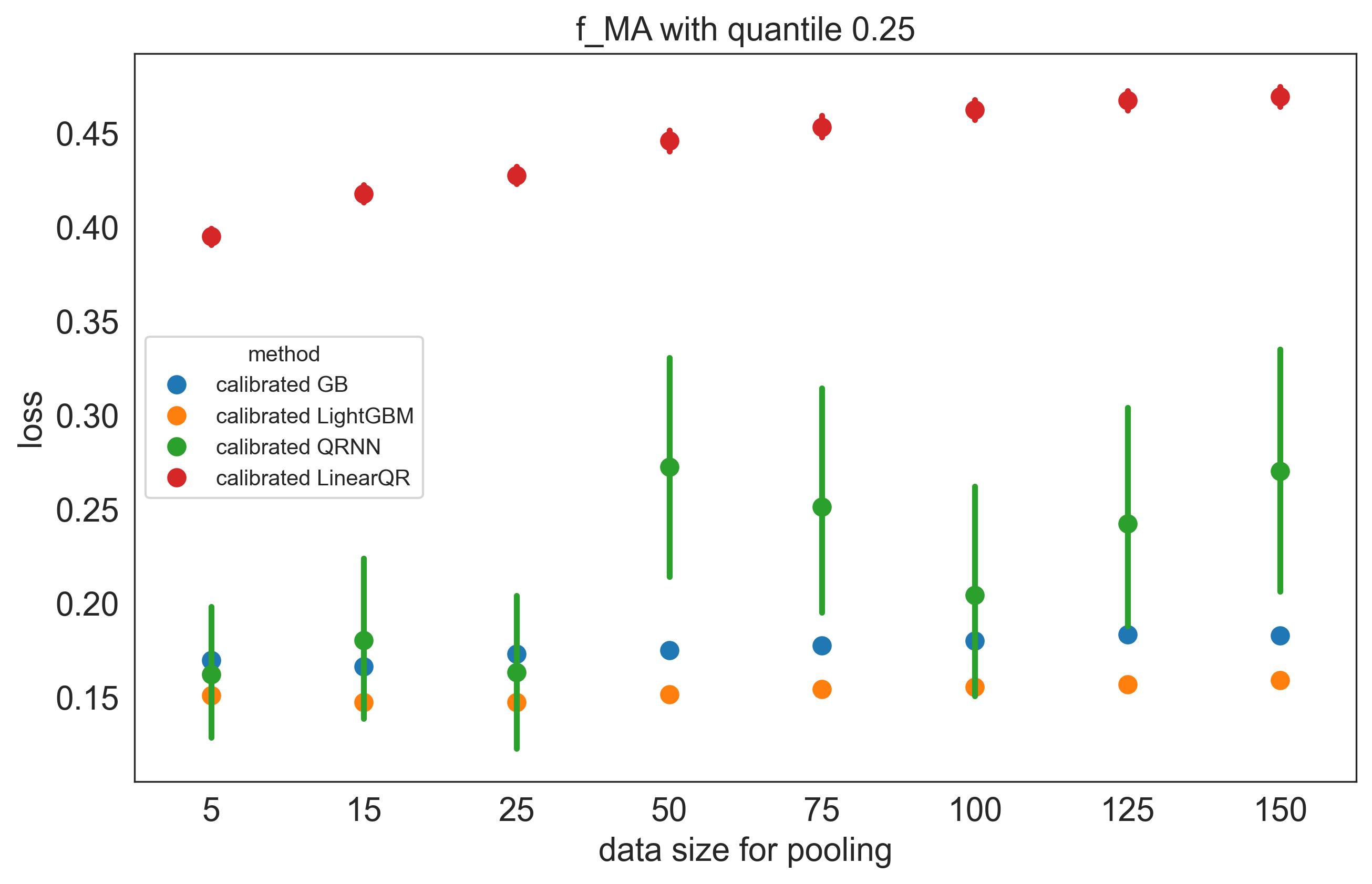}}
    \subfigure[quantile 0.5]
    {\includegraphics[width = .32\linewidth]{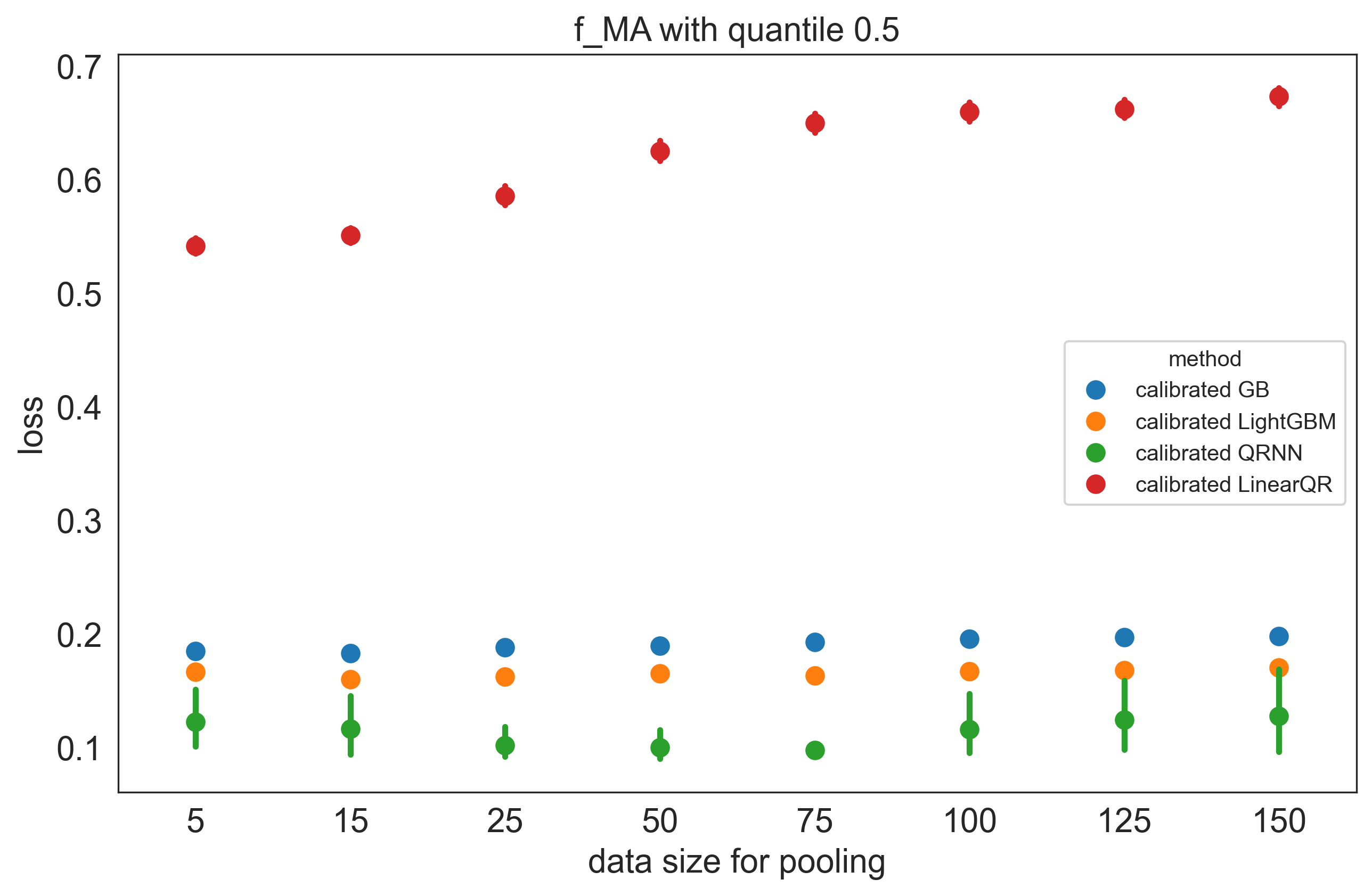}}
    \subfigure[quantile 0.75]
    {\includegraphics[width = .32\linewidth]{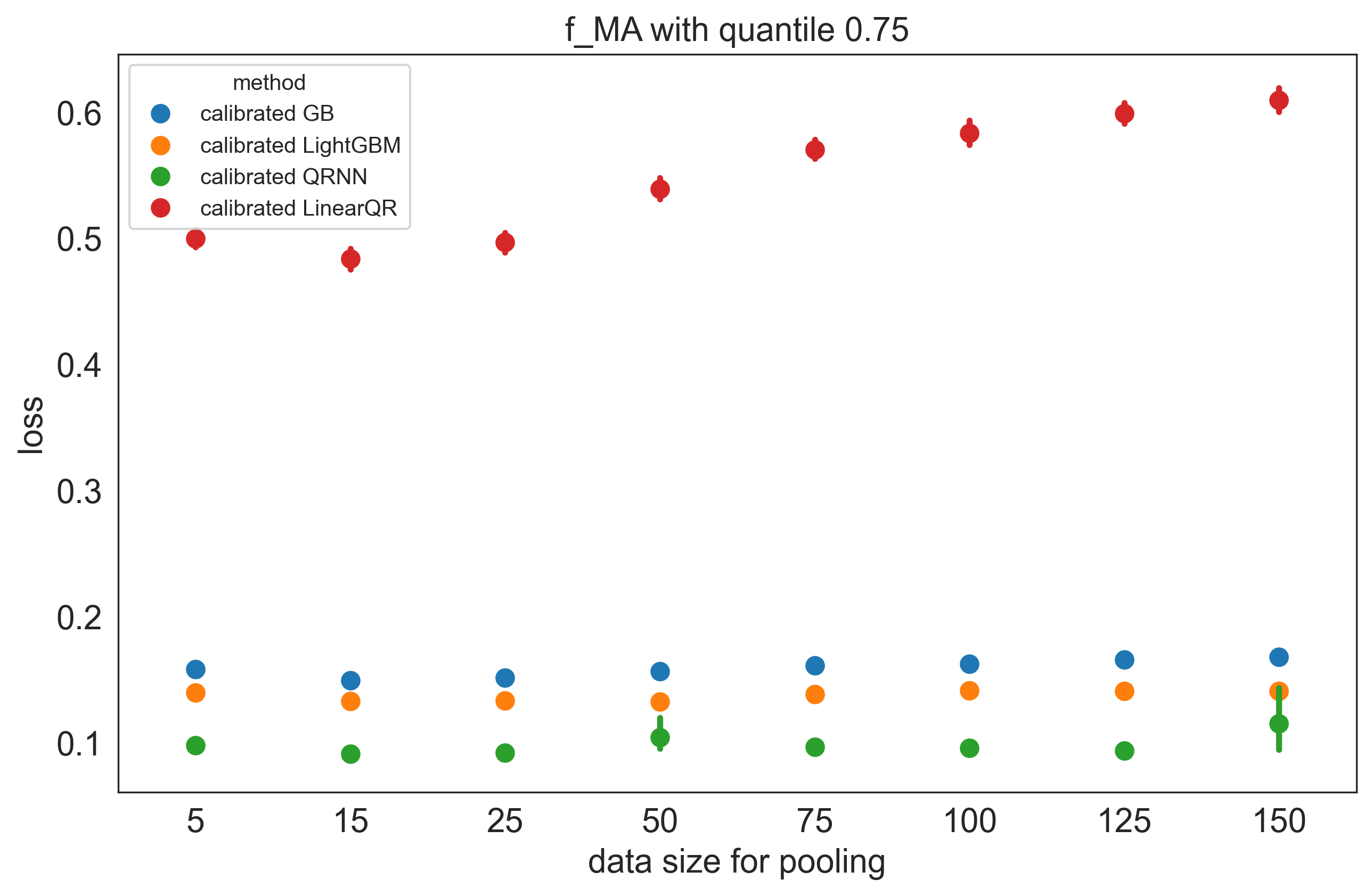}}
    \caption{Empirical pinball loss for MA model for different pooling sizes (four algorithms).}
    \label{fig: pooling MA}
\end{figure}

\begin{figure}[h!]
    \centering
    \subfigure[quantile 0.25]
    {\includegraphics[width = .32\linewidth]{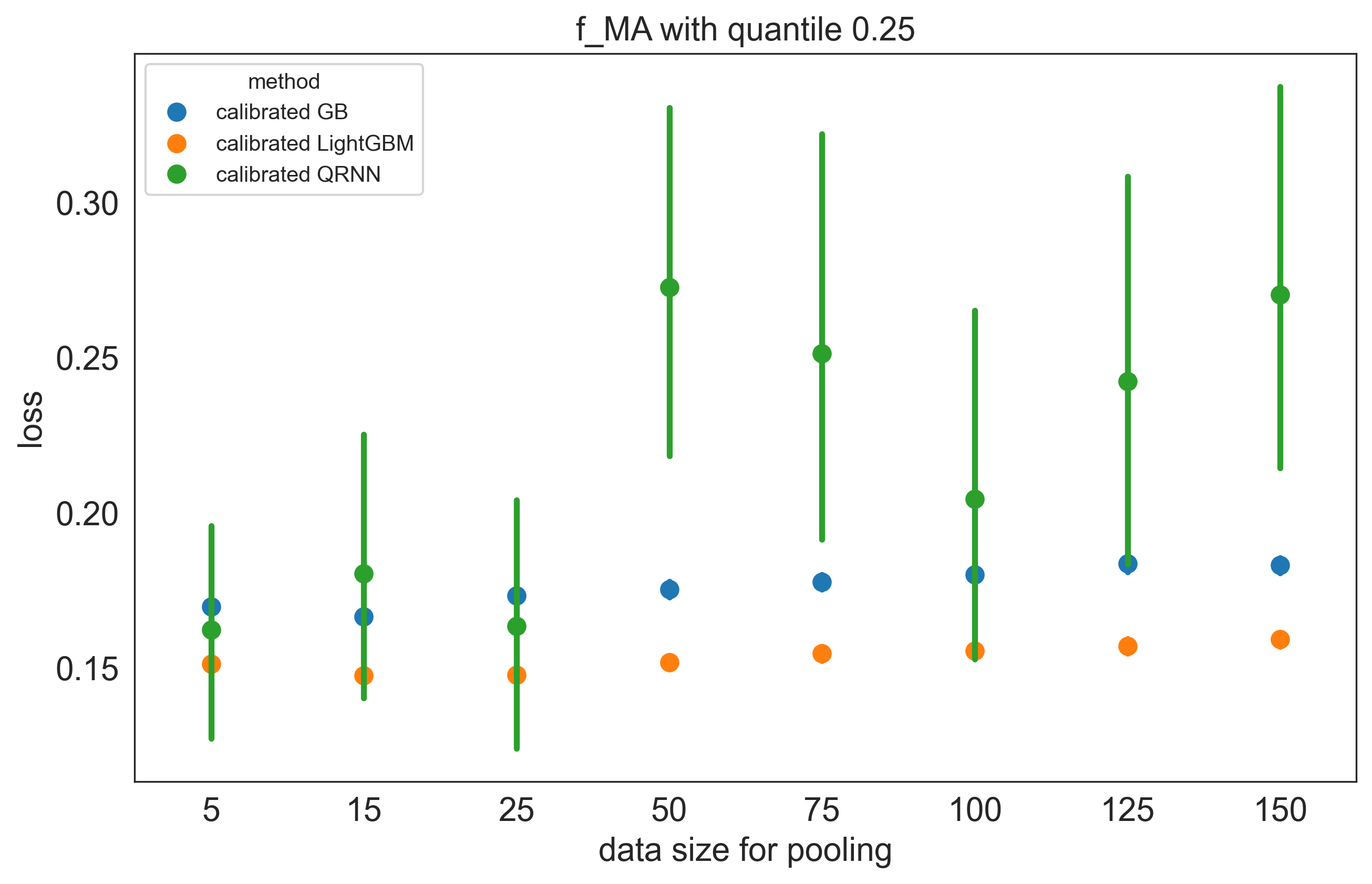}}
    \subfigure[quantile 0.5]
    {\includegraphics[width = .32\linewidth]{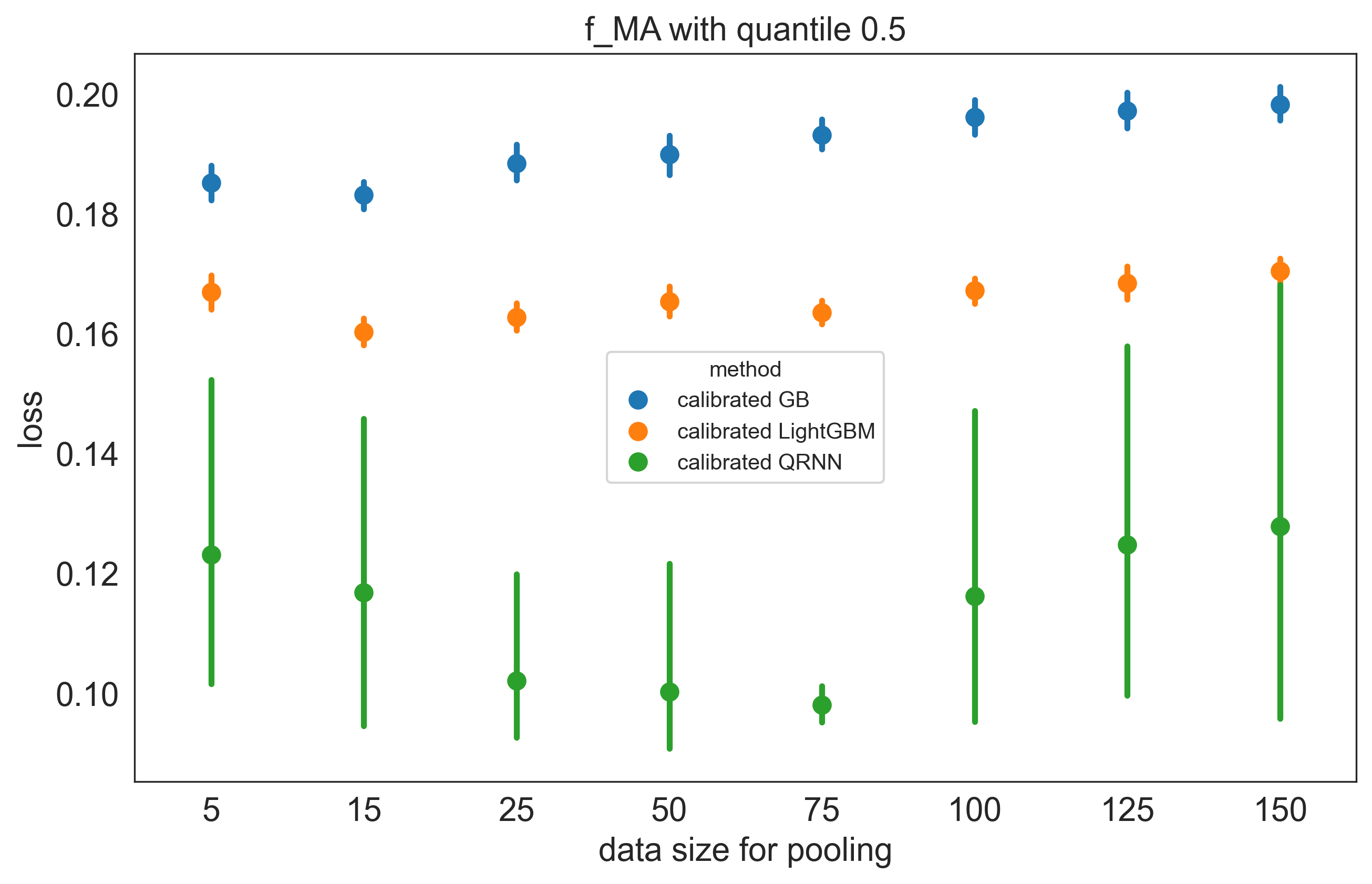}}
    \subfigure[quantile 0.75]
    {\includegraphics[width = .32\linewidth]{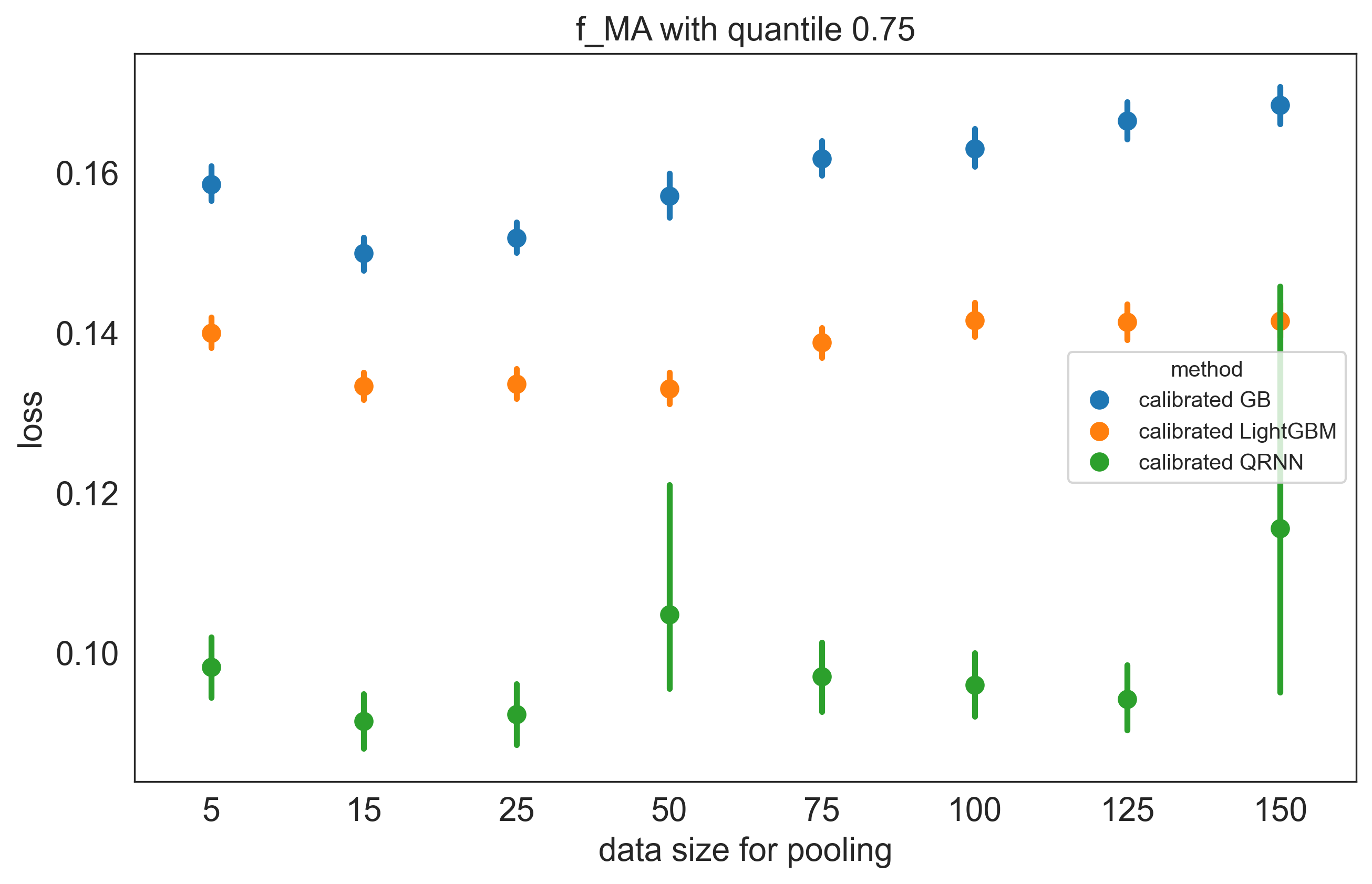}}
    \caption{Empirical pinball loss for MA model for different pooling sizes (three algorithms).}
    \label{fig: pooling MA 2}
\end{figure}

Algorithm \ref{Alg: CCQP} focuses on training and calibration across the entire region; but as our discussion in Section \ref{sec: quality vs quantity} suggests,  performance could be further enhanced by pooling neighboring data points. To improve computational efficiency, we train the model on the entire region but calibrate using pooled neighboring data points.
Figure \ref{fig: pooling MA} illustrates the effect on performance of varying pooling sizes. The results indicate that the effects of pooling differ among the various algorithms and across different quantiles. Notably, the performance of the calibrated LinearQR deteriorates as the pooling region expands.
To better highlight the trends for the other three algorithms, we present them in Figure \ref{fig: pooling MA 2}. The empirical loss of QRNN exhibits fluctuations as the pooling size increases: At quantile 0.25, it rises from 25 to 50, then decreases from 50 to 100, and subsequently increases again beyond 100. At quantile 0.5, the empirical loss initially decreases and then increases with pooling size, achieving optimal performance when pooling 75 neighboring data points. Calibrating using the entire region results in a significantly higher empirical loss compared to local calibration. For quantile 0.5, the best performance is attained with a pooling size of 75. Interestingly, the double-descent phenomenon is observable in both the 0.25 and 0.75 quantile scenarios. 
Calibrated GB and LightGBM demonstrate relative robustness with respect to pooling size, achieving optimal performance at a pooling size of about 15. 

\begin{figure}[h!]
    \centering
    \subfigure[quantile 0.25]
    {\includegraphics[width = .32\linewidth]{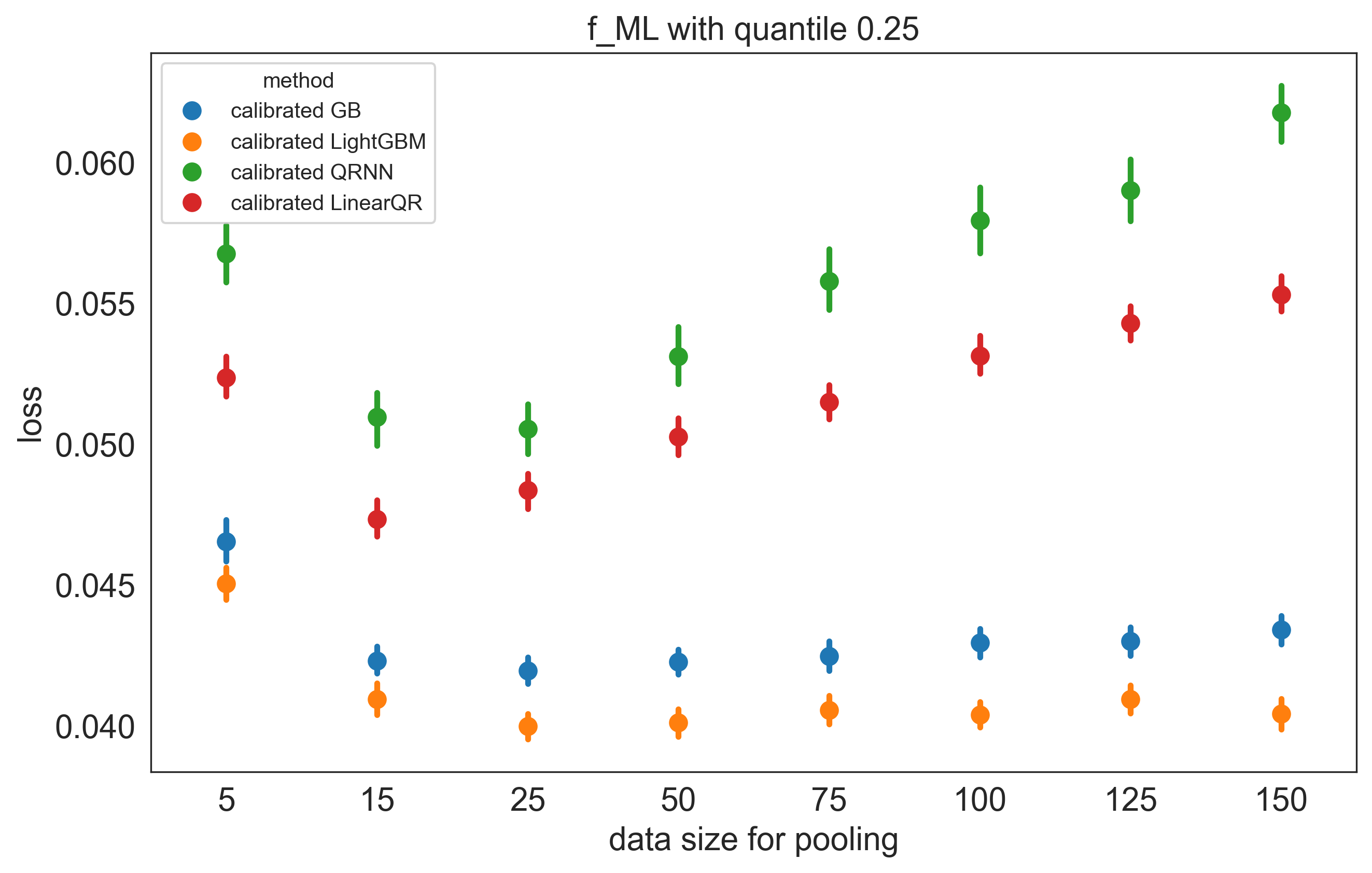}}
    \subfigure[quantile 0.5]
    {\includegraphics[width = .32\linewidth]{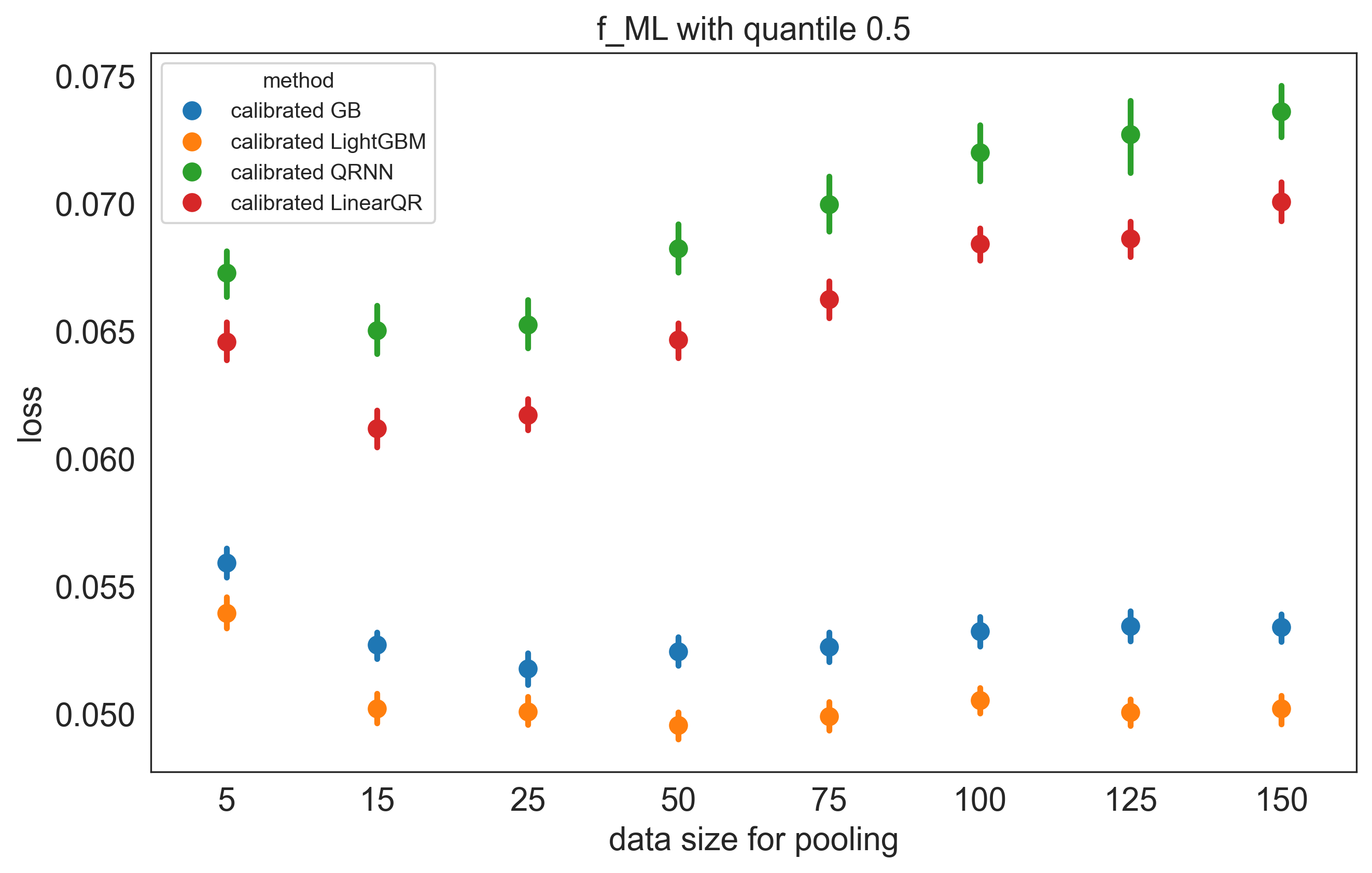}}
    \subfigure[quantile 0.75]
    {\includegraphics[width = .32\linewidth]{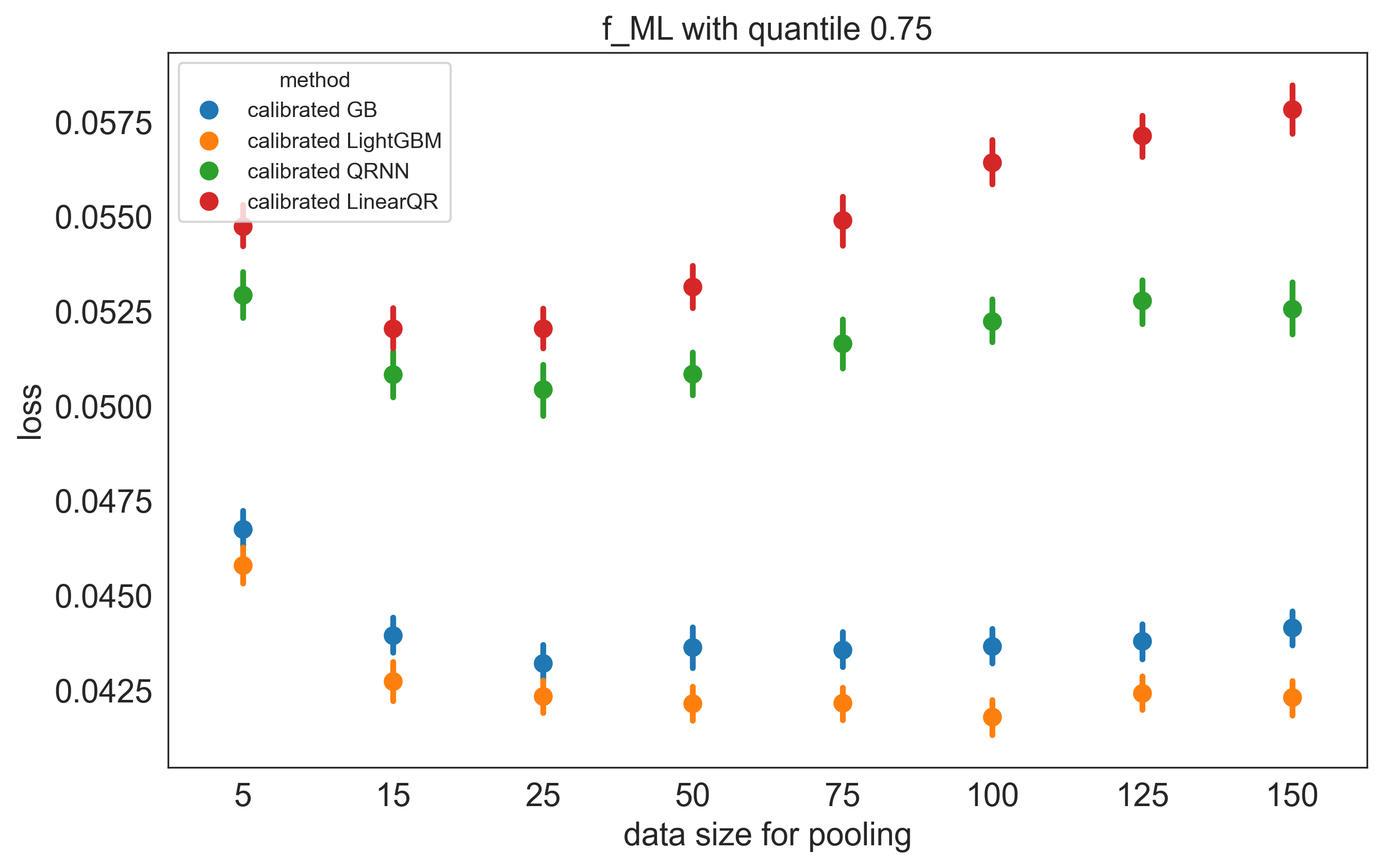}}
    \caption{Empirical pinball loss for ML model for different pooling sizes.}
    \label{fig: pooling ML}
\end{figure}

Figure \ref{fig: pooling ML} presents the empirical pinball loss of the ML model across various pooling sizes. In nearly all cases, the loss decreases initially as the pooling size increases, but then it begins to rise, with the optimal pooling size falling between 15 and 25. Calibrating on the entire dataset corresponds to the largest pooling size. This result suggests that local calibration offers performance improvements over calibration based on the entire dataset. It aligns with our theoretical findings that model bias tends to affect nearby data points similarly, meaning that pooling locally can help mitigate the bias caused by model misspecification. On the other hand, pooling too few nearby points can also result in poor performance because of the presence of observational noise.

\subsubsection{The effect of sample size}
\begin{figure}[H]
    \centering
    \subfigure[GB]
    {\includegraphics[width = .32\linewidth]{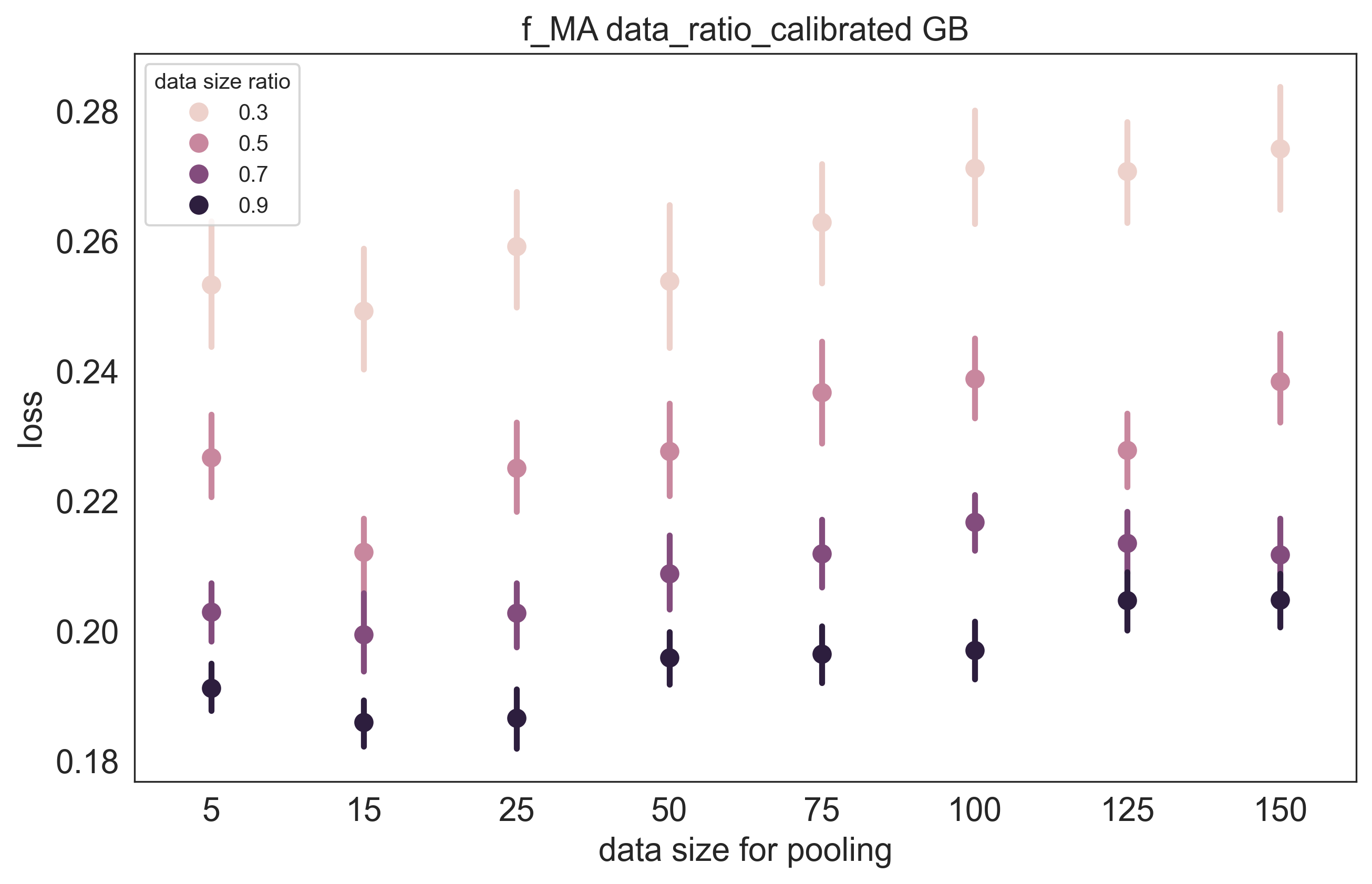}}
    \subfigure[LightGBM]
    {\includegraphics[width = .32\linewidth]{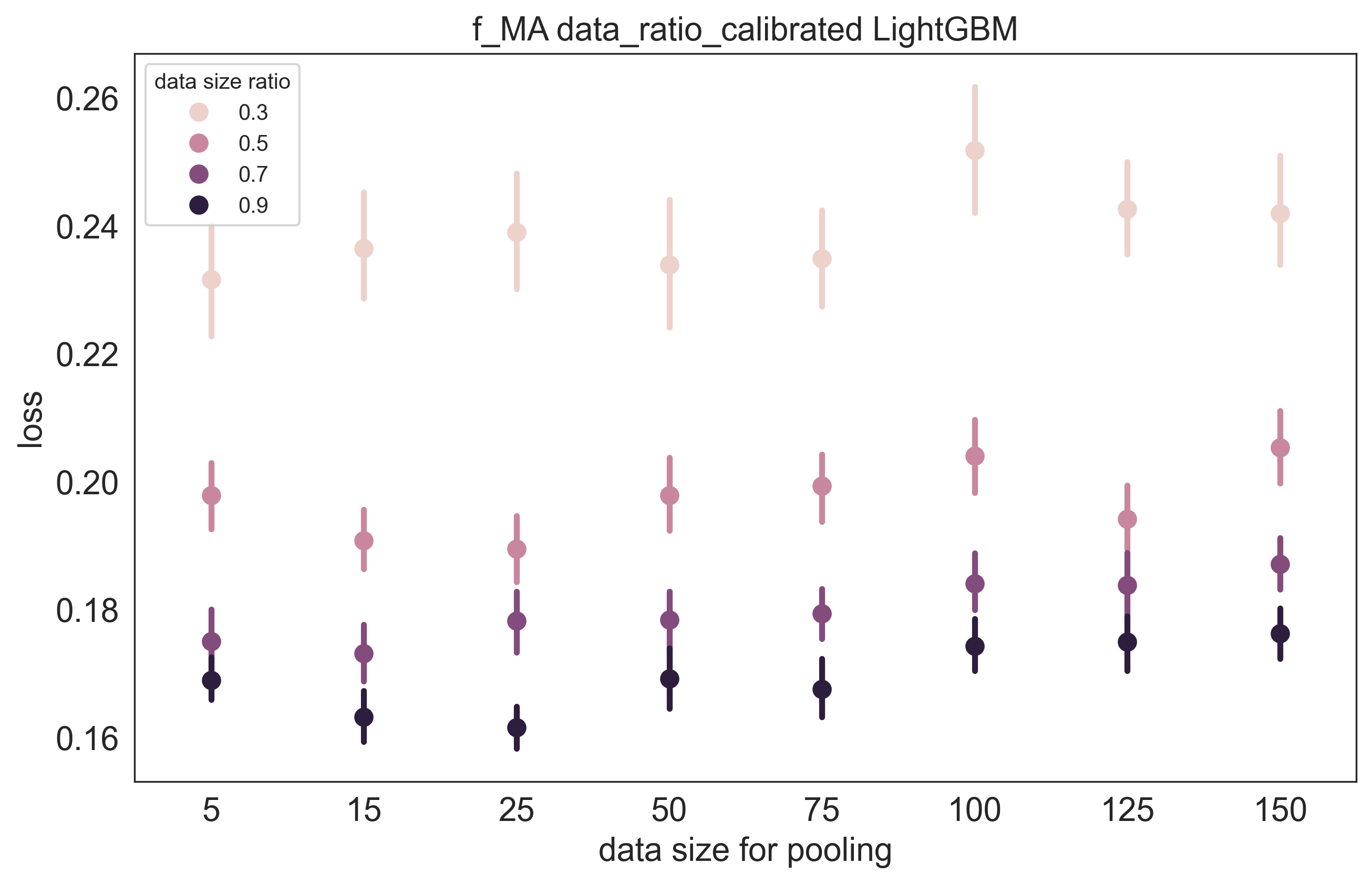}}
     \subfigure[QRNN]
    {\includegraphics[width = .32\linewidth]{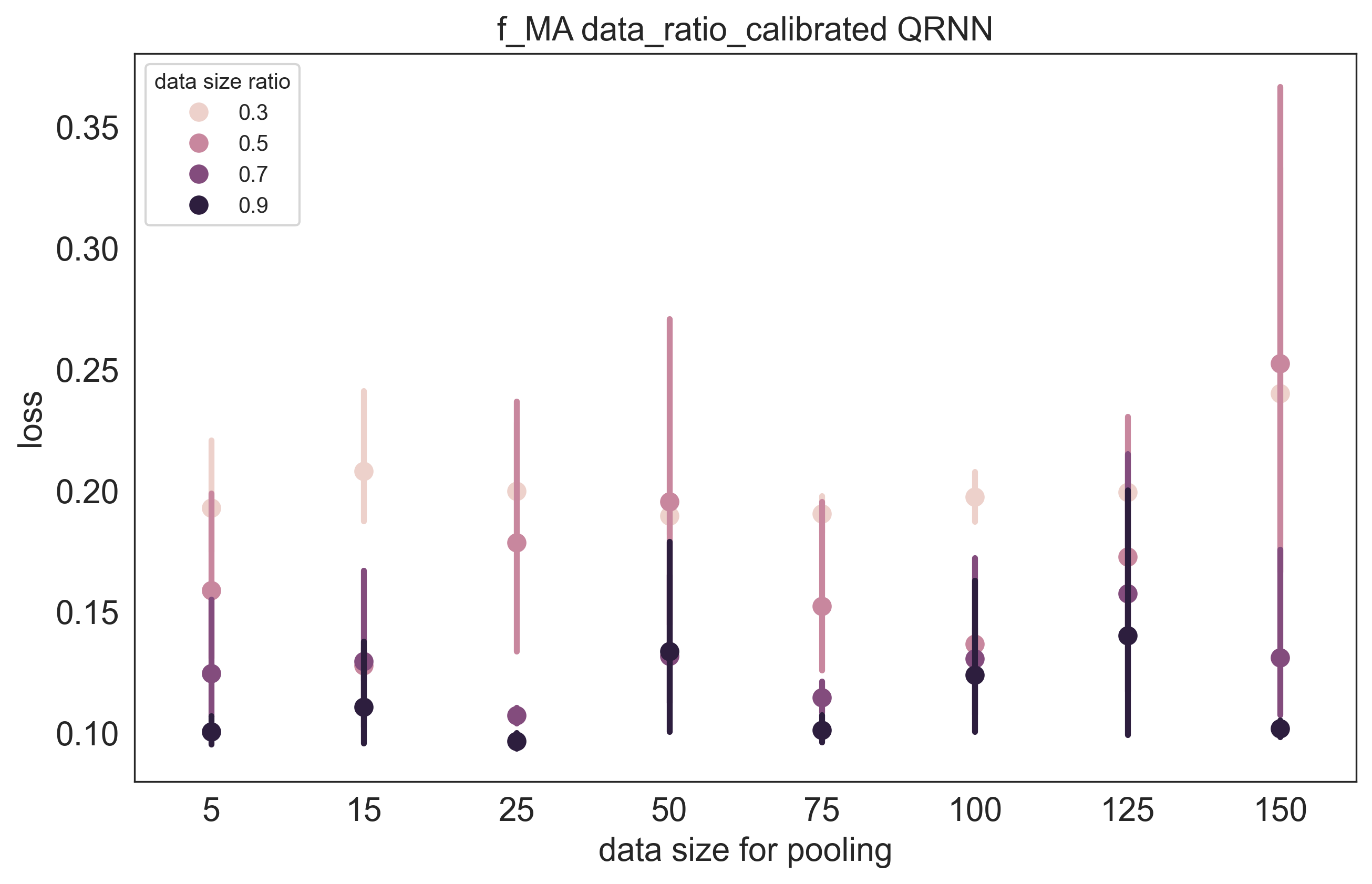}}
    \caption{Empirical pinball loss for MA model with different data sizes.}
    \label{fig: MA model data ratio}
\end{figure}

In this experiment, we vary the sample size by randomly selecting a fixed proportion of data from a total of 2,000 data points, with size ratios chosen from ${0.3, 0.5, 0.7, 0.9}$. The quantile is chosen as 0.5. Figure \ref{fig: MA model data ratio} presents the empirical pinball loss for the MA model across different data sizes. As expected, the prediction performance improves as the amount of data increases. 
In addition, Figure \ref{fig: MA model data ratio 3} in Appendix \ref{appendix: numerical} shows results for a smaller data size ratio of 0.1, which performs significantly worse than the ratio of 0.3.

We also observe distinct patterns across different algorithms. For GB and LinearQR, the best performance is consistently achieved by pooling 15 and 5 neighboring points, respectively, across all sample sizes. However, for LightGBM and QRNN, an interesting pattern emerges. In the case of LightGBM, we compare pooling sizes of 25 and 75, and when the data size is small (ratio = 0.3), pooling more data points (75 points) improves performance. However, as the data size increases, pooling fewer points (25 points) becomes more beneficial. For QRNN, when the data size is small (ratio = 0.3), the best results are achieved by pooling 50 to 75 points. In contrast, with larger data sizes (ratios of 0.7 or 0.9), pooling fewer points (25 neighboring points) yields the best performance.

\begin{figure}[H]
    \centering
    \subfigure[LightGBM]
    {\includegraphics[width = .32\linewidth]{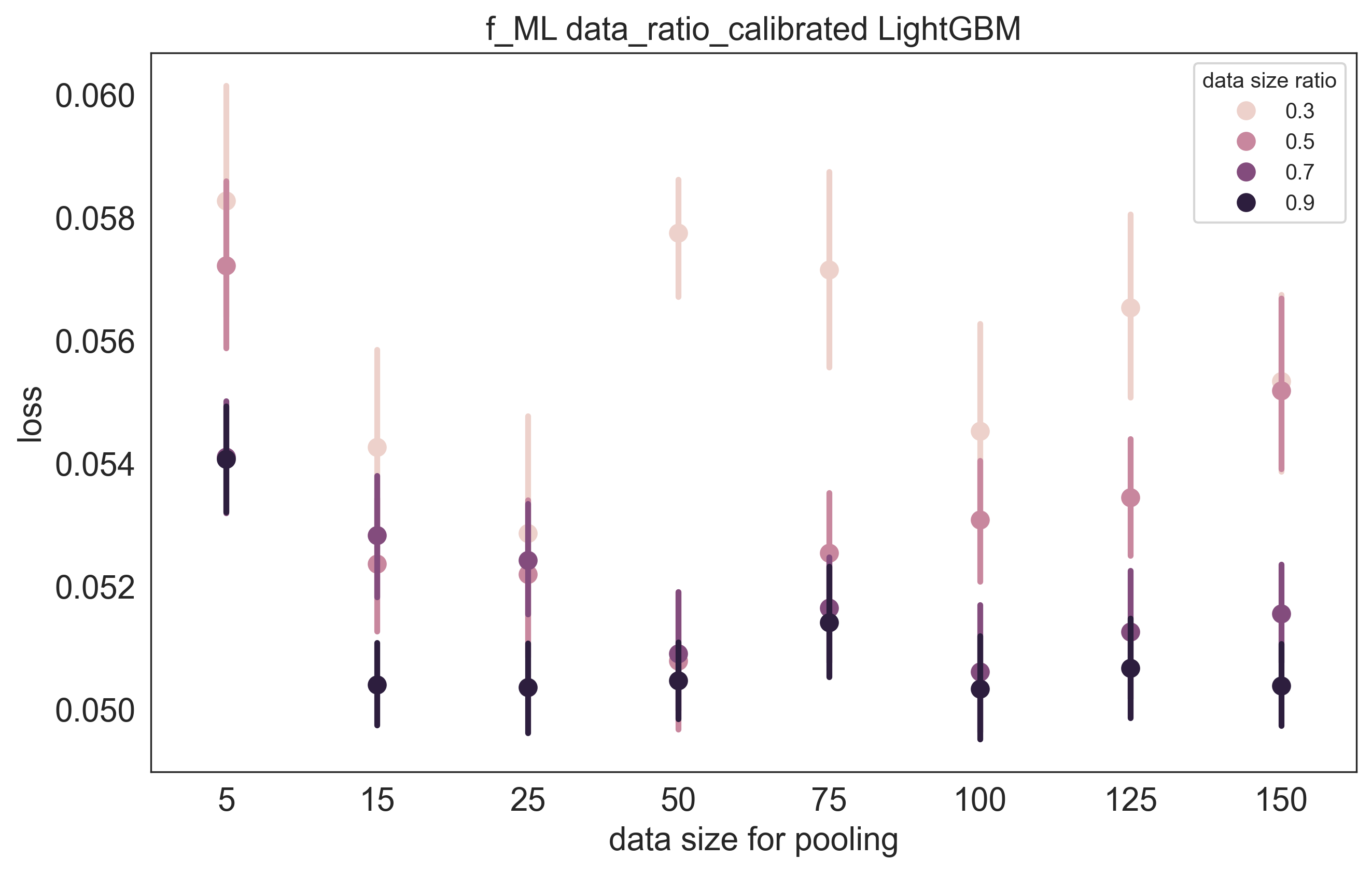}}
    \subfigure[LQR]
    {\includegraphics[width = .32\linewidth]{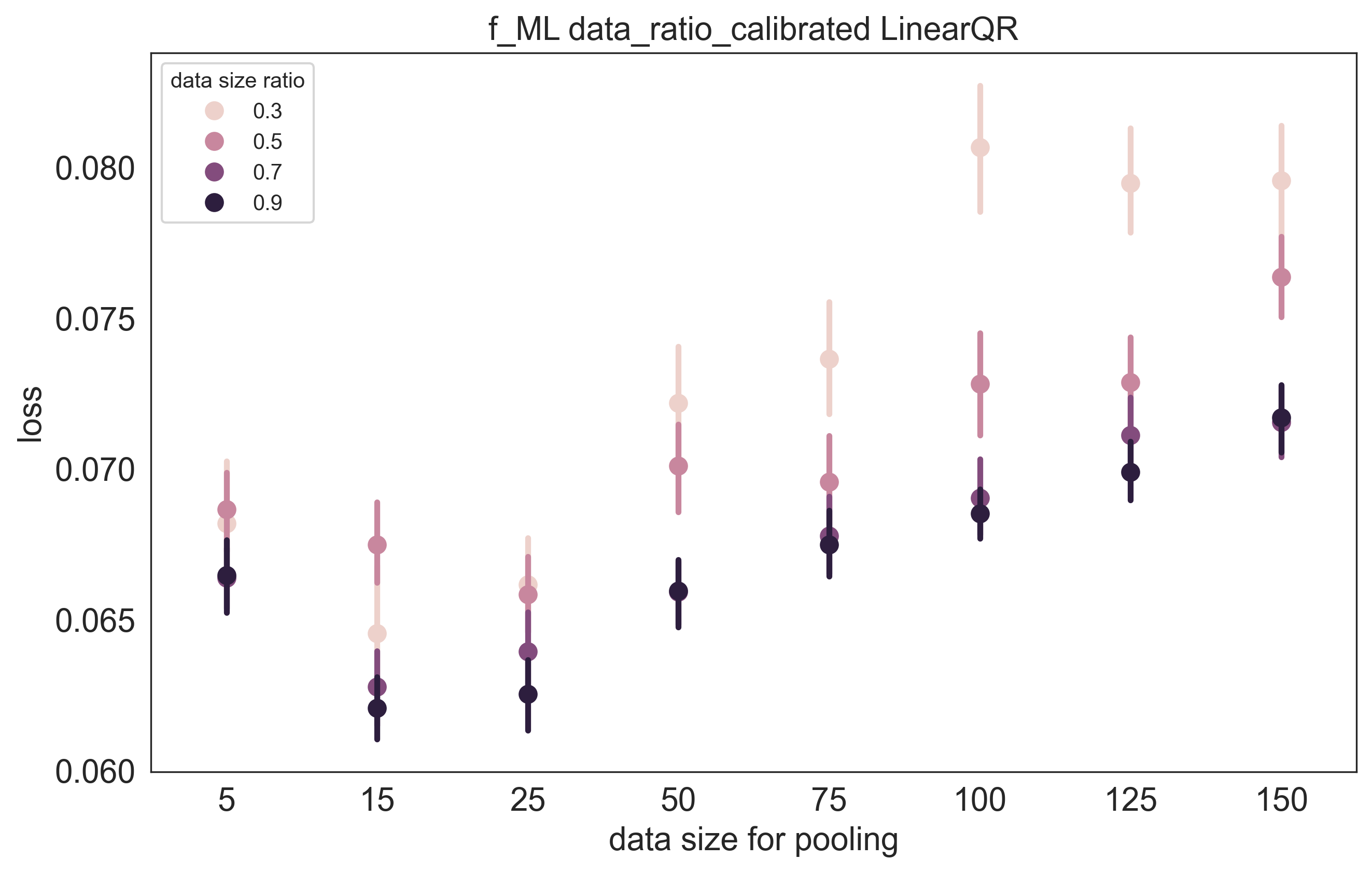}}
     \subfigure[QRNN]
    {\includegraphics[width = .32\linewidth]{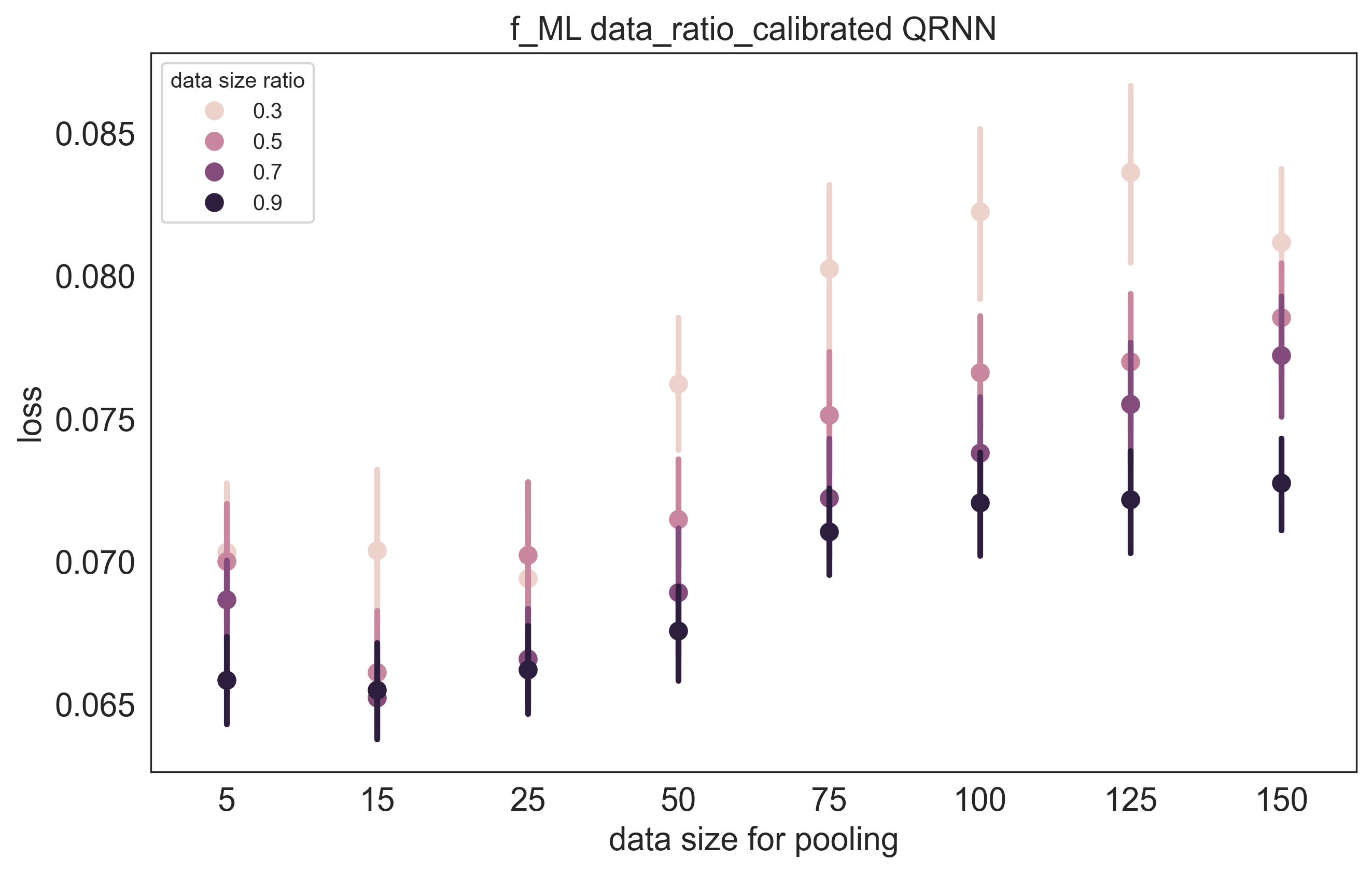}}
    \caption{Empirical pinball loss for ML model with different data sizes.}
    \label{fig: ML model data ratio}
\end{figure}

Figure \ref{fig: ML model data ratio} presents the empirical pinball loss for LightGBM, LinearQR, and QRNN across varying data sizes for the ML model. Detailed results, including those for a data ratio of 0.1, can be found in Appendix \ref{appendix: numerical}. For LightGBM, when the data size ratios are 0.5 and 0.7, the optimal pooling size is 50 data points. As the dataset increases, the performance of pooling between 15 and 50 data points becomes comparable. For LinearQR, at a data size ratio of 0.5, the optimal pooling size is 25, but when the ratio increases to 0.9, the best pooling size decreases to 15. A similar trend is observed in QRNN, where the optimal pooling size is 25 for a data ratio of 0.3, but the optimal size reduces to 15 as the dataset grows.

\subsection{Real Data: Capital Bikeshare}
 This dataset depicts the bike rental demand in the Capital Bikeshare program in Washington, DC \citep{fanaee2014event}.\footnote{https://archive.ics.uci.edu/dataset/275/bike+sharing+dataset}
 The features of the dataset include season of the year, year, month, hour, holiday, weekday, workingday, weather situation, temperature, ``feels like'' temperature, relative humidity, and wind speed. The bike rental demand is hourly-based. 

\begin{figure}[!h]
    \centering
    {\includegraphics[width = .65\linewidth]{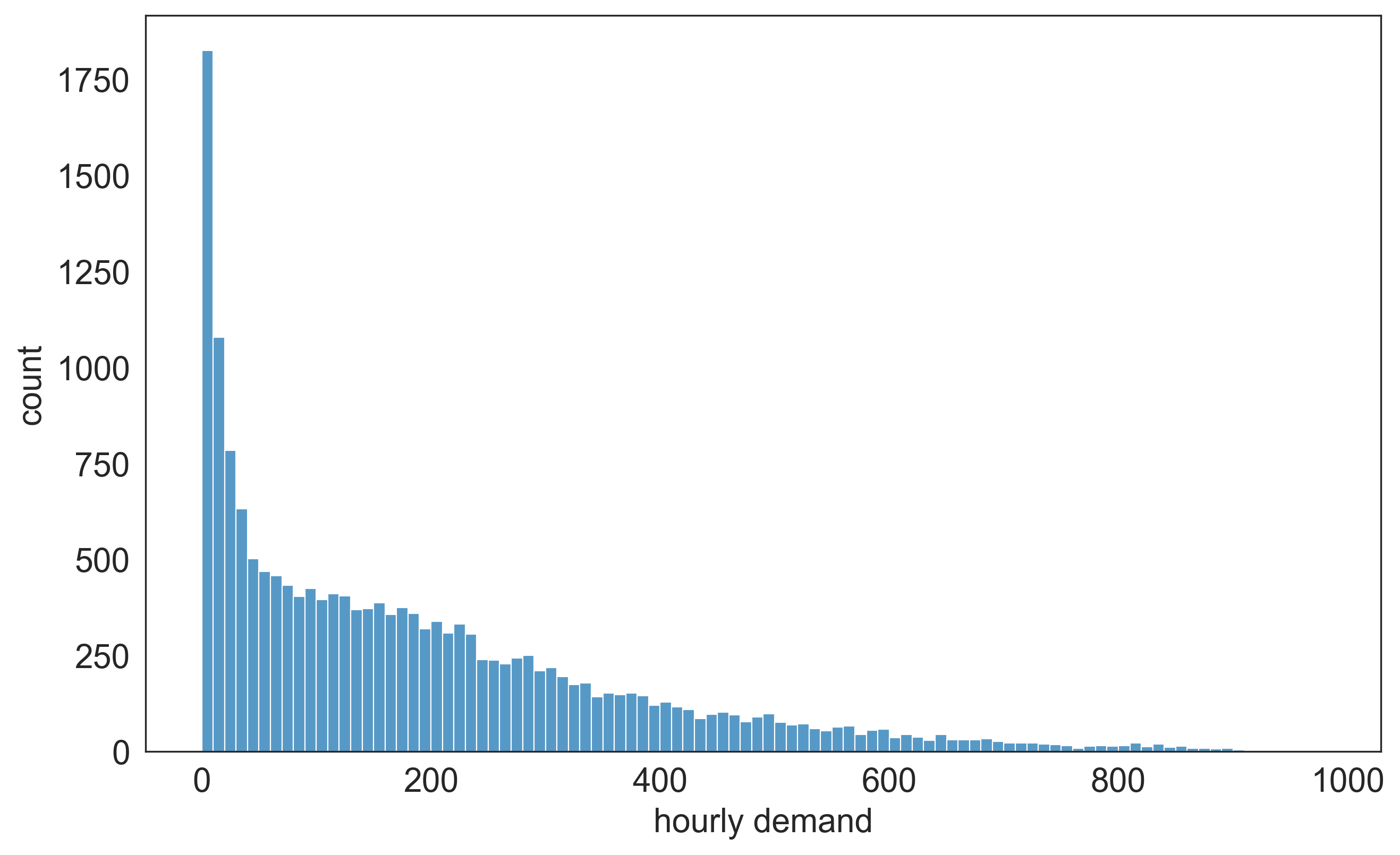}}
    \caption{Hourly demand distribution.}
    \label{fig: bike demand}
\end{figure}

\begin{figure}[!h]
    \centering
    \subfigure[Heatmap of demand by temperature and windspeed.]
    {\includegraphics[width = .48\linewidth]{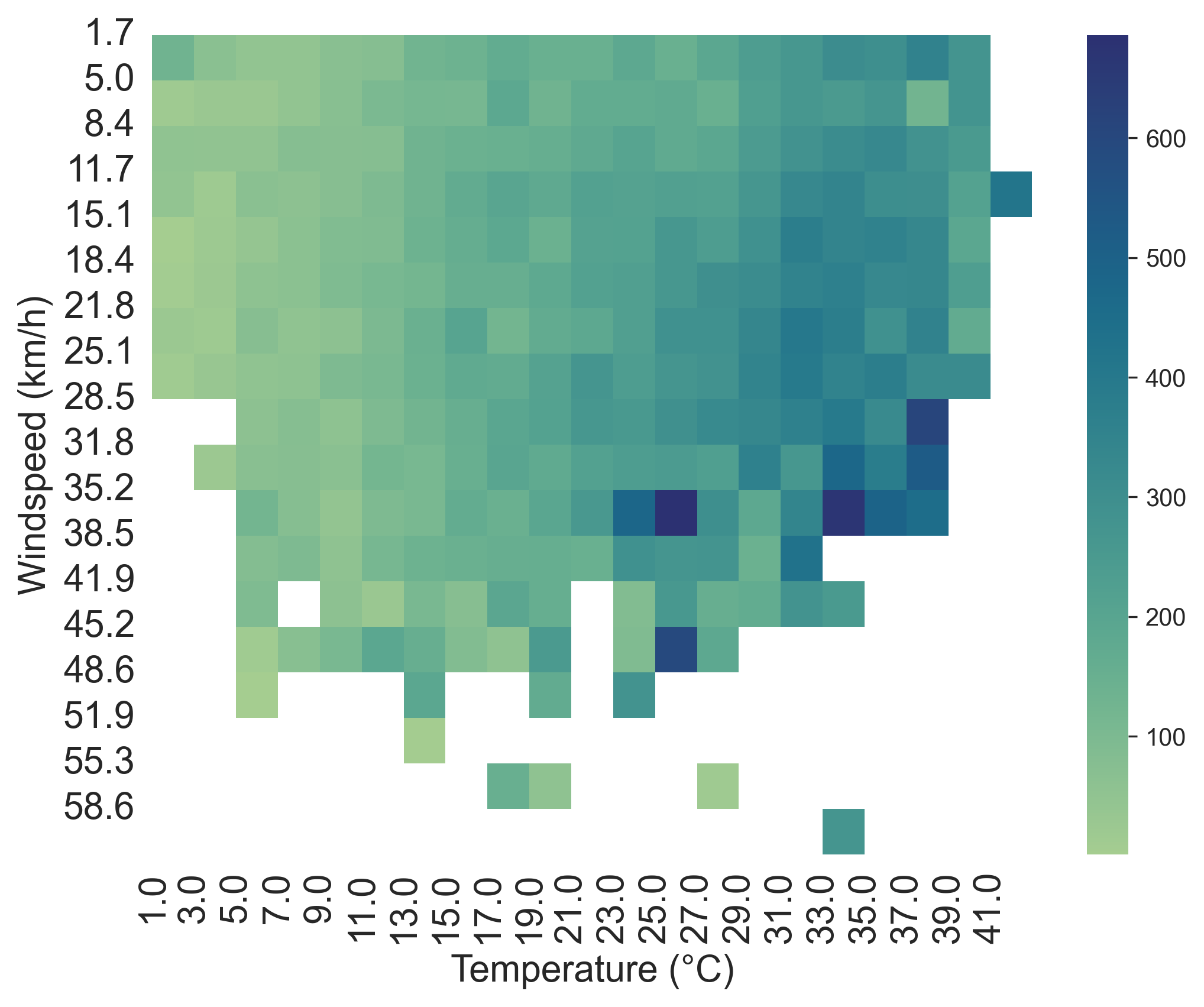}}
    \subfigure[Demand distribution during the day.]
    {\includegraphics[width = .48\linewidth]{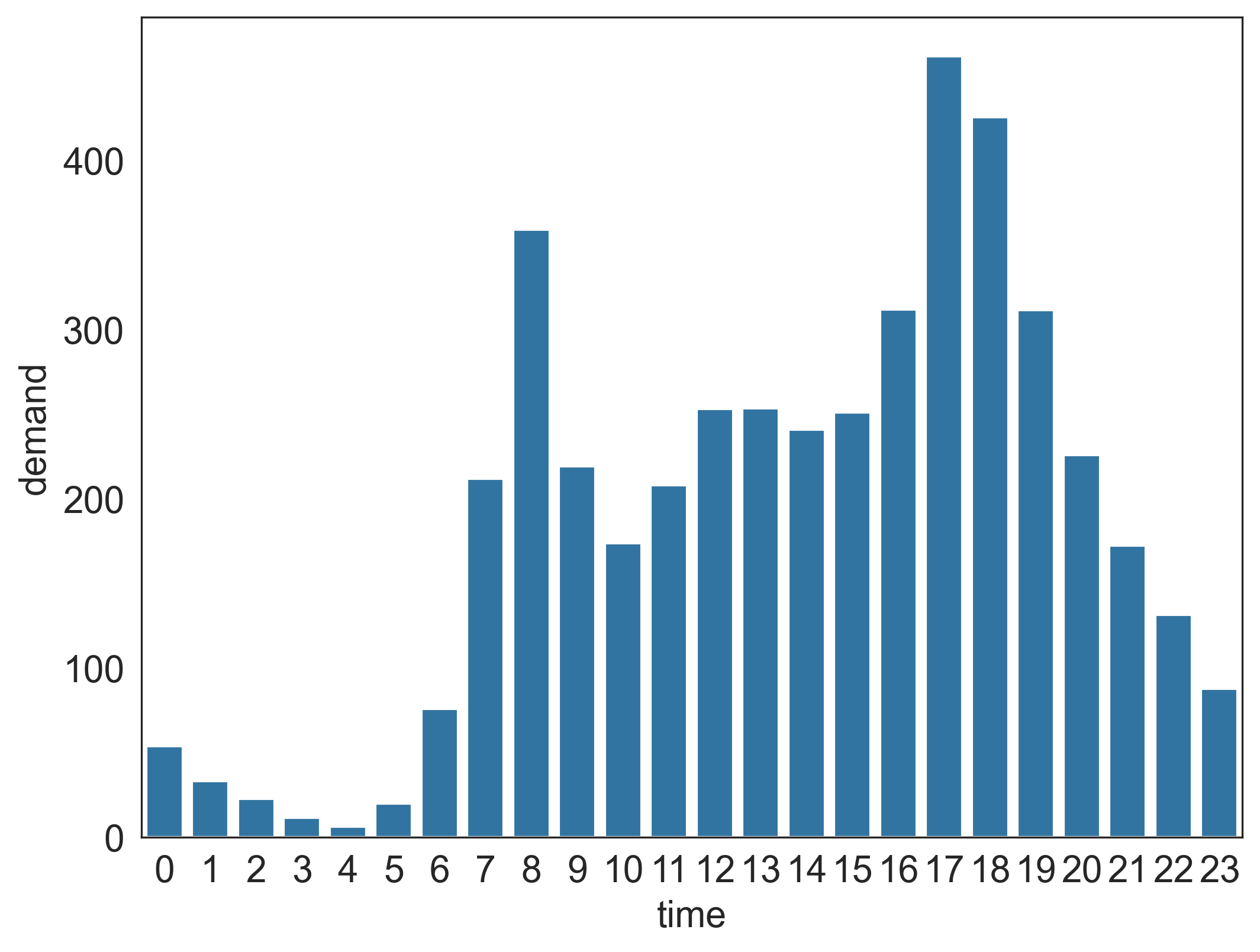}}
    \caption{Demand distribution.}
    \label{fig: bike demand 2}
\end{figure}

 The dataset comprises over 17,000 data points. Figure \ref{fig: bike demand} illustrates the hourly distribution of bike demand. The mean demand is 189, with a standard deviation of 181. The median demand is 142, while the maximum hourly demand reaches 977.
In addition, Figure \ref{fig: bike demand 2}(a) presents a heatmap depicting demand in relation to temperature and windspeed. Notably, demand tends to be lower at lower temperatures. The correlation between demand and temperature is 0.4, whereas the correlation between demand and windspeed is only 0.09.
Figure \ref{fig: bike demand 2}(b) shows the demand distribution throughout the day. There is significant variation in demand, with a pronounced increase occurring between 6 a.m. and 8 a.m. Demand remains relatively stable from 9 a.m. to 4 p.m. before peaking between 5 p.m. and 6 p.m., after which it gradually declines.

We partition the dataset into two subsets: 90\% for training and 10\% for testing. Of the 90\% in the training subset, we subdivide so that 75\% is designated for model training, while the remaining 15\% is reserved for calibration. During the calibration process, the nearest 20 data points (measured using Euclidean distance) are used. Each experiment is conducted over 50 repetitions.

\begin{figure}[h!]
    \centering
    \subfigure[GB]
    {\includegraphics[width = .48\linewidth]{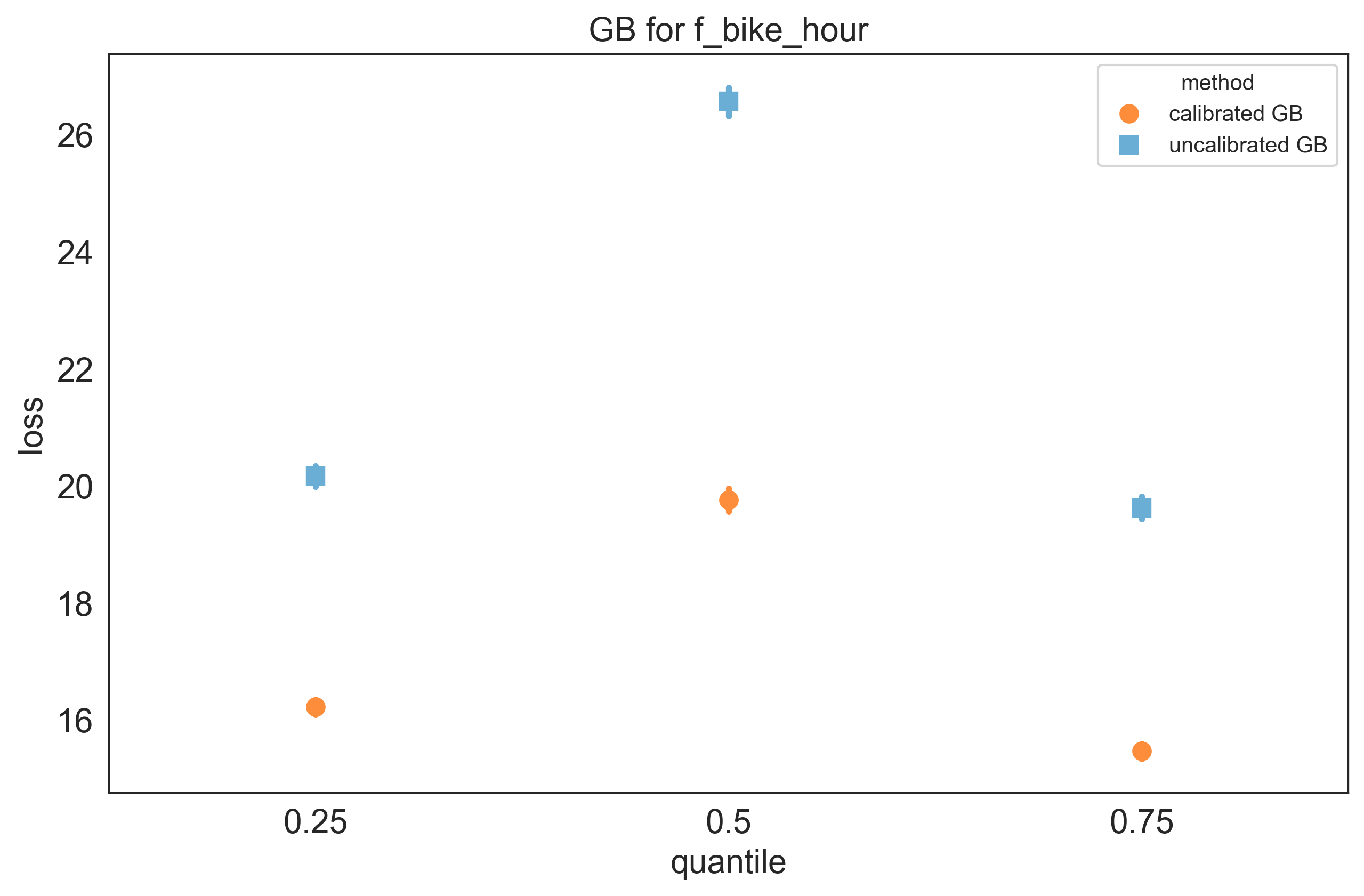}}
    \subfigure[LightGBM]
    {\includegraphics[width = .48\linewidth]{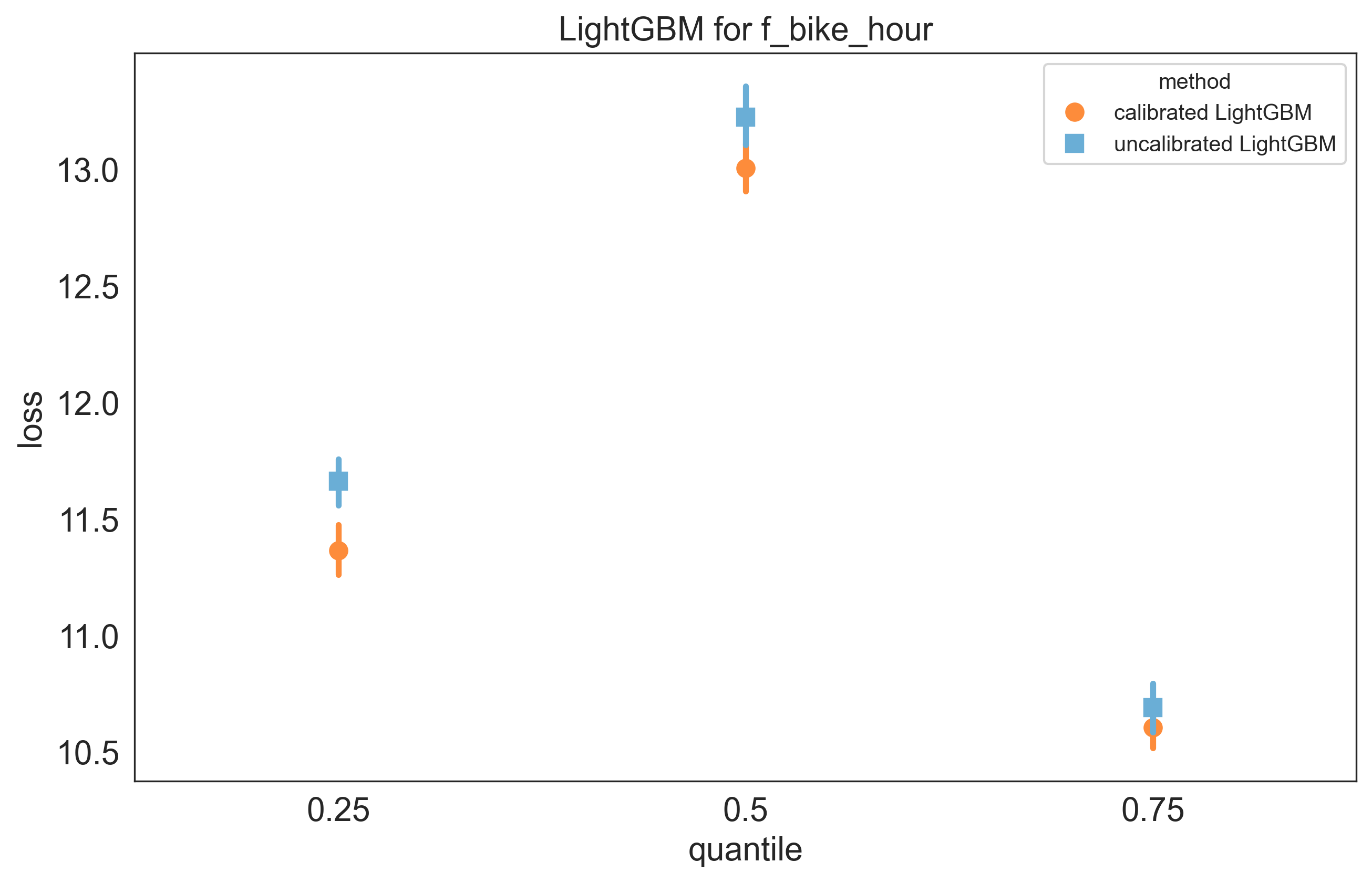}}
    \subfigure[LQR]
    {\includegraphics[width = .48\linewidth]{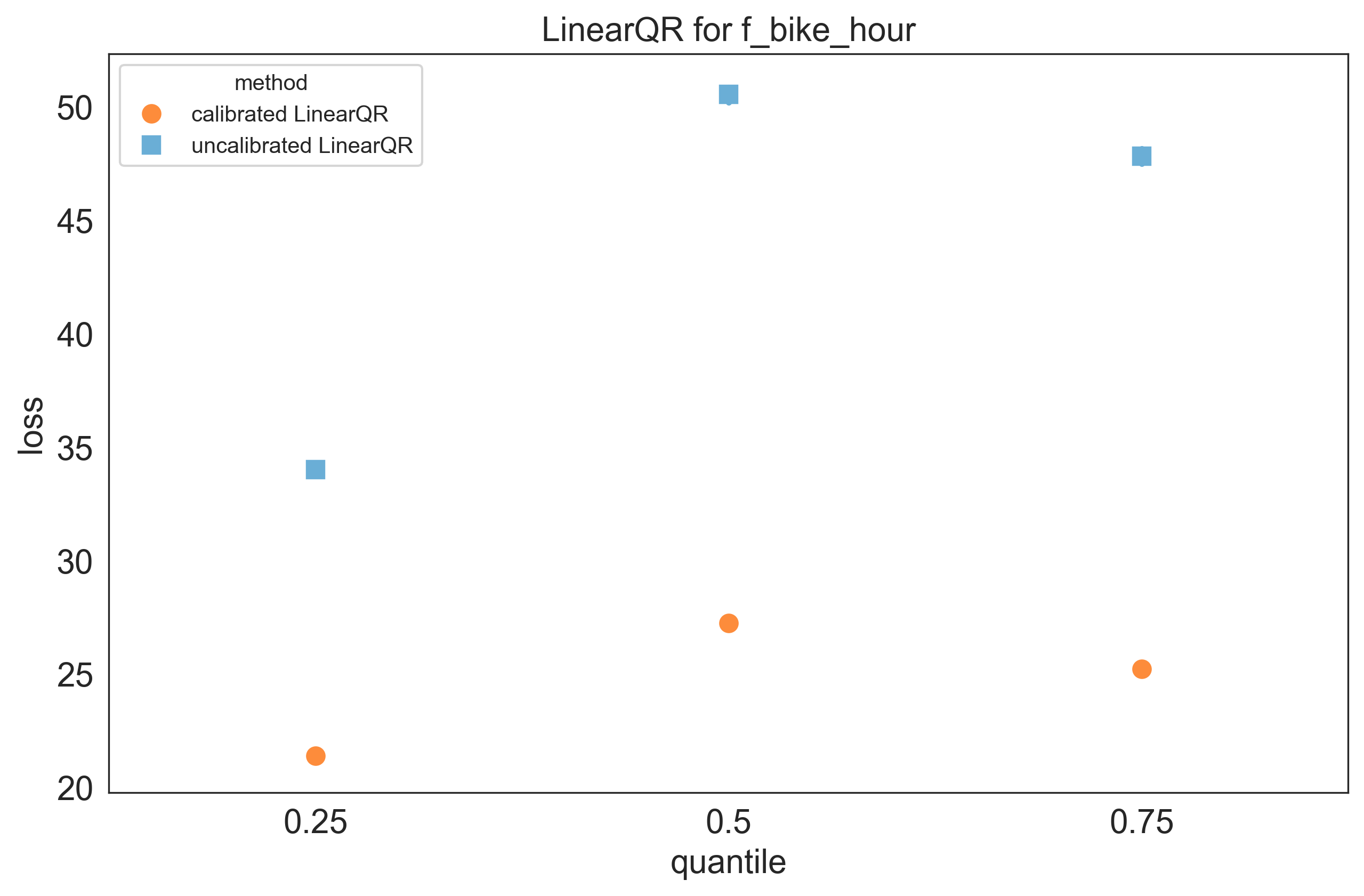}}
     \subfigure[QRNN]
    {\includegraphics[width = .48\linewidth]{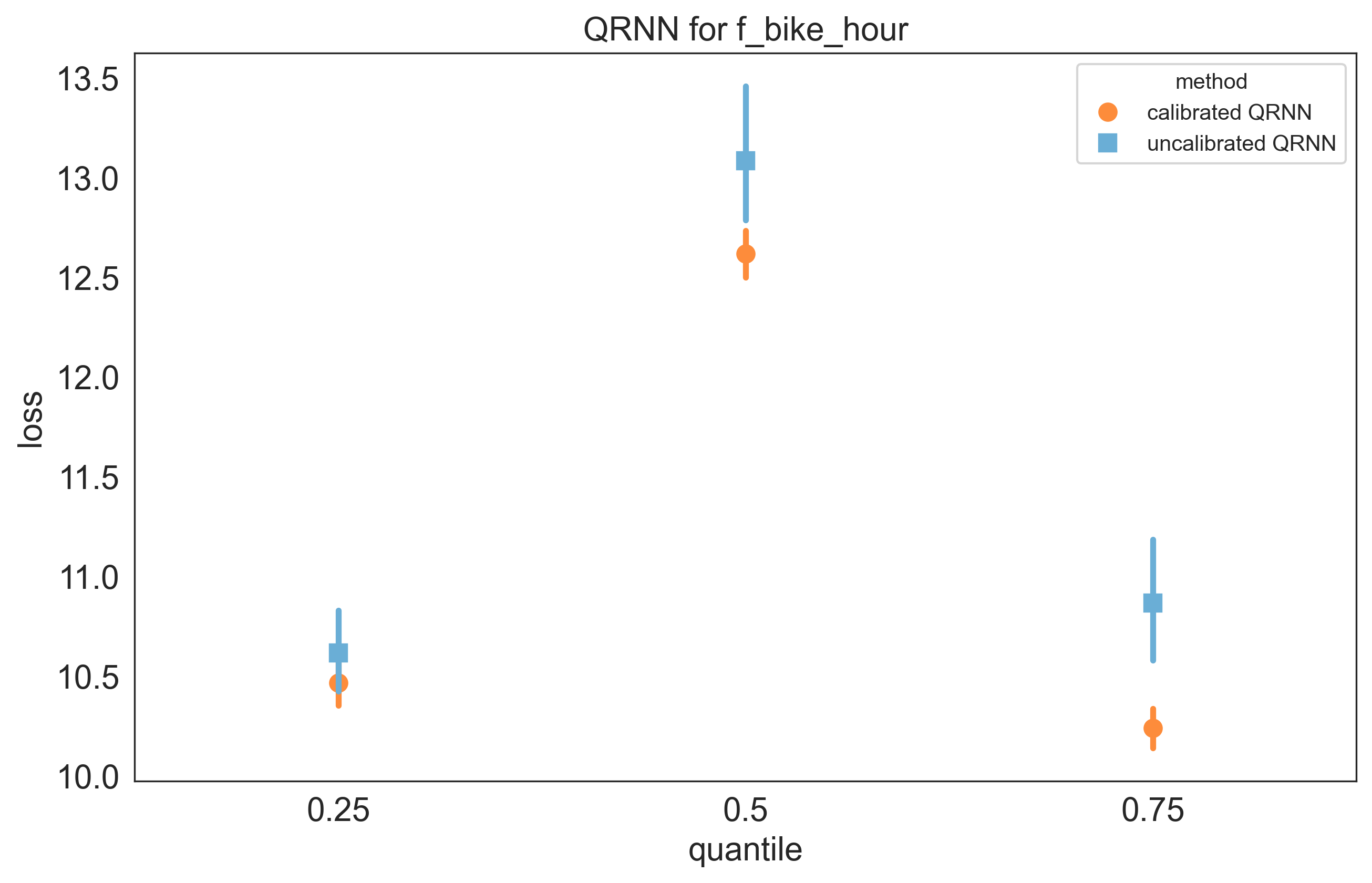}}
    \caption{Empirical pinball loss for MA model.}
    \label{fig: bike hour}
\end{figure}

Figure \ref{fig: bike hour} presents the empirical results for hourly bike demand, demonstrating that our proposed calibration method significantly outperforms the uncalibrated baseline. The improvement is particularly notable for GB and LQR. For GB, the empirical loss is reduced by approximately 19.6\%, 25.6\%, and 21.2\% at quantiles 0.25, 0.5, and 0.75, respectively. Similarly, for LQR, the loss reduction is around 37.1\%, 46.1\%, and 47.3\% at quantiles 0.25, 0.5, and 0.75, respectively. Among the four algorithms tested, QRNN shows the strongest overall performance: The loss reduction is 1.4\%, 3.5\%, and 5.8\% at quantiles 0.25, 0.5, and 0.75, respectively, and LightGBM performs comparably. Interestingly, the improvement from calibration is more pronounced for benchmark algorithms (without the calibration step) that perform worse. This finding suggests that when model misspecification is more severe, the calibration effectively adjusts for bias, leading to greater gains. Although the calibration step reduces bias introduced during model training, it does not entirely eliminate the performance gap between the weaker and stronger baseline algorithms.

\subsection{Implications}
The impact of the additional calibration step on bias correction varies across different algorithms. For highly misspecified function classes, the error reduction is more pronounced. For instance, LinearQR, which assumes a strictly linear structure, has significant model misspecification issue when the true relationship is highly nonlinear, resulting in higher empirical loss. The calibration step helps by capturing higher-order relationships, significantly reducing this loss. In contrast, more complex models, such as deep neural networks, inherently capture some higher-order relationships. Consequently, their model misspecification is less severe, leading to a smaller reduction in error through calibration.

However, we observe that the calibration step does not entirely
eliminate the performance gap between different algorithms.
It is important to note that there are numerous data-driven approaches to calibration and there is a potential for further error reduction. In our numerical experiments, we adopted one of the simplest methods: pooling neighboring data points based on a distance metric that treats all features equally. This approach, while straightforward and easy to implement, may overlook the varying importance of different features. Therefore, if we can somehow better understand and extrapolate how bias depends on individual features, there is a possibility to design more refined calibration techniques and achieve improved performance.

Closing the performance gap between different algorithms is highly desirable, if achievable. If calibration enables different approaches to deliver comparable performance, decision-makers could bypass the often time-consuming process of model selection and focus on selecting the most efficient algorithm. For example, when training a deep neural network, there would be no need to explore various configurations of layers, neurons, or activation functions through cross-validation. Instead, a single structure could be chosen, with calibration handling any residual biases. This streamlined approach would greatly simplify the decision-making process and enhance efficiency.

\section{Concluding Remarks}

This work addresses the prevalent issue of model misspecification in feature-based newsvendor problems, where incorrect model assumptions can lead to significant prediction errors and substantial revenue losses. We introduced a model-free and distribution-free framework inspired by conformal prediction, which can enhance any prediction method by conformalizing model bias and balancing data quality with data quantity. We provided rigorous statistical guarantees for our conformalized quantile, along with confidence intervals for the critical quantile, demonstrating that these intervals shrink as data quality improves and data quantity increases. In addition, our method was validated on both simulated and real-world datasets. The proposed method consistently outperforms benchmark algorithms. We also investigated the effect of local pooling and the effect of sample size.

Our work is a preliminary step toward addressing model misspecification challenges. We see this paper as a prompt for broader discussion and further research. Several potential extensions of our work are worth exploring. First, our method improves performance by selecting an appropriate pooling region, using all historical data within that region. However, if the underlying model is non-stationary over time, considering temporal pooling also becomes necessary. For example, with a growing sales trend, the sales model from last year may differ from this year’s model, highlighting the importance of selecting an appropriate temporal region for data pooling.

Second, while our primary focus has been on the newsvendor problem, exploring the adaptability of our framework to other optimization problems with varying structures could be highly valuable. For instance, how might we design a conformity score function for a general demand prediction problem, particularly in contexts that differ from the single-item inventory structure of the newsvendor problem? Furthermore, when the problem exists in a high-dimensional space, could we develop methods to improve both computational efficiency and predictive accuracy? 
Third, our framework currently trains the model for a specific quantile. In scenarios where overage and underage costs fluctuate frequently, retraining the model to find new critical quantiles may be necessary. Investigating a unified framework that trains the model once to obtain multiple quantiles could significantly improve efficiency in such cases.

\bibliographystyle{informs2014}
\bibliography{ref}

\newpage
\begin{APPENDICES}

\section{Data-Driven Method: Choosing optimal $\dis$ through estimating three functions.}\label{Appendix: data-driven}

 We introduce the third approach, which determines the optimal pooling region by estimating three key functions: $\kappa$, $\barh$, and $\lowh$.

The high-level idea is as follows: We begin by selecting an initial pooling region, $\dis$, using a clustering algorithm. Then, we estimate the functions $\barh$, $\lowh$, and $\kappa$. Based on these estimates, we iteratively update the pooling region $\dis$ according to Equation \eqref{eq: tilde Delta}, repeating this process until satisfactory performance is achieved.

\emph{$\barh$ and $\lowh$ Estimation:}  Recall that $\barh(\Delta)$ and $\lowh(\Delta)$, as defined in Assumption \ref{assum: margin}, satisfy the inequalities:
 $\alpha + \lowh(\Delta)\leq \Pr(Y\leq q_{\alpha}^*(X)+\Delta|X)\leq \alpha + \barh(\Delta)$
 and 
$\alpha- \barh(\Delta)\leq \Pr(Y\leq q_{\alpha}^*(X)-\Delta|X) \leq \alpha- \lowh(\Delta)$. Directly estimating these functions is challenging due to the unknown nature of the true quantile function $q_{\alpha}^*$. We propose an approach to estimate these functions.

First, the data is partitioned into clusters with diameter $\dis$, and a sequence of margin levels $\Delta_1 \leq \Delta_2 \leq \cdots \leq \Delta_k$ is selected. Within each cluster  $\cB_\dis$, let $\hat{f}_\alpha$ be the quantile predictor, which is the solution to the quantile regression problem defined by Equation \eqref{Eq: quantile regression}. We then estimate  $\hat{f}_\alpha$, as well as $\hat{f}_{\alpha\pm \Delta_1}(\cdot),\cdots$, $\hat{f}_{\alpha\pm \Delta_k}(\cdot)$ for different quantiles. The estimated functions $\barh(\Delta)$ and $\lowh(\Delta)$ are then computed as:
\[
\hat{\barh}(\Delta_i)= \max_x \max\left\{\hat{f}_{\alpha+\Delta_i}(x)- \hat{f}_{\alpha}(x), \hat{f}_{\alpha}(x)- \hat{f}_{\alpha-\Delta_i}(x)\right\}
\]
and 
\[
\hat{\lowh}(\Delta_i)= \min_x \min\left\{\hat{f}_{\alpha+\Delta_i}(x)- \hat{f}_{\alpha}(x), \hat{f}_{\alpha}(x)- \hat{f}_{\alpha-\Delta_i}(x)\right\}.
\]
After evaluating the values at $\Delta_1, \cdots, \Delta_k$, we can then apply non-parametric methods, such as kernel smoothing, to estimate the two functions.

\paragraph{$\kappa$ Estimation:} 
Define
\[
\tilde{\kappa}(n_1, \xi(x_1, x_2))=\|(\hat{q}_{n_1,\alpha}(x_1)-q_{\alpha}^*(x_1))-(\hat{q}_{n_1,\alpha}(x_2)-q_{\alpha}^*(x_2))\|.
\]
Here, $\kappa$ represents the tightest upper bound of $\tilde{\kappa}$. However, estimating $\tilde{\kappa}$ is challenging due to the unknown nature of the true quantile function $q_{\alpha}^*$. To address this, we approximate $\tilde{\kappa}$ by bounding it using the difference in empirical loss:
\[
\tilde{\kappa}(n_1, \dis(x_1,x_2))\leq \eta |(\cL(\hat{q}_{n_1,\alpha}(x_1))-\cL(q_{\alpha}^*(x_1)))-(\cL(\hat{q}_{n_1,\alpha}(x_2))-\cL(q_{\alpha}^*(x_2)))|,
\]
where $\eta$ is a constant.
This approximation is grounded in the property that the loss function $\cL(a)$ is Lipschitz continuous, meaning that closeness in empirical loss reflects closeness in quantile prediction errors.

We start by clustering the data based on the diameter $\dis$. Within each cluster of size $|\cB_{\dis}|$, we split the data into a training set $\cB_1$ and a testing set $\cB_2$ according to a fixed proportion. From $\cB_1$, we create multiple training sets by randomly sampling with varying proportions: $\rho_1, \rho_2, \cdots, \rho_k$. This results in training sample sizes of $n_1^1 = |\cB_1|\rho_1$, $n_1^2 = |\cB_1|\rho_2$, $\cdots$, $n_1^k = |\cB_1|\rho_k$. For each training set corresponding to these sample sizes, we compute the quantile predictor, denoted as $\hat{q}_{n_1^j,\alpha}$. For each data point $(x_i, y_i)$ in the testing set, we then calculate the empirical loss, denoted by $\ell(\hat{q}_{n_1^j,\alpha}(x_i), y_i)$. Since the true quantile function $q^*_\alpha(x)$ is unknown, we approximate it using the predictor trained on the testing data, denoted by $\hat{q}^{\text{test}}_\alpha(x)$.

After computing these values, for any pair of points $(x_i,x_j)$ in the testing dataset, we can compute
\[
\hat{\tilde{\kappa}}(n_1,\dis(x_i,x_j))=\eta|(\ell(\hat{q}_{n_1,\alpha}(x_1), y_1)-\ell(\hat{q}_{\alpha}^{test}(x_1), y_1))-(\ell(\hat{q}_{n_1,\alpha}(x_2), y_2)-\ell(\hat{q}_{\alpha}^{test}(x_2), y_2))|,
\]
where $\eta=\frac{1}{|\cB_2|}\sum_{i\in \cB_2}\frac{\hat{q}_{n_1,\alpha}(x_i)}{\ell(\hat{q}_{n_1,\alpha}(x_i), y_i)}$.
 Let $\cD=\{(n_1,\dis(x_i,x_j)), \hat{\tilde{\kappa}}(n_1,\dis(x_i,x_j)), \forall  i,j\in \cB_2\}$. We estimate the two-dimensional function from the dataset $\cD$ using non-parametric methods, such as kernel smoothing.
Once $\barh$, $\lowh$, and $\kappa$ are estimated, we select the optimal $\xi$ according to Equation \eqref{eq: tilde Delta}.

The pseudo-code for this data-driven method of selecting the pooling diameter is outlined in Algorithm \ref{Alg: Pooling}. We repeat the process of function estimation and $\dis$ optimization for several rounds until the stopping criterion is satisfied. The stopping criterion could be based on the stabilization of the pooling diameter $\dis$ or reaching a predefined number of rounds. Once $\dis$ is selected, we proceed with Algorithm \ref{Alg: CCQP}.

\begin{algorithm}
   \caption{\textnormal{\textsf{Data-driven method for selecting the pooling diameter}}}\label{Alg: Pooling}
\begin{algorithmic}[1]
\State Initialize $\dis=\dis_0$;
\While{the stopping criteria is not satisfied}
\State Select data according to the pooling diameter $\dis$;
\State Compute the predictor from Equation \eqref{Eq: quantile regression};
\State Compute $\hat{\barh}(\Delta_i)$ and $\hat{\lowh}(\Delta_i)$ on varying levels $\Delta_i$; 
\State Compute $\hat{\tilde{\kappa}}(n_1,\dis(x_i,x_j))$ on different levels of $n_1$ and $\xi$; estimate the function of $\kappa$;
\State Fit functions $\barh$, $\lowh$, and $\kappa$;
\State Select $\dis$ as a solution to Equation \eqref{eq: tilde Delta};
\EndWhile
\State Implement Algorithm \ref{Alg: CCQP} with pooling diameter $\dis$.
\end{algorithmic}
\end{algorithm}

Though Algorithm \ref{Alg: Pooling} is more complicated to implement, the advantage of Algorithm \ref{Alg: Pooling} is that once three functions are estimated in a good way, the optimal pooling diameter can be directly computed via Equation \eqref{eq: tilde Delta} when the size of dataset changes. Thus, this procedure only needs to be run for a few times.

\section{Additional Numerical Results}\label{appendix: numerical}
Figures \ref{fig: MA model data ratio 3} and \ref{fig: ML model data ratio 3} show the empirical pinball loss for the MA and ML models with varying data sizes. Prediction performance improves as the data size increases, with a data size ratio of 0.1 resulting in significantly worse performance and higher variance compared to a ratio of 0.3.

\begin{figure}[H]
    \centering
    \subfigure[GB]
    {\includegraphics[width = .45\linewidth]{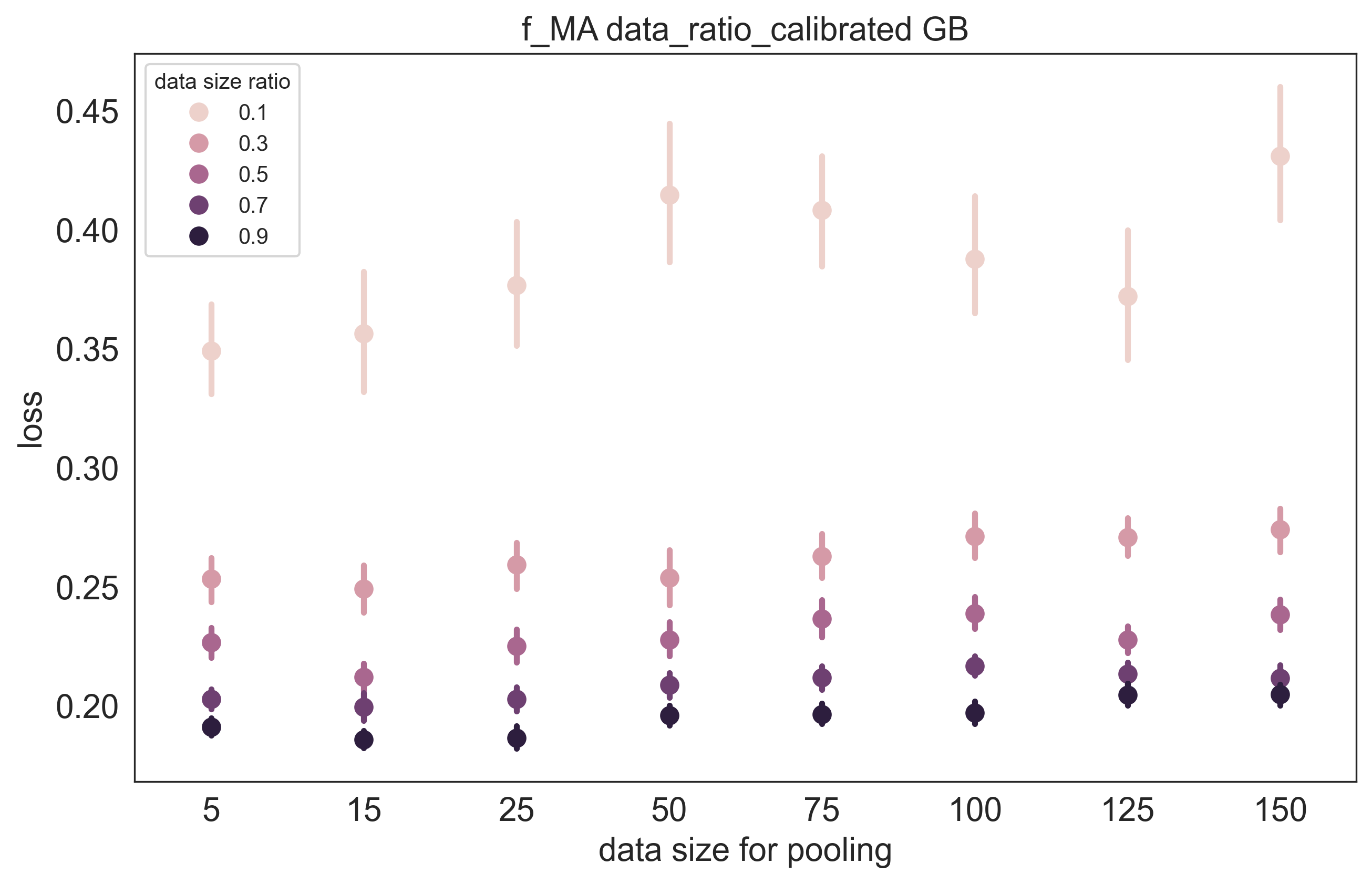}}
    \subfigure[LightGBM]
    {\includegraphics[width = .45\linewidth]{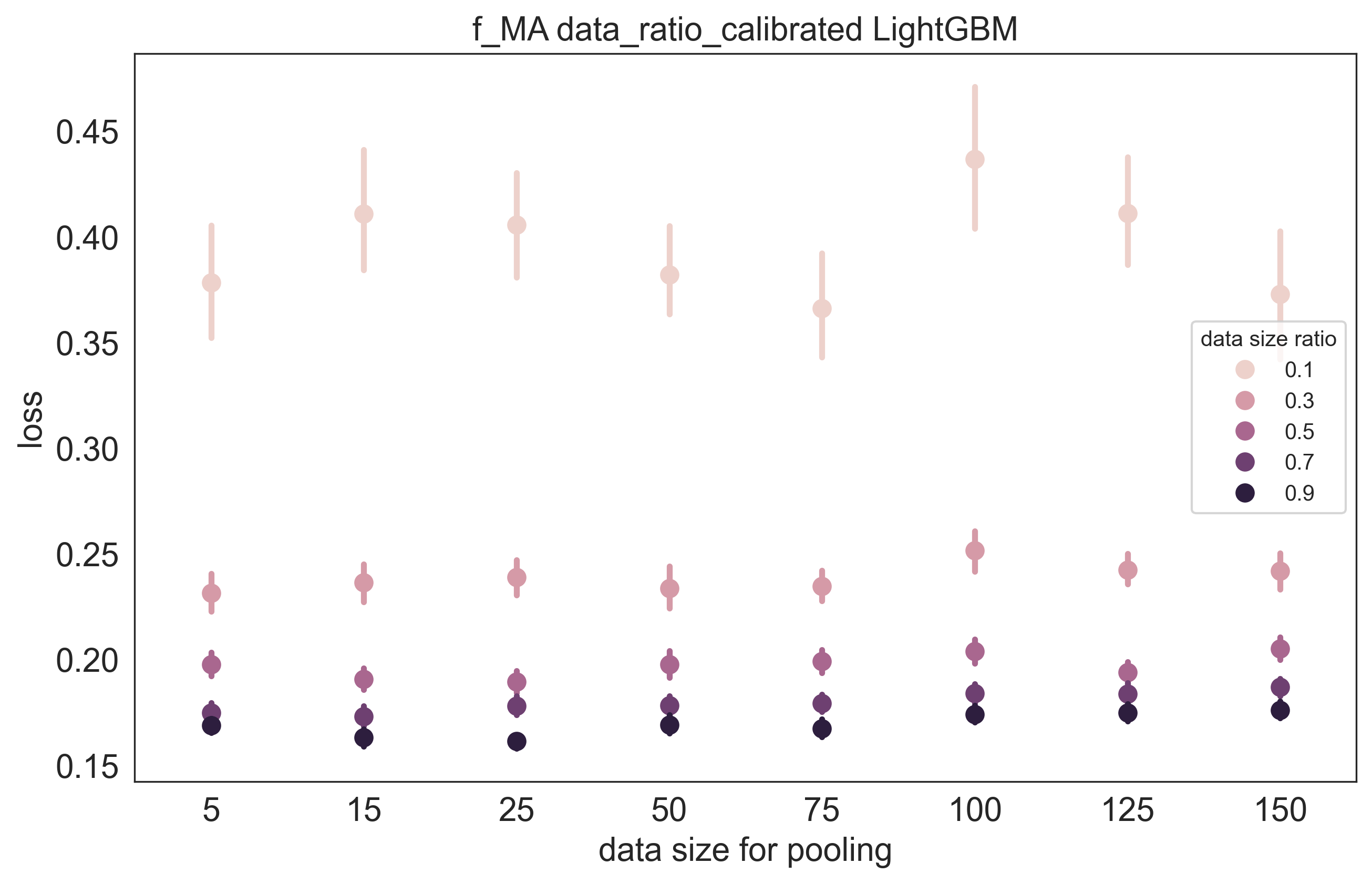}}
    \subfigure[LinearQR]
    {\includegraphics[width = .45\linewidth]{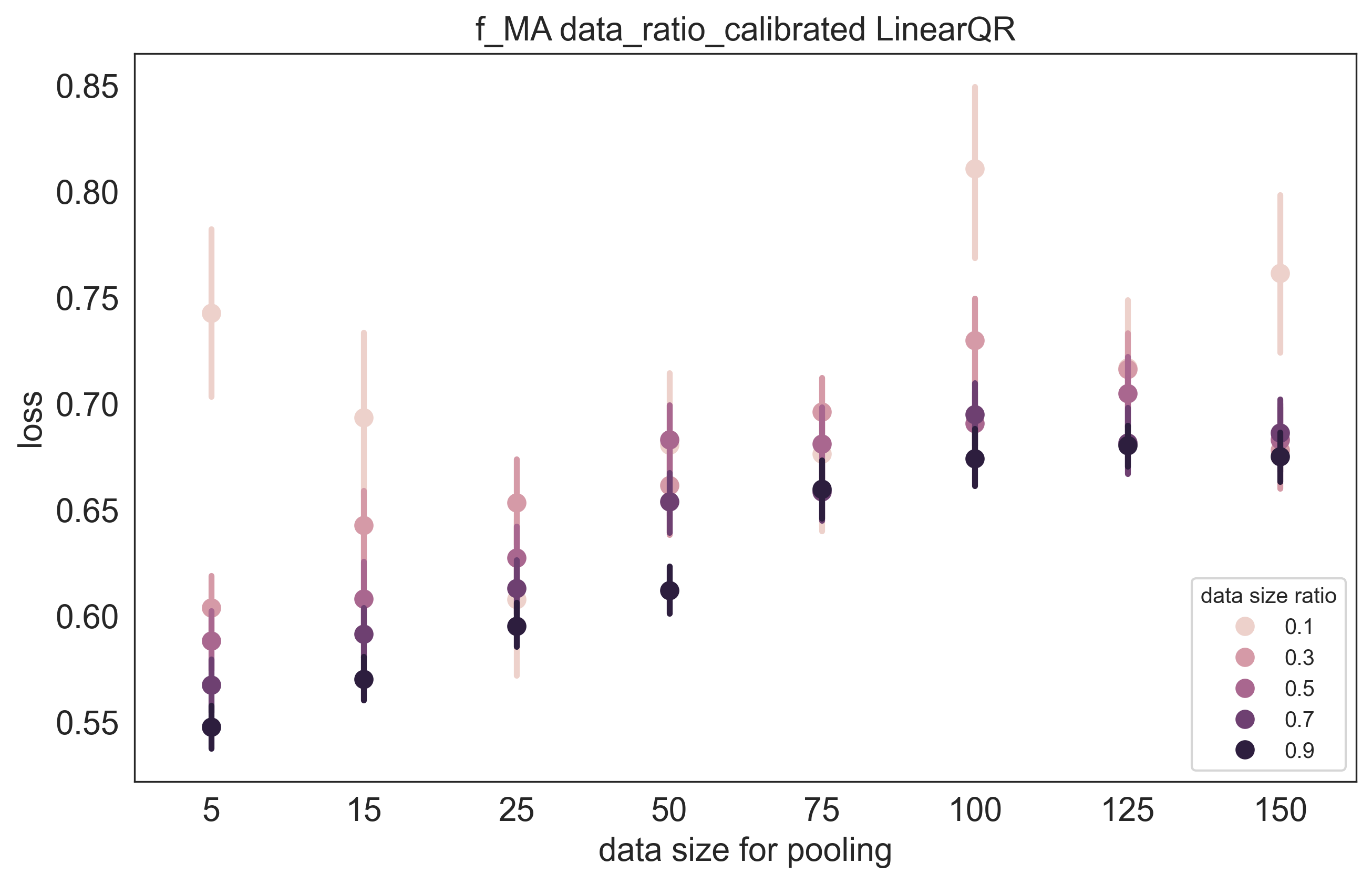}}
     \subfigure[QRNN]
    {\includegraphics[width = .45\linewidth]{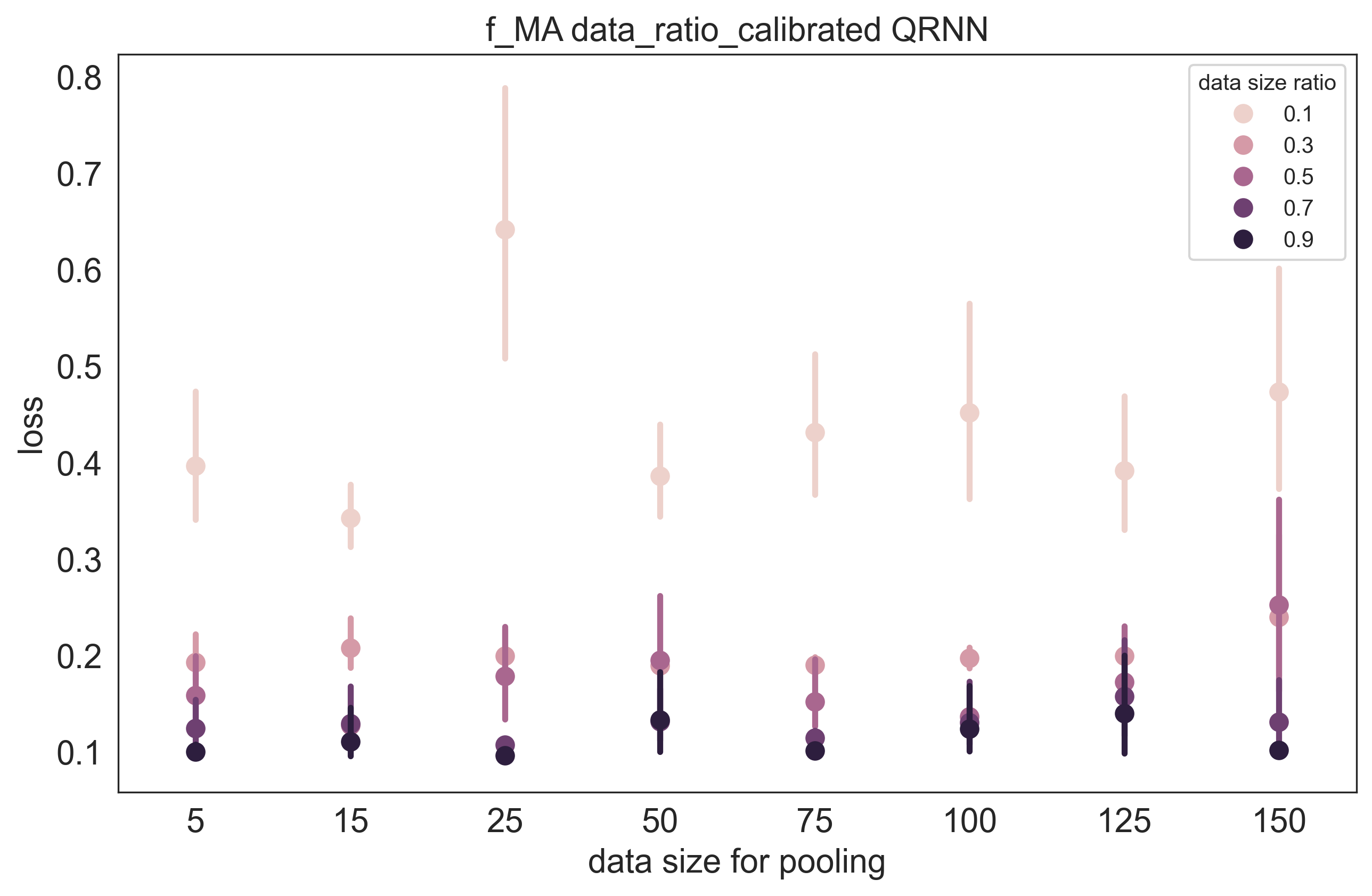}}
    \caption{Empirical pinball loss for MA model with different data sizes.}
    \label{fig: MA model data ratio 3}
\end{figure}

\begin{figure}[H]
    \centering
    \subfigure[Gradient Boosting]
    {\includegraphics[width = .45\linewidth]{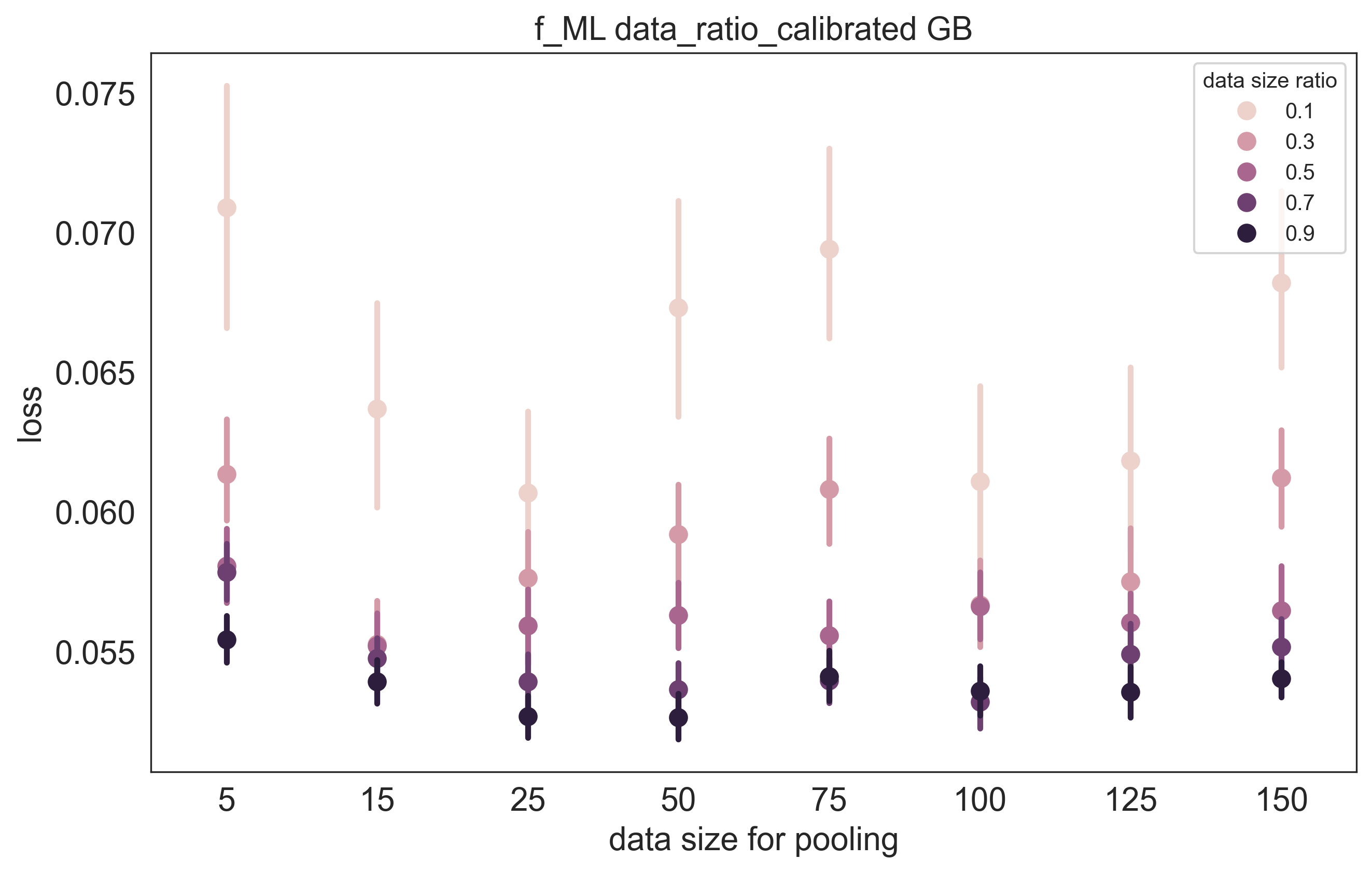}}
    \subfigure[LightGBM]
    {\includegraphics[width = .45\linewidth]{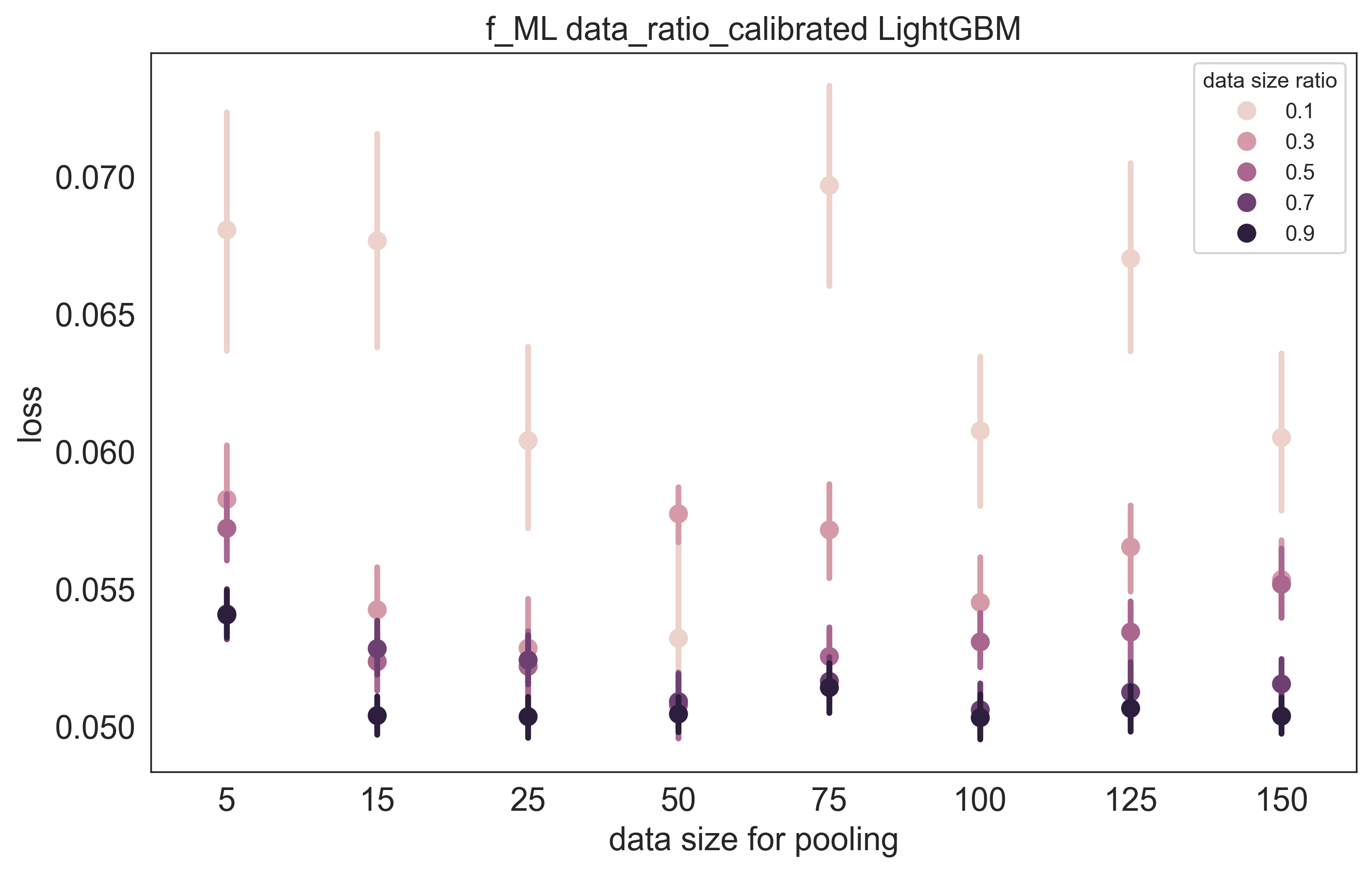}}
    \subfigure[LinearQR]
    {\includegraphics[width = .45\linewidth]{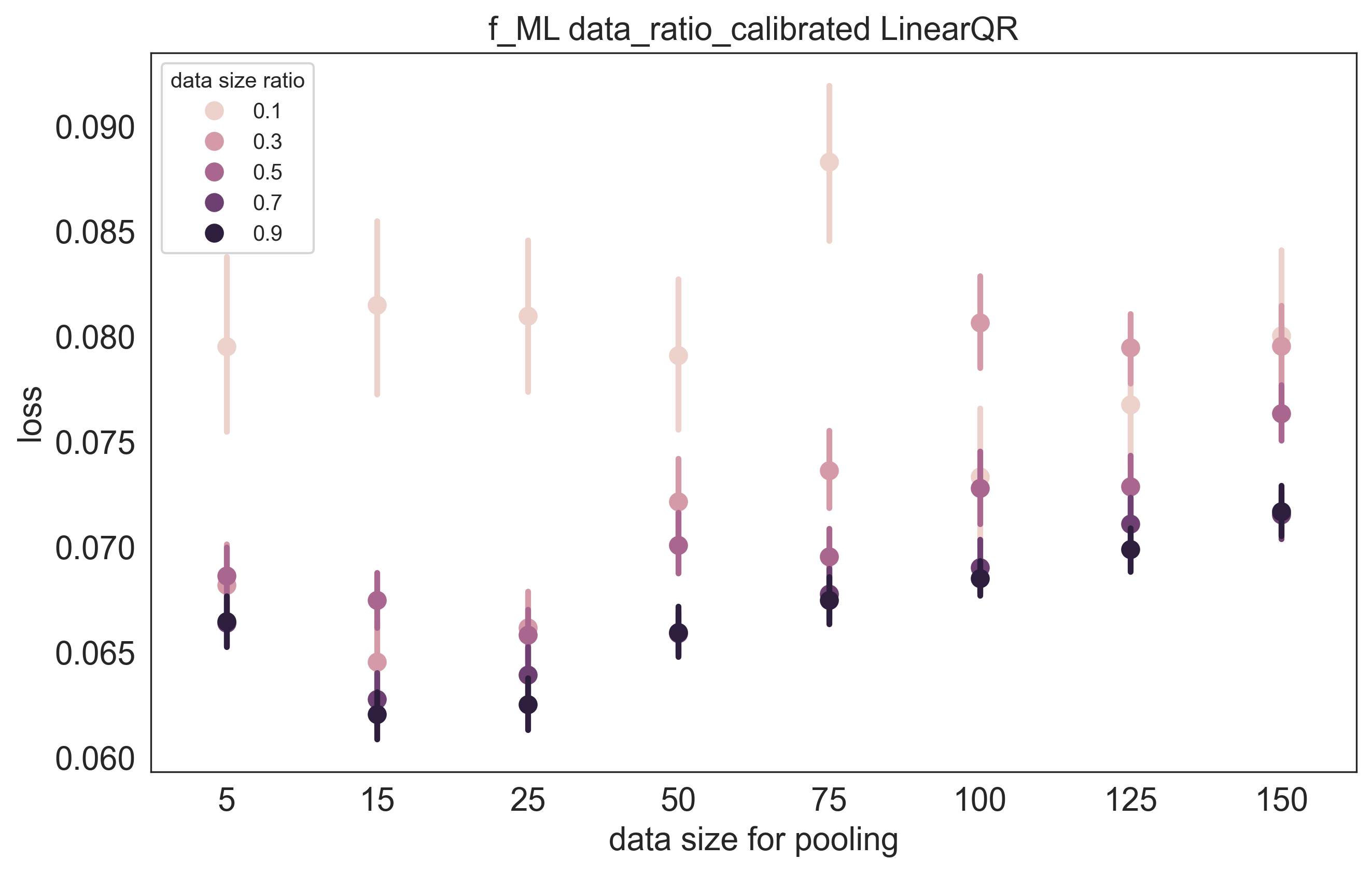}}
     \subfigure[QRNN]
    {\includegraphics[width = .45\linewidth]{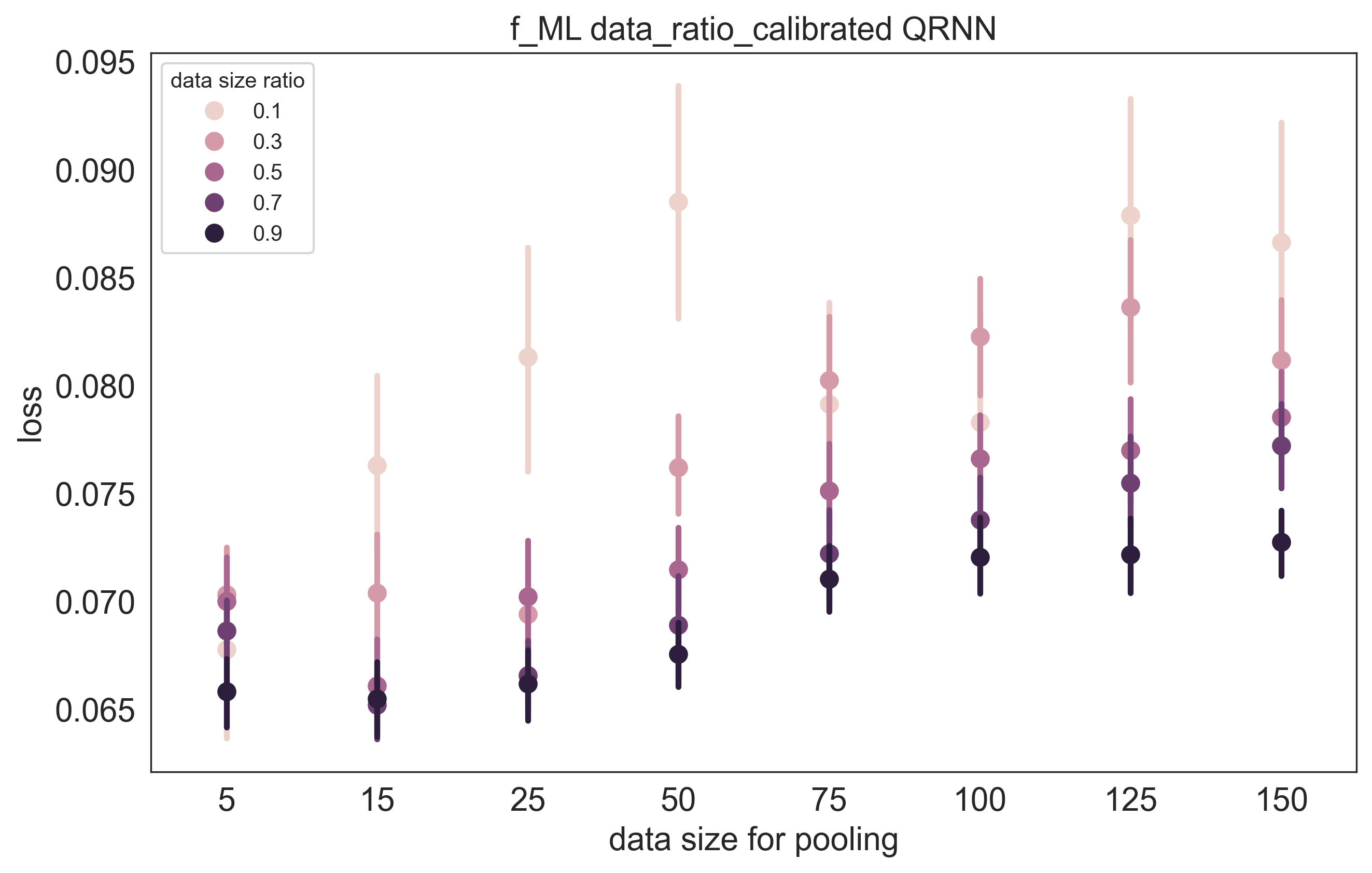}}
    \caption{Empirical pinball loss for ML model with different data sizes.}
    \label{fig: ML  model data ratio 3}
\end{figure}

\end{APPENDICES}

\end{document}